\documentclass[10pt,journal,compsoc]{IEEEtran}

\usepackage{amsmath}
\usepackage{gensymb}
\usepackage{diagbox}
\usepackage{graphics}
\usepackage{epsfig}
\usepackage{threeparttable}
\usepackage{xspace}
\usepackage{siunitx}
\usepackage{graphicx}
\usepackage{subfig}
\usepackage{placeins}
\usepackage{blindtext}
\usepackage{amssymb}
\usepackage{threeparttable}
\usepackage{color}
\usepackage[normalem]{ulem}
\usepackage{multirow}
\usepackage{float}
\usepackage{wrapfig}
\usepackage{amsfonts}
\usepackage{bm}
\usepackage{bbm}
\usepackage{array}
\usepackage{tabulary}	
\usepackage{microtype}
\usepackage{booktabs}
\usepackage{xspace}
\usepackage[table]{xcolor}
\usepackage{colortbl}
\usepackage{diagbox}
\usepackage{rotating}
\usepackage{booktabs}
\usepackage{overpic}
\usepackage{enumitem}
\usepackage{colortbl}
\usepackage{soul}
\usepackage{url}
\usepackage{makecell, verbatim, sidecap}
\usepackage{algorithm}
\usepackage{algorithmic}
\usepackage{cite}
\usepackage{tikz}
\usetikzlibrary{positioning}
\usepackage{mathrsfs}
\usepackage{longtable}
\usepackage{multicol}
\usepackage{calc}
\usepackage{pifont}

\usepackage{tikz}
\usepackage[edges]{forest}
\definecolor{hidden-draw}{RGB}{20,68,106}
\definecolor{hidden-pink}{RGB}{255,245,247}
\usepackage{amssymb}
\usepackage{placeins}

\definecolor{citecolor}{HTML}{0071bc}
\usepackage[pagebackref=true,breaklinks=true,colorlinks,citecolor=citecolor,bookmarks=false]{hyperref}

\definecolor{scorered}{HTML}{e4485a}
\definecolor{scoreblue}{HTML}{4a7ee8}
\definecolor{scoregreen}{HTML}{80ba0e}

\definecolor{purple0}{HTML}{e9e9f3}
\definecolor{purple}{HTML}{dcdaed}
\definecolor{purple1}{HTML}{bab6da}
\definecolor{blue0}{HTML}{b6d9f0}
\definecolor{blue1}{HTML}{80b1d1}
\definecolor{blue2}{HTML}{2a8ed1}
\definecolor{blue3}{HTML}{0071bc}

\definecolor{mygray00}{gray}{.3}
\definecolor{mygray0}{gray}{.6}
\definecolor{mygray}{gray}{.85}
\definecolor{mygray1}{gray}{.9}
\definecolor{mygray2}{gray}{.95}

\newcommand{\myparagraph}[1]{
\vspace{0.2cm}\noindent
\textbullet\hspace{0.1cm}\textbf{#1.}
}

\makeatletter
\newcommand{\thickhline}{%
    \noalign {\ifnum 0=`}\fi \hrule height 1pt
    \futurelet \reserved@a \@xhline
}
\makeatother

\makeatletter
\DeclareRobustCommand\onedot{\futurelet\@let@token\@onedot}
\def\@onedot{\ifx\@let@token.\else.\null\fi\xspace}

\def\eg{\emph{e.g}\onedot} 
\def\ie{\emph{i.e}\onedot}

\def\etal{\emph{et al}\onedot}

\makeatother

\newcommand{\app}{\raise.17ex\hbox{$\scriptstyle\sim$}}

\usepackage{xcolor}
\usepackage{caption}
\usepackage{etoolbox} %

\definecolor{revcolor}{RGB}{0,0,0}
\definecolor{revcolor1}{RGB}{0,0,0}

\makeatletter
\@ifpackageloaded{hyperref}{
  \newcommand{\revpdfstring}[2]{\texorpdfstring{#1}{#2}}
}{
  \newcommand{\revpdfstring}[2]{#1}
}
\makeatother

\newenvironment{revrange}{%
  \begingroup
  \color{revcolor}%
  \captionsetup{labelfont={color=revcolor}, textfont={color=revcolor}}%
  \let\revSavedSection\section
  \let\revSavedSubsection\subsection
  \let\revSavedSubsubsection\subsubsection
  \let\revSavedParagraph\paragraph
  \renewcommand{\section}[1]{%
    \revSavedSection{\revpdfstring{\textcolor{revcolor}{##1}}{##1}}}%
  \renewcommand{\subsection}[1]{%
    \revSavedSubsection{\revpdfstring{\textcolor{revcolor}{##1}}{##1}}}%
  \renewcommand{\subsubsection}[1]{%
    \revSavedSubsubsection{\revpdfstring{\textcolor{revcolor}{##1}}{##1}}}%
  \renewcommand{\paragraph}[1]{%
    \revSavedParagraph{\revpdfstring{\textcolor{revcolor}{##1}}{##1}}}%
}{%
  \endgroup
}

\newenvironment{revrange1}{%
  \begingroup
  \color{revcolor1}%
  \captionsetup{labelfont={color=revcolor1}, textfont={color=revcolor1}}%
  \let\revSavedSection\section
  \let\revSavedSubsection\subsection
  \let\revSavedSubsubsection\subsubsection
  \let\revSavedParagraph\paragraph
  \renewcommand{\section}[1]{%
    \revSavedSection{\revpdfstring{\textcolor{revcolor1}{##1}}{##1}}}%
  \renewcommand{\subsection}[1]{%
    \revSavedSubsection{\revpdfstring{\textcolor{revcolor1}{##1}}{##1}}}%
  \renewcommand{\subsubsection}[1]{%
    \revSavedSubsubsection{\revpdfstring{\textcolor{revcolor1}{##1}}{##1}}}%
  \renewcommand{\paragraph}[1]{%
    \revSavedParagraph{\revpdfstring{\textcolor{revcolor1}{##1}}{##1}}}%
}{%
  \endgroup
}

\begin{document}
\title{Controllable Generation with Text-to-Image Diffusion Models: A Survey}

\author{Pu Cao, Feng Zhou, Qing Song, Lu Yang

\IEEEcompsocitemizethanks{
\IEEEcompsocthanksitem Pu Cao, Feng Zhou, Qing Song, Lu Yang are with the Beijing University of Posts and Telecommunications,  Beijing, 100876, China (e-mail: caopu@bupt.edu.cn; zhoufeng@bupt.edu.cn; priv@bupt.edu.cn; soeaver@bupt.edu.cn)
\IEEEcompsocthanksitem Corresponding author: Lu Yang (email: soeaver@bupt.edu.cn)
\IEEEcompsocthanksitem This work was supported by the Young Scientists Fund of NSFC (Grant No. 62406035).
}
}

\markboth{IEEE TRANSACTIONS ON PATTERN ANALYSIS AND MACHINE INTELLIGENCE}%
{Shell \MakeLowercase{\textit{\etal}}: Bare Demo of IEEEtran.cls for Journals}

\IEEEtitleabstractindextext{
\begin{abstract}
In the rapidly advancing realm of visual generation, diffusion models have revolutionized the landscape, marking a significant shift in capabilities with their impressive text-guided generative functions. 
However, relying solely on text for conditioning these models does not fully cater to the varied and complex requirements of different applications and scenarios. Acknowledging this shortfall, a variety of studies aim to control pre-trained text-to-image (T2I) models to support novel conditions.
In this survey, we undertake a thorough review of the literature on controllable generation with T2I diffusion models, covering both the theoretical foundations and practical advancements in this domain. Our review begins with a brief introduction to the basics of denoising diffusion probabilistic models (DDPMs) and widely used T2I diffusion models.
Additionally, we provide a detailed overview of research in this area, categorizing it from the condition perspective into three directions: generation with specific conditions, generation with multiple conditions, and universal controllable generation. 
For each category, we analyze the underlying control mechanisms and review representative methods based on their core techniques.
For an exhaustive list of the controllable generation literature surveyed, please refer to our curated repository at \url{https://github.com/PRIV-Creation/Awesome-Controllable-T2I-Diffusion-Models}.

\end{abstract}
}
\maketitle
\IEEEdisplaynontitleabstractindextext
\IEEEpeerreviewmaketitle

\IEEEraisesectionheading{\section{Introduction}\label{sec:introduction}}
\label{sec:intro}
\IEEEPARstart{D}{iffusion} models, representing a paradigm shift in the visual generation, have dramatically outperformed traditional frameworks like Generative Adversarial Networks (GANs)\cite{goodfellow2014generative,karras2019style,karras2020analyzing,karras2021alias}. As parameterized Markov chains, diffusion models exhibit a remarkable ability to transform random noise into intricate images, progressing sequentially from noise to high-fidelity visual representations. 
With the advancement of technology, diffusion models have demonstrated immense potential in image generation and related downstream tasks.

As the quality of imagery generated by these models advances, a critical challenge becomes increasingly apparent: achieving precise control over these generative models to fulfill complex and diverse human needs. This task goes beyond simply enhancing image resolution or realism; it involves meticulously aligning the generated output with the user's specific and nuanced requirements as well as their creative aspirations.
Fueled by the advent of extensive multi-modal text-image datasets~\cite{krishna2017visual,lin2014microsoft,schuhmann2021laion,schuhmann2022laion} and development of guidance mechanism~\cite{dhariwal2021diffusion,song2021scorebased,ho2022classifier,nichol2021glide}, text-to-image (T2I) diffusion models have emerged as a cornerstone in the controllable visual generation landscape~\cite{ramesh2021zero,nichol2021glide,rombach2022high,saharia2022photorealistic,podell2023sdxl,balaji2022ediffi}. 
These models are capable of generating realistic, high-quality images that accurately reflect the descriptions provided in natural language.

\begin{figure}
    \centering
    \subfloat[Yearly paper count.]{
        \includegraphics[width=0.45\textwidth]{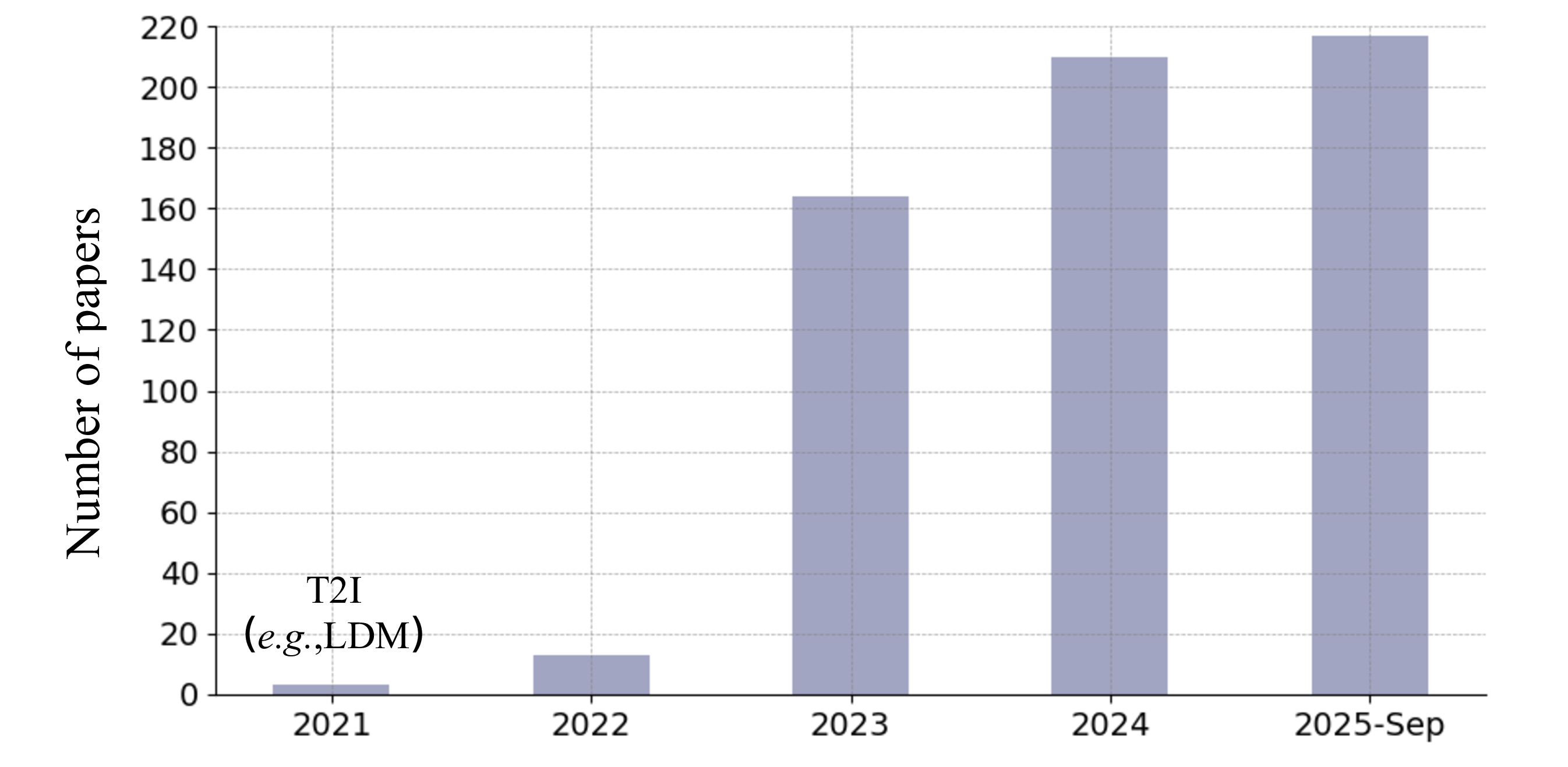}
        \label{fig:number}
    }\\
    \subfloat[Schematic diagram of controllable generation.]{
        \includegraphics[width=0.45\textwidth]{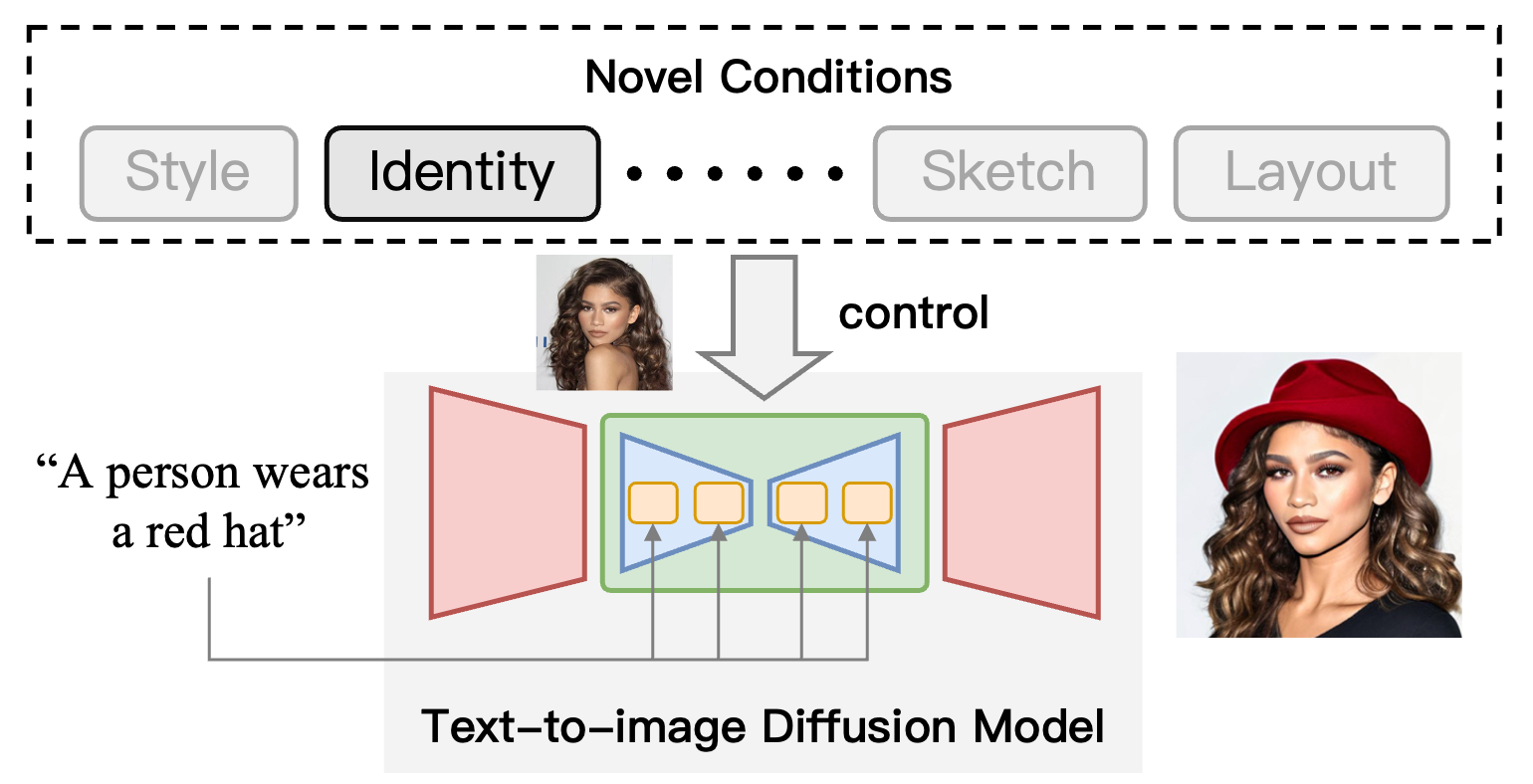}
        \label{fig:cond_gen}
    }
    \caption{\textbf{An overview of conditional generation with T2I diffusion model.} (a) We plot the number of papers on controllable generation based on T2I diffusion models, implying that it is increasing rapidly after powerful generators are released. (b) We present a schematic illustration of controllable generation using the T2I diffusion model, where novel conditions beyond text are introduced to steer the outcomes. Example images are sourced from \cite{chen2023photoverse}.}
    \label{fig:overview}
\vspace{-4mm}
\end{figure}

While text-based conditions have been instrumental in propelling the field of controllable generation forward, they inherently lack the capability to fully satisfy all user requirements. This limitation is particularly evident in scenarios where conditions, such as the depiction of an unseen person or a distinct art style, are not effectively conveyable through text prompts alone. These scenarios pose significant challenges in the T2I generation process, as the nuances and complexities of such visual representations are difficult to encapsulate in text form. 
Recognizing this gap, a substantial body of research has shifted focus towards integrating novel conditions that extend beyond the confines of textual descriptions into T2I diffusion models. This pivot has been further facilitated by the emergence of powerful and open-sourced T2I diffusion models, as illustrated in Fig.~\ref{fig:number}. These advancements have led to the exploration of diverse conditions, thereby enriching the spectrum of possibilities for conditional generation and addressing the more intricate and nuanced demands of users in various applications.

There are numerous survey articles exploring the AI-generated content (AIGC) domain, including diffusion model theories and architectures~\cite{yang2023diffusion}, efficient diffusion models~\cite{ulhaq2022efficient}, multi-modal image synthesis and editing~\cite{zhan2023multimodal}, visual diffusion model~\cite{croitoru2023diffusion,xing2023survey}, and text-to-3D applications~\cite{li2023generative}. 
However, they often provide only a cursory brief of controlling text-to-image diffusion models or predominantly focus on alternative modalities. This lack of in-depth analysis of the integration and impact of novel conditions in T2I models highlights a critical area for future research and exploration.

This survey presents a comprehensive review of controllable generation with text-to-image diffusion models, covering both theoretical foundations and practical advancements. We begin with a concise overview of T2I diffusion models, introducing a brief summary of the theory and widely adopted text-to-image models. We then provide an in-depth examination of prior studies from a technical perspective, analyzing their theoretical underpinnings and highlighting their distinctive contributions and characteristics. This discussion not only clarifies the foundations of earlier research but also deepens the understanding of the field. Furthermore, we review the diverse applications of these methods, demonstrating their practical value and influence across different contexts and related tasks.

In summary, our contributions are:
\begin{itemize}
\item We introduce a well-structured taxonomy of controllable generation methods from the condition perspective, encompassing novel condition introduction, multi-condition integration, and universal controllable generation. This taxonomy provides a clearer lens to reveal their core theoretical principles as well as the inherent challenges of each category. 
\item We summarize two fundamental paradigms for incorporating novel conditions into T2I diffusion models: conditional score prediction and condition-guided score estimation. Based on these paradigms, we systematically organize and review the corresponding methods, covering a broad spectrum of controllable generation studies. We carefully highlight the key features, distinctive contributions, and comparative advantages of each method. 
\item We further showcase the diverse applications of conditional generation with T2I diffusion models across a variety of generative tasks, illustrating its emergence as a central and influential component in the AIGC era. 
\end{itemize}

The rest of this paper is organized as follows. 
Sec.~\ref{sec:preliminaries} provides a brief introduction to the theory and widely used text-to-image diffusion models, overviews the popular controllable generation tasks, and presents a well-structured taxonomy.
Later, we summarize existing approaches for controlling the text-to-image diffusion model according to our proposed taxonomy, including novel condition introduction (Sec.~\ref{sec:method}), multi-condition integration (Sec.~\ref{sec:multi-conditon}), and universal controllable generation (Sec.~\ref{sec:universal}). 
Finally, Sec.~\ref{sec:application} demonstrates the applications of controllable text-to-image generation.

\section{Preliminaries}
\label{sec:preliminaries}
\subsection{Denoising Diffusion Probabilistic Models}

Denoising Diffusion Probabilistic Models (DDPMs) represent a novel class of generative models that operate on the principle of reverse diffusion. These models are formulated as parameterized Markov chains that synthesize images by gradually converting noise into structured data through a sequence of steps.

\myparagraph{Forward Process} The diffusion process begins with the data distribution $ x_0 \sim q(x_0) $ and adds Gaussian noise incrementally over $T$ timesteps. At each step $t$, the data $x_t$ is noised by a transition kernel:
\begin{equation}
q(x_{1:T}|x_0) := \prod_{t=1}^{T} q(x_t|x_{t-1}),
\end{equation}

\begin{equation}
q(x_t|x_{t-1}) = \mathcal{N}(x_t; \sqrt{1-\beta_t}x_{t-1}, \beta_t\mathbf{I}),
\end{equation}
where \( \beta_t \) are variance hyperparameters of the noise.

\myparagraph{Reverse Process}
During the reverse process of a DDPM, the model's objective is to progressively denoise the data, thereby approximating the reverse of the Markov chain. This process begins from the noise vector $x_T$ and transitions towards the original data distribution $q(x_0)$. The generative model parameterizes the reverse transition $p_{\theta}(x_{t-1}|x_t)$ as a normal distribution:
\begin{equation}
\label{eq:ddpm_reverse}
p_{\theta}(x_{t-1}|x_t) = \mathcal{N}(x_{t-1}; \mu_{\theta}(x_t, t), \Sigma_{\theta}(x_t, t))
\end{equation}
where deep neural networks, often instantiated by architectures like UNet, parameterize the mean $\mu_{\theta}(x_t, t)$ and variance $\Sigma_{\theta}(x_t, t)$. The UNet takes the noisy data $x_t$ and time step $t$ as inputs and outputs the parameters of the normal distribution, thereby predicting the noise $\epsilon_{\theta}$ that the model needs to reverse the diffusion process. To synthesize new data instances $x_0 $, we initiate by sampling a noise vector $x_T \sim p(x_T)$ and then successively sample from the learned transition kernels $x_{t-1} \sim p_{\theta}(x_{t-1}|x_t)$ until we reach $t = 1$, completing the reverse diffusion process.

\begin{revrange}
Subsequent research has extended the DDPM framework toward more general and theoretically grounded formulations, including the Denoising Diffusion Implicit Model (DDIM)~\cite{song2021denoising}, score-based generative modeling~\cite{song2021scorebased}, Flow Matching~\cite{lipman2023flow}, and Elucidated Diffusion Models (EDM)~\cite{karras2022elucidating}. Among these, we focus on the flow-matching technique, as it underpins many state-of-the-art text-to-image systems such as Stable Diffusion 3~\cite{esser2024scaling} and Flux~\cite{flux2024}.

\subsection{Flow Matching and Rectified Flow}
\label{sec:flow-matching}

\myparagraph{Flow Matching (FM)~\cite{lipman2023flow}}
Flow Matching provides a deterministic reformulation of diffusion-based generative modeling by directly learning a continuous vector field that transports probability mass from the data distribution to a simple prior. 
Let $x_t$ denote the data state at time $t \in [0,1]$ evolving under
\begin{equation}
\frac{dx_t}{dt} = u   _t(x_t), \quad x_0 \sim p_{\text{data}}, \quad x_1 \sim p_1,
\end{equation}
where $v_t(x)$ represents the velocity field. 
To enable supervised training, a conditional Gaussian path $p_t(x\mid x_0)$ is defined with analytical velocity
\begin{equation}
u_t(x_t\mid x_0) = \frac{\dot{\sigma}_t}{\sigma_t}\big(x_t - \mu_t\big) + \dot{\mu}_t,
\end{equation}
parameterized by time-dependent schedules $\mu_t$ and $\sigma_t$ (which may depend on $x_0$). 
A neural network $u_\theta(x_t,t)$ is trained to approximate this target velocity via the \emph{flow matching loss}:
\begin{equation}
\mathcal{L}_{\text{FM}} = \mathbb{E}_{x_0,\epsilon,t}\!\left[\|u_\theta(x_t,t)-u_t(x_t|x_0)\|^2\right],
\end{equation}
where $x_t=\mu_t(x_0)+\sigma_t\epsilon$.  Here, $\mu_t(x_0)$ denotes a deterministic interpolation function, distinct from the learnable mean $\mu_\theta(x_t,t)$ used in DDPMs.
After training, generation proceeds deterministically by integrating the learned ODE from noise to data, removing stochasticity and improving sampling efficiency compared with DDPMs.

\myparagraph{Rectified Flow (RF)~\cite{liu2023flow}}
Rectified Flow simplifies the FM formulation by adopting a straight-line interpolation between the prior and the data:
\begin{equation}
x_t = (1-t)x_1 + t x_0, \qquad u_t(x_t\mid x_0) = x_0 - x_1,
\end{equation}
with $x_0 \sim p_{\text{data}}$, $x_1 \sim p_1$, and $t \sim \mathcal{U}(0,1)$.
The training objective remains a mean-squared error on the predicted velocity:
\begin{equation}
\mathcal{L}_{\mathrm{RF}} =
\mathbb{E}_{x_0,x_1,t}\!\left[\|\,u_\theta(x_t,t) - (x_0 - x_1)\|^2\right].
\end{equation}
This rectification enforces a monotonic and low-curvature probability flow, resulting in smoother vector fields and faster ODE integration. 

\end{revrange}

\begin{table*}[t]
\centering
\caption{\textbf{Collection of primary and used text-to-image diffusion models in this survey.} $^\dagger$: number of UNet and text encoder's parameters (default refers only to UNet). $f$: downsampling factor of autoencoder in latent-space diffusion models. \underline{CLIP}: open source implementation of CLIP. $^*$: train from scratch. 
\textbf{Resolution}: the maximum supported image resolution generated by the model.
}
\label{tab:t2idm}
\resizebox{\textwidth}{!}
{
\renewcommand{\arraystretch}{1.3}
\setlength{\tabcolsep}{3pt}
\begin{tabular}{l|c||c|c|>{\centering\arraybackslash}p{0.5cm}|c|c|c|c}
\hline\thickhline
\rowcolor{mygray1}
\textbf{Model}  & \textbf{Pub.}  & \textbf{Param.}   & \textbf{Resolution}&\textbf{$f$}  &\textbf{Text Encoder}&\textbf{Arch.} & \textbf{Training Dataset} & \textbf{Open} \\ 
\hline\hline
\multicolumn{7}{l}{\emph{\textbf{Pixel Space Diffusion Models}}}\\
\hline
GLIDE\cite{nichol2021glide} &  ICML 2022& 5.0B$^\dagger$  & $256^2$ &- & plain Transformer$^*$\cite{vaswani2017attention} & U-Net& DALL$\cdot$E\cite{ramesh2021zero} & \ding{51} \\
Imagen\cite{saharia2022photorealistic} &  NeurIPS 2022 & 3.0B&  $1024^2$ &-& T5-XXL\cite{raffel2020exploring}& U-Net & $>$LAION-400M\cite{schuhmann2021laion}  & \ding{55} \\
DALL·E\ 2/3\cite{Ramesh2022HierarchicalTI}  & arXiv 2022 & 4.5B & 1024$^2$ & - & CLIP$^*$\cite{radford2021learning} \& Diffusion prior$^*$ & U-Net& CLIP\cite{radford2021learning} \& DALL$\cdot$E\cite{ramesh2021zero} &\ding{55} \\
DeepFloyd IF~\cite{stabilityai2023deepfloydif} &-& 1.5B$\sim$6.2B  & $1024^2$  &-& T5-XXL\cite{raffel2020exploring} & U-Net&  $>$LAION-A 1B\cite{schuhmann2021laion}   & \ding{51} \\
\hline 
\multicolumn{7}{l}{\emph{\textbf{Latent Space Diffusion Models}}}\\
\hline
LDM\cite{rombach2022high}& CVPR 2022& 903M  & $256^2$ &$8$&BERT-tokenizer\cite{devlin2018bert} & U-Net& LAION-400M\cite{schuhmann2021laion}  &\ding{51} \\
SD 1.x\cite{rombach2022high}& CVPR 2022 & 860M & $512^2$ & 8 & CLIP-L\cite{radford2021learning}& U-Net& LAION-2B\cite{schuhmann2022laion} & \ding{51} \\
SD 2.x\cite{rombach2022high}& CVPR 2022 & 865M  & $512^2$/$768^2$ & 8 & \underline{CLIP}-H/14\cite{radford2021learning}& U-Net& LAION-5B\cite{schuhmann2022laion}  & \ding{51}  \\ 
SD XL\cite{podell2023sdxl}& ICLR 2024& 2.6B  & $1024^2$  &8& \underline{CLIP}-G \& \underline{CLIP}-L\cite{radford2021learning} & U-Net& internal dataset   & \ding{51} \\
SD 3.x\cite{esser2024scaling}& ICML 2024 & 2B$\sim$8.1B  & $2048^2$  &8& \underline{CLIP}-G\&L\cite{radford2021learning} \& T5-XXL\cite{raffel2020exploring}& MM-DiT & internal dataset   & \ding{51} \\
PixArt-$\alpha$\cite{chen2023pixart}& ICLR 2024& 0.25B  & $1024^2$  &8& T5-XXL\cite{raffel2020exploring} & DiT& mixture & \ding{51} \\

Flux.1x~\cite{flux2024}& arXiv 2025& 12B  & $2048^2$  &8& \underline{CLIP}-L\cite{radford2021learning} \& T5-XXL\cite{raffel2020exploring}& MM-DiT & internal dataset  & \ding{51} \\
\bottomrule
\end{tabular}
}
\vspace{-3mm}
\end{table*}

\subsection{Text-to-Image Diffusion Models}

Text-to-image (T2I) diffusion models synthesize images from textual descriptions by learning a conditional denoising process. 
A central design question in such models is how textual information is injected into the denoising network. We highlight here the representative architectural paradigms that define modern T2I systems regarding task formulation and architecture-level analysis (Tab.~\ref{tab:t2idm}).

\myparagraph{Cross-Attention–Based U-Net Architectures} 
Most early and widely adopted T2I models employ a U-Net backbone augmented with cross-attention layers to model. 
In this mechanism, latent image features act as queries, while text embeddings provide keys and values, enabling fine-grained alignment between linguistic concepts and visual features. 
GLIDE~\cite{nichol2021glide} first demonstrated that replacing class-conditioning in diffusion models with free-form text, combined with classifier-free guidance (CFG)~\cite{ho2022classifier}, yields strong improvements in photorealism and text alignment. 
Imagen~\cite{saharia2022photorealistic} further showed that scaling the text encoder (a frozen large language model, e.g., T5~\cite{raffel2020exploring}) improves generation quality more than enlarging the diffusion model itself, and confirmed cross-attention as the most effective conditioning approach.  
Latent Diffusion Models (LDM)~\cite{rombach2022high} introduced a major efficiency breakthrough by performing diffusion in a compressed latent space, enabling high-resolution synthesis on limited compute. 
Stable Diffusion (SD)~\cite{rombach2022high} builds on the LDM formulation and its v1.x, v2.x, and SDXL variants adopt the U-Net with cross-attention design that has become the de facto standard in open-source T2I generation.

\myparagraph{Transformer-Based Diffusion Models: DiT and MMDiT}
Recent T2I systems, such as Stable Diffusion 3~\cite{esser2024scaling} and FLUX~\cite{flux2024}, increasingly replace U-Nets with transformer-based diffusion backbones due to their improved scalability and expressiveness. 
Diffusion Transformers (DiT)~\cite{peebles2023scalable} treat image latents as patch tokens and perform denoising entirely through self-attention, showing superior scaling behavior compared to convolutional models. 
Beyond traditional cross-attention, Multimodal Diffusion Transformers (MMDiT) introduce \emph{joint attention} mechanisms in which text and image tokens interact bidirectionally within the same transformer blocks. 
This unified multimodal modeling enables richer text–image dependency structures than unidirectional cross-attention.

\subsection{Controllable Generation Tasks}
\label{sec:tasks}
\begin{figure*}[t!]
	\begin{center}
		\includegraphics[width=1.0\linewidth]{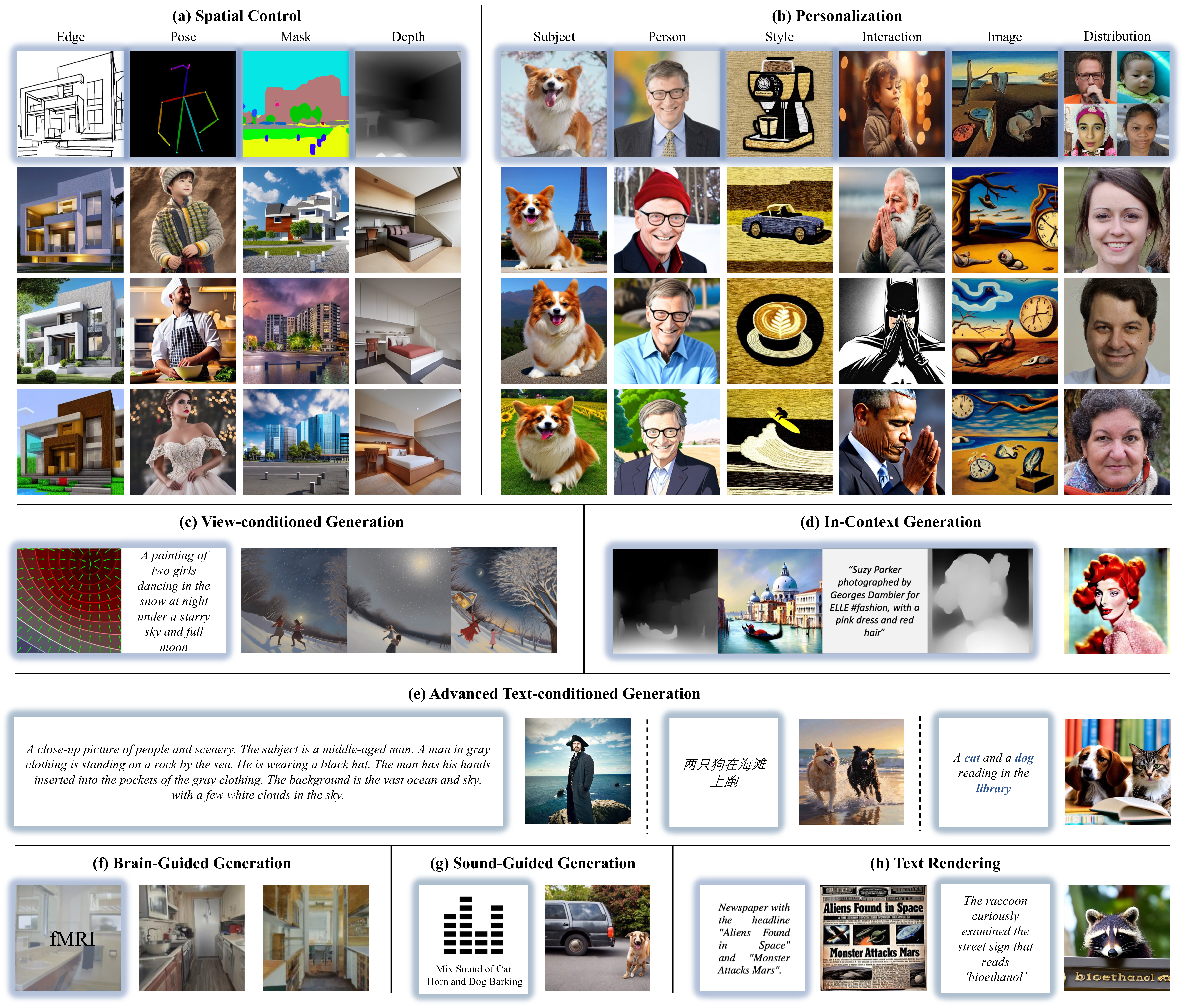}
	\end{center}
\vspace{-3mm}
	\captionsetup{font=small}
	\caption{\small{
\textbf{Illustration of controllable text-to-image generation with specific conditions.} The condition is marked in blue background. Examples are sourced from \cite{wei2023elite,chen2023photoverse,chen2023artadapter,huang2023learning,ramesh2022hierarchical,cao2025image,zhang2023adding,wang2023context,wu2023paragraph,lu2023minddiffuser,qin2023gluegen,zhang2023brush,bernal2025precisecam,tan2024empirical}.}}
\vspace{-3mm}
\label{fig:personalized}
\end{figure*}

Controllable generation extends the capabilities of text-to-image diffusion models by introducing diverse conditions that guide the synthesis process beyond plain textual prompts. These conditions enable more precise alignment with user intentions. In this part, we introduce the major categories of controllable generation tasks and illustrate them in Fig.~\ref{fig:personalized}:

\myparagraph{Spatial Control} 
Since text alone struggles to represent structural information such as position and dense labels, spatial signals have become crucial conditions for text-to-image diffusion. Typical spatial inputs include layout, human pose, depth, and segmentation masks.

\myparagraph{Image Personalization} 
The personalization task aims to capture and utilize concepts from exemplar images that cannot be easily described by text, integrating them as generative conditions for controllable synthesis.

\myparagraph{View-conditioend Generation}
View-conditioned generation aims to synthesize images from specific viewpoints or across multiple views, ensuring geometric consistency and structural coherence. By leveraging conditions such as camera parameters, depth maps, or multi-view correspondences, these methods extend controllable diffusion to scenarios like novel view synthesis, 3D-aware image generation, and panoramic rendering.

\myparagraph{Advanced Text-Conditioned Generation}
Although text is the fundamental condition in text-to-image diffusion models, several challenges remain. For example, text-guided synthesis often suffers from misalignment, particularly when dealing with complex prompts involving multiple entities or rich contextual descriptions. Moreover, the dominance of English datasets in training leads to limited multilingual capabilities. To address these issues, recent works propose novel strategies for improving alignment and expanding the linguistic scope of text-conditioned generation.

\myparagraph{In-Context Generation}
In-context generation focuses on understanding and executing specific tasks on query images by leveraging task-specific example pairs of images and text guidance. This setting enables models to adapt to new tasks with minimal supervision, expanding the flexibility of controllable diffusion.

\myparagraph{Brain- \& Sound-guided Generation}
Brain-guided generation seeks to control image synthesis directly from neural activity, such as electroencephalogram (EEG) recordings or functional magnetic resonance imaging (fMRI), bypassing the intermediate step of translating thoughts into text. Similarly, sound-guided generation explores how auditory signals can be leveraged as direct conditions for visual synthesis, enabling cross-modal creativity.

\myparagraph{Text Rendering}
Rendering coherent and legible text within generated images is a critical task, given its wide applications in posters, book covers, advertisements, and memes. Effective text rendering not only enhances practicality but also pushes the boundary of fine-grained controllability in diffusion models.

\begin{table}[t]
\centering
\caption{Summary of evaluation dimensions and commonly used metrics for controllable T2I models.}
\label{tab:evaluation_metrics}

\begin{tabular}{p{0.2\linewidth} p{0.6\linewidth}}
\toprule
\textbf{Dimension} & \textbf{Metrics} \\
\midrule

Image Quality &
FID~\cite{heusel2017gans}; 
IS~\cite{salimans2016improved}; 
Aesthetic Score\cite{schuhmann2022laionaesthetics};Human Preference(HPS~\cite{wu2023human},PickScore~\cite{kirstain2023pick})
\\

\midrule

Conditional Alignment &
Textual Similarity(CLIP~\cite{hessel2021clipscore,radford2021learning}); Visual Similarity(CLIPScore\cite{radford2021learning}, DINO~\cite{oquab2023dinov2}); Spatial Consistency(mIoU, MSE); Content Accuracy\cite{huang2023t2i,ghosh2023geneval}
\\

\bottomrule
\end{tabular}

\vspace{-3mm}
\end{table}

\subsection{Evaluation of Controllable T2I Generation}
Since the types of conditions used in controllable T2I methods are highly diverse and vary across approaches, we focus here on the overarching evaluation dimensions rather than specific task settings. In what follows, we outline the main perspectives from which controllable T2I models are commonly assessed and summarize the widely adopted metrics associated with each dimension (Tab.~\ref{tab:evaluation_metrics}).

\myparagraph{Image Quality}
Image quality is commonly measured by distribution-based metrics such as Inception Score (IS)~\cite{salimans2016improved} and Fréchet Inception Distance (FID)~\cite{heusel2017gans}, computed on deep features from a pretrained classifier (e.g., Inception~V3~\cite{szegedy2016rethinking}). In addition, aesthetic predictors~\cite{schuhmann2022laionaesthetics} and learned human-preference models (HPS~\cite{wu2023human}, PickScore~\cite{kirstain2023pick}) estimate visual appeal and perceived quality directly from generated images.

\myparagraph{Conditional Alignment}
Conditional alignment measures how faithfully the generated images follow the specified conditions (textual or otherwise). Text–image alignment is typically quantified with CLIP-based similarity scores~\cite{hessel2021clipscore,radford2021learning}, which leverage a large-scale contrastively trained vision–language model to compare prompts and outputs. For other types of conditions, alignment is computed as the discrepancy between the input condition and its prediction from the generated image, for example by using CLIP or DINOv2~\cite{oquab2023dinov2} features for image-conditioned tasks, or detection metrics such as mAP for layout-conditioned generation.

\subsection{Taxonomy}
Conditional generation with text-to-image diffusion models is inherently complex and can be broadly organized into three sub-tasks from the perspective of conditioning signals.
The first and most studied direction augments pretrained diffusion models with novel conditions. We group these methods by their theoretical foundations, namely conditional score prediction and condition-guided score estimation. The main challenge is to flexibly inject new condition types alongside text prompts without sacrificing image quality.
The second direction targets multi-condition control, such as combining a character’s identity with a specific pose. Methods are categorized by their technical strategies, including joint training, continual learning, weight fusion, attention-based integration, and guidance composition. Here, the core difficulty is to fuse multiple signals so that all conditions are faithfully expressed.
The third direction pursues condition-agnostic generation, aiming to build unified frameworks that can robustly leverage diverse condition types across a wide range of inputs.

\begin{revrange1}

\section{Control Text-to-Image Diffusion Models with Novel Conditions}
\label{sec:method}
Following \cite{luo2022understanding}, we can set the approximate denoising transition mean $\mu_{\theta}(x_t, t)$ in Eq.~\ref{eq:ddpm_reverse} as:
\begin{equation}
\label{eq:mu}
\mu_{\theta}(x_t, t) = \frac{1}{\sqrt{\alpha_t}} x_t - \frac{1 - \alpha_t}{\sqrt{\bar{\alpha}_t}} s_{\theta}(x_t, t) 
\end{equation}
where $s_{\theta}(x_t, t)$ is a neural network that learns to predict the score function $\nabla_{x_{t}}\log p_t(x)$.
Hence, we have:
\begin{equation}
\nabla_{x_{t}}\log p_t(x)=-\frac{1}{\sqrt{1-\bar{\alpha}_t}}\epsilon
\end{equation}
where $\epsilon \sim \mathcal{N}(0, \mathbf{I})$ is the Gaussian noise used in forward process, $\alpha_t:=1-\beta_t$, and $\bar\alpha_t:=\prod_{s=0}^{t}\alpha_s$.
Then, Eq.~\ref{eq:mu} can be written as:
\begin{equation}
\label{eq:mu_epsilon}
\mu_{\theta}(x_t, t) = \frac{1}{\sqrt{\alpha_t}} \left(x_t - \frac{1 - \alpha_t}{\sqrt{1 - \bar{\alpha}_t}} \hat\epsilon(x_t, t) \right)
\end{equation}
where $\hat\epsilon(x_t, t)$ predicts $\epsilon$.

In conditional generation ($c$ denotes condition), the score function is extended with a posterior probability term $\nabla_{x_{t}}\log p_t(c|x_{t})$ and becomes $\nabla_{x_{t}}\log \left (p_t(x_{t})p_t^w(x_{t}|c) \right)$ ($w$ represents a hyper-parameter to control condition intensity), following \cite{dhariwal2021diffusion,ho2022classifier}. 
To employ a neural network for conditional generation, classifier-free guidance (CFG)\cite{ho2022classifier}  transforms it to:
\begin{align}
    \nabla_{x_{t}}\log &\left (p_t(x_{t})p_t^w(c|x_{t}) \right) \nonumber\\
    &= \nabla_{x_{t}}\log p_t(x_{t})+w\nabla_{x_{t}}\log p_t(c|x_{t}) \nonumber \\
    & = \nabla_{x_{t}}\log p_t(x_{t}) + w\nabla_{x_{t}}\log \frac{p_t(x_{t}|c)}{p_t(x_{t})} \nonumber \\
    & = (1-w)\nabla_{x_{t}}\log p_t(x_{t}) + w\nabla_{x_{t}}\log p_t(x_{t}|c)
\end{align}
where $\nabla_{x_{t}}\log p_t(x_{t})$ and $\nabla_{x_{t}}\log p_t(x_{t}|c)$ can be predicted by training a model $\epsilon_{\theta}(x_t, \cdot, t)$, which predict the former via $\epsilon_{\theta}(x_t, \phi, t)$ and the latter via $\epsilon_{\theta}(x_t, c, t)$.

Existing T2I diffusion models train $\epsilon_{\theta}(x_t, \cdot, t)$ by randomly dropping the text prompt, and the denoising process with CFG is as follows:
\begin{align}
\label{eq:cfg}
    \hat\epsilon(x_t, c_{text}, t)=(1-w)\epsilon_{\theta}(x_t, \phi, t)+w\epsilon_{\theta}(x_t, c_{text}, t)
\end{align}
and $\hat\epsilon(x_t, c_{text}, t)$ is used in Eq.~\ref{eq:mu} for conditional synthesis.

Hence, the key to controlling text-to-image models with novel conditions $c_{novel}$ is to model score $\nabla_{x_{t}}\log p_t(x_{t}|c_{text},c_{novel})$.
Following \cite{dhariwal2021diffusion,zhang2023text}, there are two types of mechanisms, \ie, conditional score prediction and conditioned-guided score estimation, which we illustrate below.

\myparagraph{Conditional Score Prediction (Sec.~\ref{sec:method_csp})} While T2I diffusion models leverage $\epsilon_{\theta}(x_t, c_{text}, t)$ to predict $\nabla_{x_{t}}\log p_t(x_{t}|c_{text})$, a fundamental and powerful way for steering diffusion models is through conditional score prediction in the sampling process, where these methods introduce $c_{novel}$ into $\epsilon_{\theta}(x_t, c_{text}, t)$, constructing a $\tilde\epsilon(x_t, c_{text}, c_{novel}, t)$ to straightforwardly predict $\nabla_{x_{t}}\log p_t(x_{t}|c_{text},c_{novel})$. The CFG-based denoising update then reads:
\begin{align}
    \label{eq:cfg_cond}
    \hat\epsilon(x_t, c_{text}, c_{novel}, t)=(1-w)&\tilde\epsilon(x_t, \phi, t) \nonumber\\
    +w&\tilde\epsilon(x_t, c_{text}, c_{novel}, t)
\end{align}
We here illustrate several mainstream ways to attain $\tilde\epsilon(x_t, c_{text}, c_{novel}, t)$.

\myparagraph{Condition-guided Score Estimation (Sec.~\ref{sec:method_cse})}
Unlike conditional score prediction straightforwardly predicting $\nabla_{x_{t}}\log p_t(x_{t}|c_{text},c_{novel})$, condition-guided estimation approaches estimate $\log p_t(c_{novel}|x_t)$ with a likelihood/critic and obtain $\nabla_{x_{t}}\log p_t(c_{novel}|x_t)$ by backpropagation, which is then injected into the sampler. And the denoising process now reads:
\begin{align}
\label{eq:condition_guided}
\hat\epsilon(x_t, c_{text}, c_{novel}, t)=&\hat\epsilon(x_t, c_{text}, t) \nonumber\\
&+ \gamma \nabla_{x_{t}} \log p_{t}(c_{novel}|x_t)
\end{align}
where $\gamma$ is a hyper-parameter to adjust the conditional score and $\hat\epsilon(x_t, c_{text}, t)$ is the original score prediction of text-conditioned diffusion models with CFG. 

\subsection{Conditional Score Prediction}
\label{sec:method_csp}

Conditional score prediction approaches focus on empowering pre-trained denoising models supporting novel conditions to straightforwardly predict denoised latents. According to mechanisms, these methods can be categorized into tuning-based (Sec.~\ref{sec:method_csp_tune}), adapter-based (Sec.~\ref{sec:method_csp_model}), and training-free  (Sec.~\ref{sec:method_csp_free}) manners.

\begin{figure}[t!]
\begin{center}
    \includegraphics[width=1.0\linewidth]{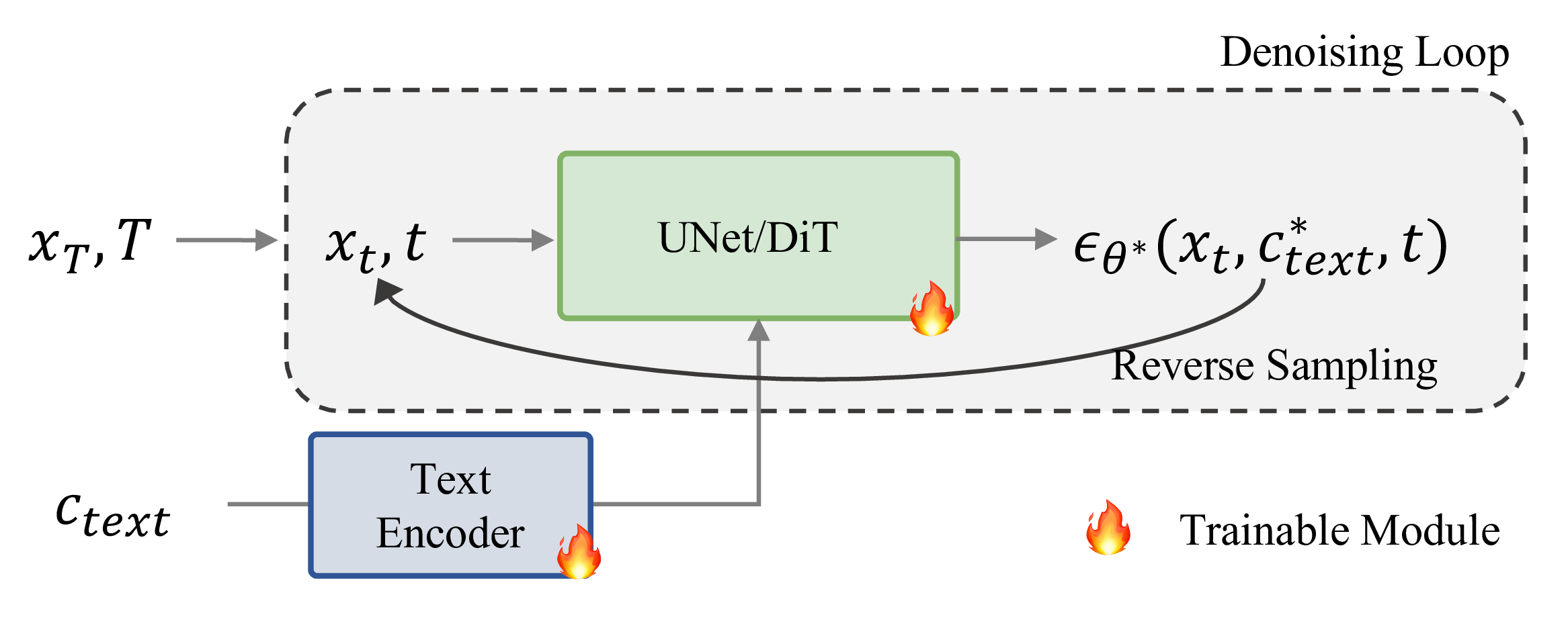}
\end{center}
\vspace{-4mm}
\captionsetup{font=small}
\caption{\small{
\textbf{Illustration of tuning-based conditional score prediction.}}}
\vspace{-4mm}
\label{fig:method_csp_tune}
\end{figure}

\subsubsection{Tuning-based Conditional Score Prediction}
\label{sec:method_csp_tune}
Tuning-based methods typically focus on adapting to a specific condition, often in scenarios with limited data, such as single or few-shot examples. These methods achieve conditional prediction by transforming either the text condition $c_{text}$ or the model parameters $\theta$ into a form specific to the given condition, as shown in Fig.~\ref{fig:method_csp_tune}. This can be represented as:
\begin{align}
\tilde\epsilon(x_t, c_{text}, c_{novel}, t)&=\epsilon_{\theta^*}(x_t, c^*_{text}, t)
\end{align}
where condition information is memorized in $c_{text}$ and $\theta$.

\myparagraph{Basic Tuning-based Methods} 
A straightforward yet effective strategy for learning controlling concepts from given samples is to fine-tune the diffusion models, thereby adapting text-to-image models to reflect the target conditions. 
The updated parameters are specialized to capture the desired conditions~\cite{dong2022dreamartist,gal2022image,ruiz2023dreambooth,voynov2023p+,zhang2023prospect,roy2023diffnat,zhao2023videoassembler,chatterjee2024getting}.

As the fundamental input of text-to-image diffusion models, text plays a central role in adapting these models to user-specific requirements. 
Textual Inversion (TI)\cite{gal2022image} introduces an innovative idea by embedding user-provided concepts into new ``words'' in the text embedding space. This expands the tokenizer’s dictionary and optimizes additional tokens through a denoising process applied on provided images. 
DreamBooth\cite{ruiz2023dreambooth} adopts a similar approach but encodes concepts with low-frequency words (\ie, \emph{sks}), and additionally updates the UNet parameters with a class-specific prior preservation loss to enhance output diversity. 
Together, the simplicity and adaptability of TI and DreamBooth have established them as foundational frameworks for many subsequent tuning-based methods.
Building on these foundations, Custom Diffusion~\cite{kumari2023multi} examines weight deviations during fine-tuning and identifies cross-attention layer parameters—particularly the key and value projections ($W^k$ and $W^v$)—as pivotal. It thus narrows updates to these projections and augments the process with extra text tokens and a regularization loss. 

To further improve textual inversion, recent works explore layer-specific distinctions in UNet~\cite{voynov2023p+,zhang2023prospect,jin2024image}, refine embedding initialization strategies~\cite{pang2024cross}, and sampling distribution~\cite{zhao2025dreamdistribution}. These methods apply distinct text embeddings across layers to capture finer variations. In contrast, CatVersion \cite{zhao2025catversion} moves away from tuning text embeddings or UNet parameters, and instead learns concatenated embeddings within the dense feature space of the text encoder. This design proves effective for capturing subtle differences between a personalized concept and its base class, thereby helping preserve prior knowledge.

\myparagraph{Parameter-Efficient Fine-Tuning (PEFT)} 
Beyond full fine-tuning, parameter-efficient methods (PEFT)\cite{houlsby2019parameter,hu2021lora, valipour2022dylora,chavan2023one} have become increasingly important in personalization~\cite{xiao2023comcat}. 
Among them, Low-rank Adaptation (LoRA)\cite{hu2021lora} has been widely adopted in various personalization pipelines~\cite{ruiz2023dreambooth,chen2023disenbooth,gu2023mix,smith2023continual,guo2023animatediff,marjit2024diffusekrona}. 
Xiang \etal propose ANOVA~\cite{xiang2023closer}, which employs adapters~\cite{houlsby2019parameter} and demonstrates that placing them after the cross-attention block notably boosts performance. 
COMCAT~\cite{xiao2023comcat} develops an efficient ViT~\cite{dosovitskiy2020image} compression method based on model factorization. 
Similarly, DiffuseKronA~\cite{marjit2024diffusekrona} introduces a Kronecker product-based adaptation module that surpasses LoRA-DreamBooth in parameter efficiency and stability, offering consistent high-quality generation with greater interpretability. 
To support research and evaluation in this area, LyCORIS~\cite{yeh2023navigating} provides a comprehensive open-source library\footnote{https://github.com/KohakuBlueleaf/LyCORIS} covering numerous PEFT methods (\eg, LoRA, LoHa, DyLoRA~\cite{valipour2022dylora}) and a framework for their systematic assessment, thus promoting the progress of diffusion model personalization.

\myparagraph{Condition Disentanglement}
A further challenge in introducing novel conditions lies in disentangling the desired concept from confounding inputs. 
Many studies~\cite{chen2025consislora,zhang2023motioncrafter,guo2023animatediff,sohn2023styledrop,avrahami2023break,chen2023disenbooth,cai2024decoupled,li2023generate,motamed2023lego,jones2024customizing,zhang2024attention,zhu2024isolated,zhang2024compositional,hyung2024magicapture,safaee2024clic,jang2024identity} observe that irrelevant information—such as background context or co-occurring objects—tends to be entangled with the target concept. 
To address this, several works~\cite{avrahami2023break,jin2023image,safaee2023clic} employ explicit masks to isolate object regions. 
In this direction, Disenbooth~\cite{chen2023disenbooth} and DETEX~\cite{cai2024decoupled} reduce the impact of backgrounds, with DETEX further decoupling pose from subject appearance. 
PACGen~\cite{li2023generate} instead applies aggressive data augmentation, altering the size and position of concepts to help separate spatial cues from the core identity.

Other approaches disentangle conditions by adjusting fine-tuning strategies. 
DreamVideo~\cite{wei2024dreamvideo} separates subject learning and motion learning, while B-LoRA~\cite{frenkel2024implicit} jointly optimizes LoRA weights of two blocks to implicitly decouple style and content. 
ReVersion~\cite{huang2024reversion} introduces a relation-steering contrastive learning scheme to capture object relationships more effectively.  

Training techniques can also be exploited to reduce condition entanglement~\cite{kim2024selectively,huang2024learning,wu2024u}. 
Selective Information Description~\cite{kim2024selectively} leverages a VLM to produce refined text descriptions, ensuring the model emphasizes target objects over contextual biases. 
Similarly, U-VAP~\cite{wu2024u} adopts a decoupled self-augmentation strategy, where an LLM generates paired target and non-target prompts, facilitating dual concept learning to decouple conditions.

\myparagraph{Prior Preservation}
Another important challenge in fine-tuning diffusion models is the preservation of prior knowledge. Without careful design, models risk overfitting to narrow concepts, sacrificing generality and controllability. 
To address this, prior preservation techniques have been widely explored~\cite{ruiz2023dreambooth,kumari2023multi,tewel2023key,qiu2023controlling,han2023svdiff,wang2023hifi,zeng2024infusion,qiao2024facechain}. 
Perfusion~\cite{tewel2023key} mitigates overfitting by locking cross-attention keys to prior categories and applying a gated rank-1 concept update. 
SVDiff~\cite{han2023svdiff} regulates singular values in weight matrices to reduce risks such as language drift, while OFT~\cite{qiu2023controlling} introduces orthogonal fine-tuning to preserve semantic capacity by maintaining hyperspherical energy. Together, these methods help balance fidelity to new data with the retention of broad generative ability.

\myparagraph{Training Techniques}  
Beyond the strategies above, alternative training techniques have been explored to improve efficiency, reduce computational overhead, and enhance performance~\cite{voronov2023loss,fei2023gradient,agarwal2023image,yang2023controllable,he2023data,zhao2025catversion,zhang2024generative}. 
For example, DVAR~\cite{voronov2023loss} proposes a variance-based early stopping criterion to replace unreliable convergence metrics, thereby accelerating training. 
Gradient-Free Textual Inversion~\cite{fei2023gradient} divides optimization into dimensionality reduction and non-convex gradient-free search, achieving faster convergence with minimal performance loss. 
MATTE~\cite{agarwal2023image} investigates the roles of timesteps and UNet layers for different concept categories, aiming for broader adaptability.

Another notable direction is the introduction of auxiliary losses to boost generation performance~\cite{Wang2024TokenCompose,jiang2024comat,wang2024detdiffusion,wu2024relation}. 
TokenCompose~\cite{Wang2024TokenCompose} improves multi-object composition and photorealism through token-wise consistency losses between images and segmentation maps. 
Similarly, CoMat~\cite{jiang2024comat} addresses text-image misalignment via a concept activation loss and enhances visual quality with adversarial loss. 

Lastly, data-centric strategies have been proposed to enhance the training process. 
COTI~\cite{yang2023controllable} improves Textual Inversion through active and controllable data selection, while He \etal~\cite{he2023data} generate text- and image-level regularization datasets to better preserve model generalization.  

\myparagraph{Inference Techniques}
In addition to training methods, several approaches enhance controllability during inference by modulating cross-attention to better align outputs with target conditions~\cite{zhou2024magictailor,ham2024personalized,ng2024partcraft,zhu2025multibooth}. 
For example, MagicTailor~\cite{zhou2024magictailor} combats semantic pollution and imbalance with dynamic masked degradation and dual-stream balancing. 
OMG~\cite{kong2024omg} provides an occlusion-friendly framework that adopts a two-stage sampling process—first generating layout and visual comprehension for occlusion handling, then applying noise blending to integrate concepts—leading to superior identity preservation and visual harmony. 

Additionally, DreamBlend~\cite{ram2025dreamblend} addresses the trade-off between prompt fidelity, subject fidelity, and diversity by leveraging multiple checkpoints and combining their strengths through cross-attention guidance. 
Prompt-aligned personalization~\cite{arar2024palp} improves complex text alignment via score distillation sampling while supporting multi-/single-shot scenarios, subject composition, and reference-guided generation.  

\begin{figure}[t!]
\begin{center}
    \includegraphics[width=1.0\linewidth]{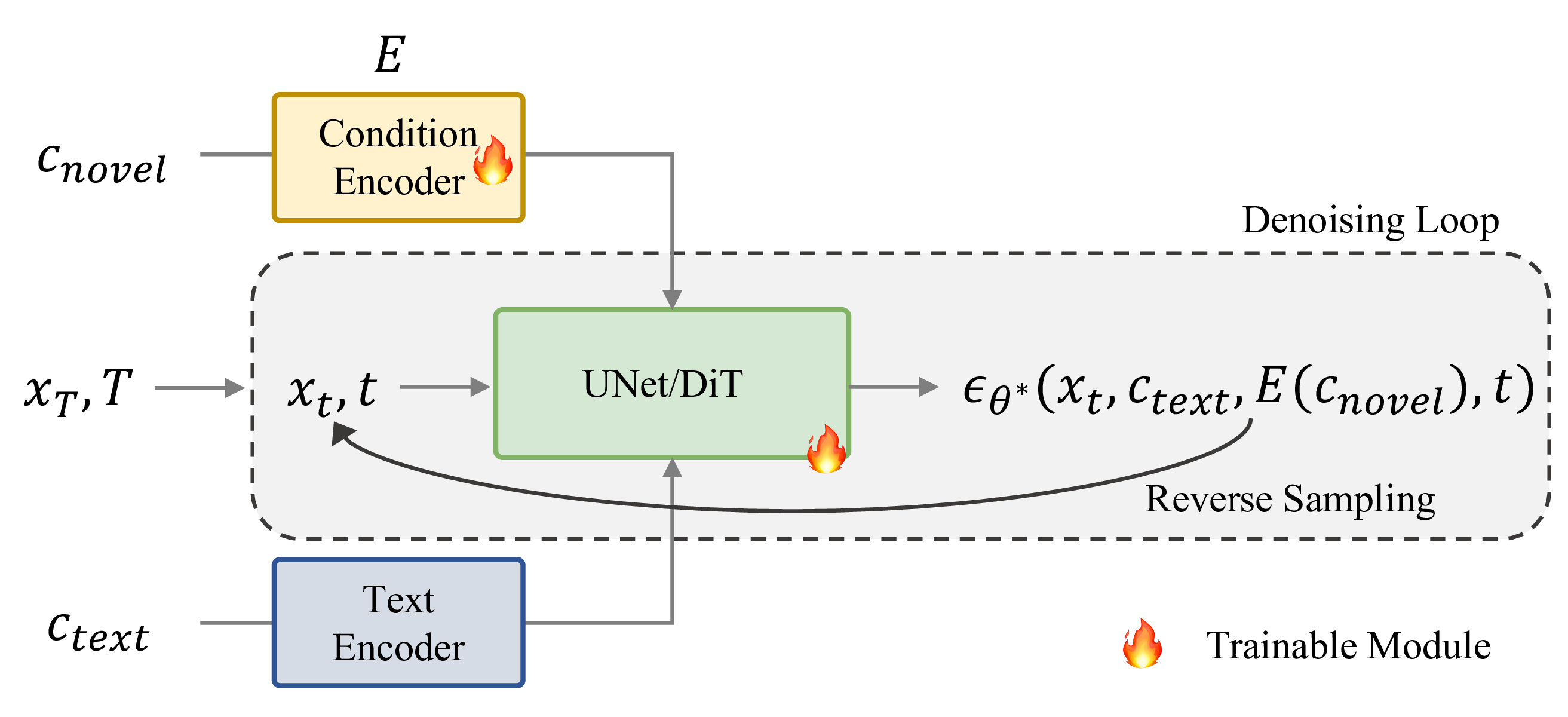}
\end{center}
\vspace{-4mm}
\captionsetup{font=small}
\caption{\small{
\textbf{Illustration of adapter-based conditional score prediction.}}}
\vspace{-5mm}
\label{fig:method_csp_model}
\end{figure}

\subsubsection{Adapter-based Conditional Score Prediction}
\label{sec:method_csp_model}

To eliminate test-time tuning cost, a class of methods introduces an additional encoder $E$ that maps novel conditions to feature embeddings and feeds them into the noise predictor. The conditional score prediction then reads
\begin{align}
\tilde\epsilon(x_t, c_{text}, c_{novel}, t)
&= \epsilon_{\theta^*}\!\big(x_t,\, c_{text},\, E(c_{novel}),\, t\big),
\end{align}
where $\theta^*$ and $E$ are learned \emph{offline} and kept fixed at inference. Fig.~\ref{fig:method_csp_model} gives a schematic overview. 

Adapter-based variants primarily differ by the condition families they target (\eg, geometry, style, layout, controls), which dictate how the introduced adapter $E(\cdot)$ interfaces with the backbone (\eg, concatenation to token embeddings, cross-attention keys/values, or feature-wise affine modulation). In what follows, we group methods by task type and provide a systematic review; task definitions are summarized in Sec.~\ref{sec:tasks}.

\myparagraph{Spatial Control}
ControlNet~\cite{zhang2023adding} stands out among generalized spatial controllers, earning recognition as a seminal work and winning the Marr Award in 2023. Distinct from approaches that simply fine-tune the base diffusion model parameters~\cite{ruiz2023hyperdreambooth,kumari2023multi}, ControlNet introduces an auxiliary encoder branch mirroring the U-Net and couples it to the original layers via zero convolutions to mitigate overfitting and catastrophic forgetting. Owing to its simplicity and adaptability, ControlNet has proved effective and has become a widely adopted baseline in subsequent studies~\cite{hu2023cocktail,jia2023ssmg,qin2023unicontrol,yang2023meta,zavadski2023controlnet,xiao2023ccm,lv2024place,wang2024instancediffusion}.
Similarly, T2I-Adapter~\cite{mou2023t2i} aligns internal knowledge in text-to-image diffusion models with external spatial control signals.

While ControlNet~\cite{zhang2023adding} requires training a separate model for each control type, subsequent work seeks general controllers that handle diverse spatial signals~\cite{qin2023unicontrol,hu2023cocktail,zhao2023uni,yang2023meta,feng2025simplifying,linctrl,zhou2024migc}. 
Qin \etal~\cite{qin2023unicontrol} propose UniControl, a task-aware HyperNetwork that modulates the diffusion backbone across condition types: conditions are encoded via a mixture-of-experts (MoE) adapter, while task instructions are embedded by the HyperNet and injected through zero-conv gating to precisely regulate how condition features enter the model. 
Meta ControlNet~\cite{yang2023meta} adapts the pretrained ControlNet to another condition domain by meta learning, apparently reducing the learning steps.
In parallel, Ctrl-Adapter~\cite{linctrl} and X-Adapter~\cite{ran2024x} transfer diverse controls to arbitrary diffusion backbones by adapting pretrained ControlNets.

For layout-conditioned generation, many methods arrange regions and bind them to textual concepts~\cite{li2023gligen,ham2023modulating,xue2023freestyle,zheng2023layoutdiffusion,avrahami2023spatext,jia2023ssmg,wang2023enhancing,qi2023layered,tian2023interactdiffusion,chen2023integrating,voynov2023anylens}. 
GLIGEN~\cite{li2023gligen} grounds language with structured inputs and injects the grounding via gated trainable layers, enabling controllable placement. SpaText~\cite{avrahami2023spatext} builds a spatio-textual representation by stacking CLIP-derived object embeddings in masks at their target positions to enforce layout. 
Related efforts focus on the face domain, synthesizing faces under face-parsing constraints~\cite{cheng2023layoutdiffuse,giambi2023conditioning}.

From the objective perspective, several approaches jointly denoise dense spatial conditions to improve alignment. 
JointNet~\cite{zhang2023jointnet} augments a pretrained T2I backbone with a dense-modality branch (\eg, depth) that is tightly coupled to the RGB branch, enabling rich cross-modality interactions. 
Liu \etal~\cite{liu2023hyperhuman} propose a Latent Structural Diffusion Model that co-denoises depth and surface normals alongside RGB synthesis. 
Complementarily, adding auxiliary spatial-consistency losses further enhances control~\cite{li2024controlnet++,koley2024s,li2024adversarial}.
Specifically, ControlNet++~\cite{li2024controlnet++} enforces pixel-level cycle consistency between outputs and controls, while Li \etal~\cite{li2024adversarial} introduce a segmentation-based discriminator to provide explicit spatial feedback.

While many aim for strict adherence to provided controls, other methods support coarse or incomplete spatial inputs~\cite{zhao2023uni,bhat2023loosecontrol,liu2024smartcontrol}. Specifically, LooseControl~\cite{bhat2023loosecontrol} extracts proxy depth from images to define 3D box controls and fine-tunes ControlNet~\cite{zhang2023adding} via LoRA~\cite{hu2021lora}, enabling the creation of complex environments (\eg, rooms, street scenes) by specifying only scene boundaries and key object locations.

\myparagraph{Image Personalization}
Adapter-based image personalization methods employ encoders to embed concepts (\eg, subject, human identity, style), offering a significant speed advantage over tuning-based approaches when extracting concepts from images.

Some methods adopt a domain-agnostic strategy, training encoders on open-world images to extract generalized subject-level conditions~\cite{wei2023elite,ma2023unified,chen2023subject,arar2023domain,ma2023subject,jiang2023videobooth,pan2023kosmos,song2024moma,huang2024realcustom,dat2025vsc,song2025harmonizing,purushwalkam2024bootpig}. 
These methods typically leverage large pretrained encoders such as CLIP~\cite{radford2021learning} and BLIP-2~\cite{Li2023BLIP2BL}, fine-tuning only lightweight projection layers~\cite{wei2023elite,ma2023unified,arar2023domain}. 
ELITE~\cite{wei2023elite}, for example, integrates a global mapping network and a local mapping network based on CLIP~\cite{radford2021learning}. The global network transforms hierarchical image features into multiple text embeddings, while the local network infuses patch features into cross-attention layers for detailed reconstruction. 
BLIP-Diffusion~\cite{li2023blip} advances customization by pre-training a BLIP-2~\cite{Li2023BLIP2BL} encoder for text-aligned image representation and developing a task for learning subject representations, enabling the generation of novel subject renditions.
Following on E4T~\cite{gal2023designing}, Arar \etal~\cite{arar2023domain} introduce an encoder for acquiring text embeddings and propose a hypernetwork to predict LoRA-style attention weight offsets in UNet. SuTI~\cite{chen2023subject} takes a unique approach inspired by apprenticeship learning~\cite{abbeel2004apprenticeship}, training a vast array of expert models on millions of internet image clusters. The apprentice model is then taught to imitate these experts' behaviors.
CAFE~\cite{zhou2023customization} build a customization assistant based on pre-trained large language model and diffusion model.

Some works design domain-aware encoders tailored to targeted domains~\cite{gal2023designing,shi2023instantbooth,cheong2023visconet}. 
In person-driven settings, facial images are encoded to provide identity conditions~\cite{ye2023ip,xiao2023fastcomposer,valevski2023face0,chen2023dreamidentity,chen2023photoverse,li2023photomaker,ruiz2023hyperdreambooth,peng2024portraitbooth,wang2024high,guo2024pulid,wang2024moa,cui2024idadapter,shiohara2024face2diffusion,liang2024caphuman,gal2024lcm,liu2024towards}. 
Face0~\cite{valevski2023face0} and DreamIdentity~\cite{chen2023dreamidentity} employ pretrained face recognition encoders~\cite{cao2018vggface2}; Face0 uses Inception-ResNet-V1~\cite{szegedy2017inception}, while DreamIdentity introduces a ViT-style M$^2$ID encoder~\cite{dosovitskiy2020image}. 
Beyond multimodal and face-recognition encoders~\cite{cao2018vggface2}, $\mathscr{W}^+$ Adapter~\cite{li2023stylegan} and PreciseControl~\cite{parihar2024precisecontrol} leverage GAN inversion~\cite{cao2022lsap,cao2024decreases} encoders as an alternative identity pathway.

Furthermore, some person-driven methods study the mechanism of combining textual embeddings with identity embeddings~\cite{xiao2023fastcomposer,li2023photomaker,cui2024idadapter}. 
To balance identity preservation and editability, FastComposer~\cite{xiao2023fastcomposer} and PhotoMaker~\cite{li2023photomaker} fuse text prompts with visual features from reference images; specifically, FastComposer mixes human-related text tokens (\eg,``man", ``woman") with visual features via a multilayer perceptron, and PhotoMaker applies two MLP layers to fuse image embeddings with the corresponding human-related embedding, then updates the latter with the fused representation. Beyond an identity encoder, some works additionally employ a spatial condition encoder to improve generation quality~\cite{han2024face}. T
his task also benefits from face segmentation masks and skeletal cues obtained from off-the-shelf models or annotations~\cite{xiao2023fastcomposer,li2023photomaker,achlioptas2023stellar,peng2023portraitbooth,ju2023humansd}. For example, Stellar~\cite{achlioptas2023stellar} uses face masks to remove background during preprocessing, sharpening focus on identity, while other methods leverage face masks to construct~\cite{xiao2023fastcomposer,li2023photomaker,peng2023portraitbooth} or adjust~\cite{hyung2023magicapture} loss functions.

Notably, task-specific encoders can be effective for image personalization. For style-driven generation, several studies use VGG features to better capture low-level style~\cite{chen2024artadapter,qi2024deadiff}. Prompt-free Diffusion~\cite{xu2024prompt} introduces SeeCoder, composed of a backbone encoder, a decoder, and a query transformer, enabling reference images to serve as conditions in lieu of text prompts.

\myparagraph{View-conditioned Generation}
Adapter-based view-conditioned generation aims to leverage explicit geometric or panoramic constraints within generative models to ensure spatial consistency across different viewpoints~\cite{seo2024genwarp,bai2024integrating,yuan2023customnet,zhang2024taming,wang2024customizing,kumari2024customizing,bernal2025precisecam,parihar2025compass}.
PreciseCam~\cite{bernal2025precisecam} converts four simple extrinsic and intrinsic camera parameters to a PF-US map and then controls generation by a trained ControlNet.
StitchDiffusion~\cite{wang2024customizing} extends adapter-based controllable generation to the domain of 360-degree panoramas by fine-tuning a T2I diffusion model with LoRA and introducing a stitching-aware denoising strategy. This approach ensures seamless global geometry and strong generalization for panoramic scene synthesis.

\myparagraph{Advaned Text-conditioned Generation}
To better extract faithful textual semantics, some works leverage LLM to replace the CLIP text encoder~\cite{chen2023tailored,wu2023paragraph,tan2024empirical}.
To improve the textual alignment of a long paragraph (up to 512 words), Wu \etal~\cite{wu2023paragraph} introduce an informative-enriched diffusion model for paragraph-to-image generation task, termed ParaDiffusion, which employ a large language model (\eg, Llama V2~\cite{touvron2023llama}) to encode long-form text, followed by fine-tuning with LoRA~\cite{hu2021lora} to align text-iamge feature spaces in generation.

Other methods design an extra pipeline or textual encoders to improve textual controllability~\cite{chen2023tailored,yang2024emogen}.
Tailored Visions~\cite{chen2023tailored} introduces a prompt rewriting system, leveraging historical user interactions to rewrite user prompts to enhance the expressiveness and alignment of user prompts with their intended visual outputs.

The text encoder is also studied to extend to multilingual version.
GlueGen~\cite{qin2023gluegen} aligns multilingual language models (\eg, XLM-Roberta~\cite{conneau2019unsupervised}) with existing text-to-image models, allowing for the generation of high-quality images from captions beyond English.
PEA-Diffusion~\cite{ma2023pea} is a proposed simple plug-and-play language transfer method based on knowledge distillation, where a lightweight MLP-like parameter-wefficient adapter with only 6M parameters is trained under teacher knowledge distillation along with a small parallel data corpus.

\myparagraph{In-Context Generation}
Wang et al.~\cite{wang2023context} introduced Prompt Diffusion, a novel approach that is jointly trained over multiple tasks using in-context prompts. This method has shown impressive results in high-quality in-context generation for trained tasks and effectively generalizes to new, unseen vision tasks with relevant prompts. 
Building upon this, Chen \etal~\cite{chen2023improving} further enhance Prompt Diffusion by incorporating a vision encoder-modulated text encoder. This innovation addresses several challenges, including costly pre-training, restrictive problem formulations, limited visual comprehension, and insufficient generalizability to out-of-distribution tasks.
Moreover, Najdenkoska \etal~\cite{najdenkoska2024context} propose a novel framework that separates the encoding of the visual context and preserves the structure of the query images. This results in the ability to learn from the visual context and text prompts, but also from either one of them.

\myparagraph{Brain-guided Generation}
\label{sec:brain}
The brain-guided generation tasks focus on controlling image creation directly from brain activities, such as electroencephalogram (EEG) recordings and functional magnetic resonance imaging (fMRI), bypassing the need to translate thoughts into text. More recently, advancements have been made with the adoption of visual diffusion models, offering enhanced capabilities in accurately translating complex brain activities into coherent visual representations\cite{chen2023seeing,takagi2023high,Ozcelik2023NaturalSR,lu2023minddiffuser,ni2023natural,bai2023dreamdiffusion,fu2023brainvis}. 

Chen \etal~\cite{chen2023seeing} present a Sparse Masked Brain Modeling with Doubled-Conditioned Latent Diffusion Model (MinD-Vis) for human vision decoding. They first learn an effective self-supervised representation of fMRI data using mask modeling and then augment latent diffusion model with double-conditioning. 
MindDiffuser~\cite{lu2023minddiffuser} is also a two-stage image reconstruction model. In the first stage, the VQ-VAE latent representations and the CLIP text embeddings decoded from fMRI are put into the image-to-image process of Stable Diffusion, which yields a preliminary image that contains semantic and structural information. Then, it utilizes the low-level CLIP visual features decoded from fMRI as supervisory information, and continually adjust the two features in the first stage through backpropagation to align the structural information.

While the above methods reconstruct visual results from fMRI, some approaches choose electroencephalogram (EEG)\cite{bai2023dreamdiffusion,fu2023brainvis}, which is a non-invasive and low-cost method of recording electrical activity in the brain. 
DreamDiffusion~\cite{bai2023dreamdiffusion} leverages
pre-trained text-to-image models and employs temporal masked signal modeling to pre-train the EEG encoder for effective and robust EEG representations. Additionally, the method further leverages a CLIP image encoder to provide extra supervision to better align EEG, text, and image embeddings with limited EEG-image pairs. 

\myparagraph{Sound-Guided Generation}
\label{sec:sound}
For sound-guided generation, some works develop an additional audio encoder, utilized to embed input audio into text embedding space or latent feature space to control generation~\cite{qin2023gluegen,zhang2024c3net,yang2023align}. 
GlueGen~\cite{qin2023gluegen} aligns multi-modal encoders such as AudioCLIP with the Stable Diffusion model, enabling sound-to-image generation.
Yang \etal~\cite{yang2023align} propose a unified framework ``Align, Adapt, and Inject" (AAI) for sound-guided image generation, editing, and stylization. In particular, this method adapts input sound into a sound token, like an ordinary word, which can plug and play with existing powerful diffusion-based Text-to-Image models.

\myparagraph{Text Rendering}
Drawing inspiration from the analysis in unCLIP~\cite{Ramesh2022HierarchicalTI}, which highlights the inadequacy of raw CLIP text embeddings in accurately modeling the spelling information in prompts, subsequent efforts such as eDiff-I~\cite{balaji2022ediffi} and Imagen~\cite{saharia2022photorealistic} have sought to harness the capabilities of large language models like T5~\cite{raffel2020exploring}, trained on text-only corpora, as text encoders in image generation. 
Additionally, DeepFloyd IF, following the design principles of Imagen~\cite{saharia2022photorealistic}, has demonstrated impressive proficiency in rendering legible text on images, showcasing a significant advancement in this challenging domain.
Meanwhile, some approaches are designed to improve text rendering capability for existing text-to-image diffusion models~\cite{liu2022character,chen2023textdiffuser,tuo2023anytext,chen2023textdiffuser2,zhao2023udifftext,zhang2023brush,chen2024textdiffuser,liu2024glyph}.
GlyphControl~\cite{yang2023glyphcontrol} leverages additional glyph conditional information to enhance the performance of the off-the-shelf Stable-Diffusion model in generating accurate visual text. 
TextDiffuser \etal~\cite{chen2023textdiffuser} first generates the layout of keywords extracted from text prompts and then generates images conditioned on the text prompt and the generated layout. The authors also contribute a large-scale text images dataset with OCR annotations, MARIO-10M, containing 10 million image-text pairs with text recognition, detection, and character-level segmentation annotations.
Zhang \etal~\cite{zhang2023brush} proposed Diff-Text, a training-free scene text generation framework for any language. Diff-text leverages rendered sketch images as priors to render text by ControlNet~\cite{zhang2023adding} and proposes a localized attention constraint to address the unreasonable position problem of scene text.

\begin{figure}[t!]
\begin{center}
    \includegraphics[width=1.0\linewidth]{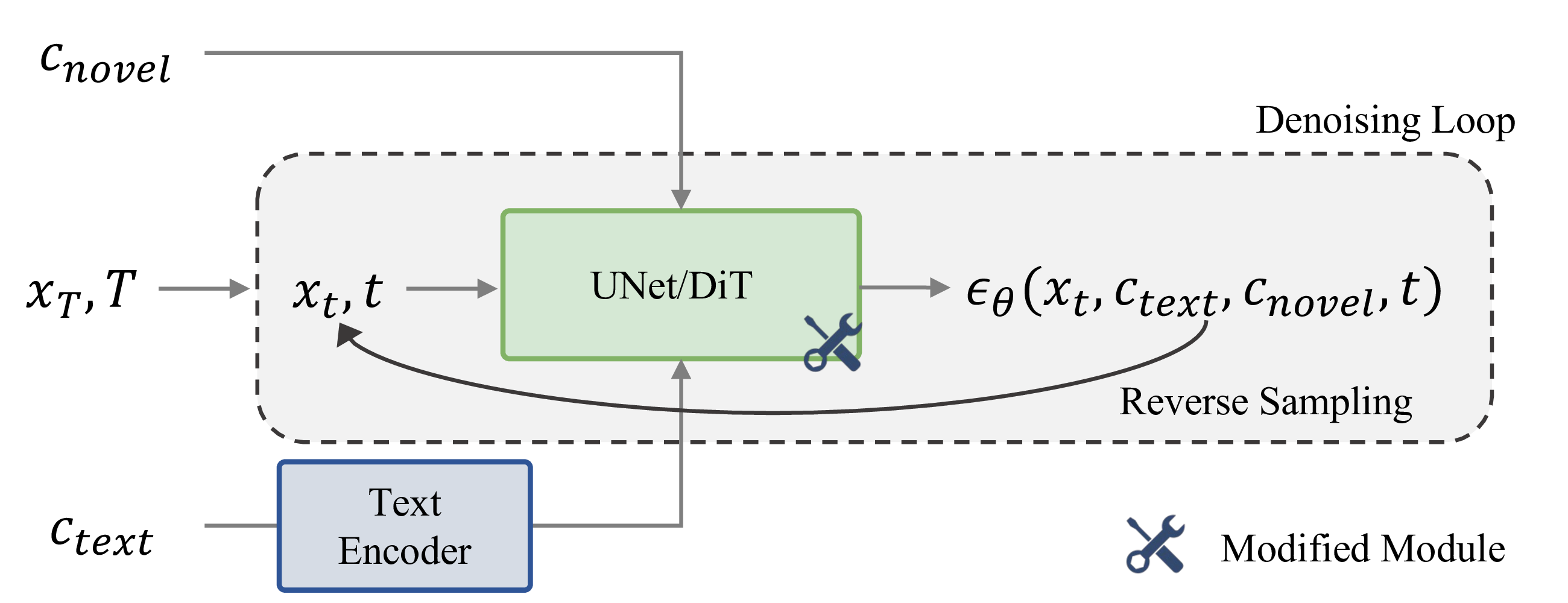}
\end{center}

\captionsetup{font=small}
\vspace{-2mm}
\caption{\small{
\textbf{Illustration of training-free conditional score prediction.}}}
\label{fig:mechanism_free}
\vspace{-4mm}
\end{figure}

\subsubsection{Training-free Conditional Score Prediction}
\label{sec:method_csp_free}
While the above techniques require training, a complementary line of work operates in a training-free fashion (see Fig.~\ref{fig:mechanism_free}). These methods analyze and exploit the model’s intrinsic properties so that a pretrained model can directly support novel conditions.

\myparagraph{Attention-map Adjustment}
Since attention maps are inherently interpretable, many controllable generation methods build upon attention map adjustment. 
Attention maps explicitly capture the relationships between tokens, particularly establishing correspondences between image features and textual tokens. 
This property enables direct control over the placement of textual concepts in generated images, including object positioning (i.e., layout-to-image)~\cite{zhao2023loco,wu2023harnessing,balaji2022ediffi,Kim2023DenseTG,zhao2023loco,mo2023freecontrol,zhao2025ltos,phung2024grounded,chen2024region,lee2024reground,wang2024compositional}, fine-grained control of objects and their attributes (i.e., attribute binding)~\cite{rassin2023linguistic}, and improving textual concept representation\cite{dahary2024yourself}.

For instance, since attention mechanisms explicitly model the relationships between text and image tokens, modulating the attention map becomes a pivotal training-free technique for controlling structure in score prediction~\cite{balaji2022ediffi,Kim2023DenseTG,zhao2023loco,mo2023freecontrol}.
eDiff-I~\cite{balaji2022ediffi} presents a technique named \emph{``paint-with-words"} (also known as pww), rectifying the cross-attention maps of each word by the correspondence segmentation maps to control the location of objects. 
Additionally, DenseDiffusion~\cite{Kim2023DenseTG} introduces a more extensive modulation method by devising multiple regularization, enhancing the precision and flexibility of layout control in score prediction.
Furthermore, Chen \etal~\cite{chen2024region} introduce a soft refinement phase to dismiss the visual boundaries and enhance adjacent interactions.

For the attribute binding task, Ge \etal~\cite{ge2023expressive} study the image generation from enriched textual description and propose a region-based diffusion, which constrains the object-level description into the object area via an attention map. Similarly, Structure Diffusion~\cite{feng2023trainingfree} employs linguistic insights to manipulate the cross-attention map, aiming for more accurate attribute binding and improved image composition.

\myparagraph{Feature Injection in Attention}
By injecting additional key–value pairs into the attention module, the denoising process can dynamically incorporate visual information from the provided references.
\cite{fan2024refdrop,li2024tuning,hertz2024style,hertz2024style,mo2024freecontrol,ding2024freecustom,gu2024analogist,heaid,hertz2023style}
Li \etal~\cite{li2024tuning} inject object feature from one of the reference images into the inversion process of another image to realize object placement in self-attention.
StyleAligned~\cite{hertz2023style} is designed to produce a series of images that adhere to a given reference style. This method introduces a novel attention sharing mechanism within the self-attention layers, which facilitates the interaction between the features of individual images and those of an additional reference image. Such a design enables the generation process to consider and incorporate style elements from multiple images simultaneously.
FreeControl~\cite{mo2024freecontrol} performs PCA to self-attention feature and replaces the principal components of feature in generation by reference components to control object appearance or spatial arrangement.
Furthermore, feature injection can also be applied to cross-attention~\cite{lv2024pick,liu2025training}.
Pick-and-draw~\cite{lv2024pick} extracts cross-attention map in reference image inversion and generation process, and then injection the inversion feature into the generation process via Earth Movers Distance(EMD) algorithm.

\myparagraph{Others}
FreeU \etal~\cite{si2024freeu} enhances diffusion image generation by strategically reweighting skip-connection and backbone feature contributions during inference—boosting coherence and fidelity without any additional training or parameters.
Basu \etal~\cite{basu2024mechanistic} propose Mechanistic Localization in text-to-image models, demonstrating that knowledge of visual attributes (\eg, “style,” “objects,” “facts”) can be localized to a small subset of UNet layers, thereby enabling more efficient model editing.

To synthesize high-resolution images, MultiDiffusion~\cite{bar2023multidiffusion} formulates an optimization problem that enforces each crop to remain consistent with its denoised counterpart. Although individual denoising steps may introduce conflicting directions, the method fuses them into a unified global denoising step, ultimately producing seamless and high-quality images.
Zhou \etal~\cite{zhou2025exploring} find that the padding is the pivotal mechanism to object arrangement, which is degraded in high-resolution generation. Based on that, they introduce a Progressive Boundary Complement, which creates dynamic virtual image boundaries inside the feature map to enhance position information propagation.

Meanwhile, from the noise initialization perspective, InitNo~\cite{guo2024initno} first samples a lot of initial noise and then designs a cross-attention response score and the self-attention conflict score to evaluate them to find a better one.

\begin{figure}[t!]
\begin{center}
    \includegraphics[width=1.0\linewidth]{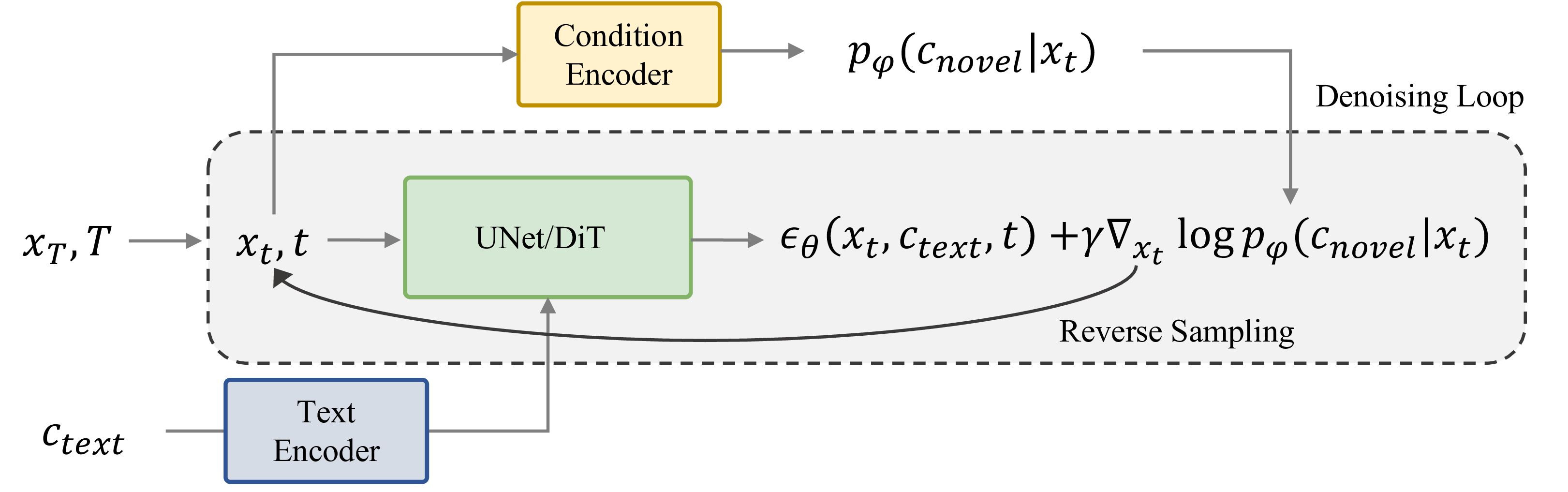}
\end{center}
\captionsetup{font=small}
\vspace{-2mm}
\caption{\small{
\textbf{Illustration of condition-guided conditional score estimation.}}}
\label{fig:method_cse}
\vspace{-5mm}
\end{figure}

\subsection{Condition-guided Score Estimation}
\label{sec:method_cse}

While numerous methods adhere to the paradigm of conditional score prediction, some studies explore controllable generation by performing condition prediction from latents or intermediate features during the generative process, and then computing losses against the given conditions to provide gradient guidance for denoising~\cite{liu2023late,voynov2023sketch,bansal2023universal,liu2023late,yu2023freedom,xiaor,couairon2023zero,luo2025adding},  as shown in Fig.~\ref{fig:method_cse}.

LGP~\cite{voynov2023sketch} stands as an early pioneer, which innovatively introduces a Latent Edge Predictor, designed to extrapolate sketch information from a series of intermediate features within a UNet architecture. It employs the degree of similarity between the condition sketch and predicted sketch to compute gradients, which are then utilized to guide the score estimation process. Its methodologies and insights have been a source of inspiration for numerous subsequent research endeavors in this field\cite{liu2023late,phung2023grounded,couairon2023zero,xiao2023r}.
Furthermore, Universal Guidance~\cite{bansal2023universal} and FreeDom~\cite{yu2023freedom} are proposed to leverage image-space off-the-shelf predictors to guide denoising. At each denoising step, it attains the clean image by one-step denoising to calculate the guidance gradient.

While the aforementioned methods require a condition predictor to backpropagate condition guidance, layout and segmentation guidance can also be directly estimated through attentiomap, eliminating the need for additional trained models~\cite{zhao2025local,xiaor,couairon2023zero,phung2024grounded,lv2024pick,xie2023boxdiff,chen2025versagen,wang2024magic,patel2025enhancing}.
For instance, BoxDiff~\cite{xie2023boxdiff} designs three spatial constraints (i.e., Inner-Box, Outer-Box, and Corner Constraints) to guide the denoising process.
ZestGuide~\cite{couairon2023zero} leverages segmentation maps extracted from cross-attention layers, aligning generation with input masks through gradient-based guidance during denoising. 
To place the object at a specific position, VersaGen~\cite{chen2025versagen} calculates the loss from the object-token attention map and the given segmentation.
Additionally, VODiff~\cite{liang2025vodiff} studies the objects' visibility order and proposes visibility-order-aware loss.

Measuring the representation of textual concepts as a denoising guidance can help improve textual alignment~\cite{chefer2023attend,rassin2023linguistic,li2023divide}. 
Attend-and-Excite~\cite{chefer2023attend}(A\&E) represents an early effort in this area, introducing an attention-based Generative Semantic Nursing (GSN) mechanism. This mechanism refines cross-attention units to more effectively ensure that all subjects described in the text prompt are accurately generated. 
EBAMA~\cite{zhang2024object} extend A\&E by introducing an attribute binding loss to address semantic misalignment.
Additionally, SynGen~\cite{rassin2023linguistic} employs a unique methodology in text-to-image generation by first conducting a syntactic analysis of the text prompt. This analysis aims to identify entities and their modifiers within the prompt. Following this, SynGen utilizes a novel loss function designed to align the cross-attention maps with the linguistic bindings as indicated by the syntax. 

\end{revrange1}

\section{Controllable Generation with Multiple Conditions}
\label{sec:multi-conditon}

\begin{figure}[t!]
	\begin{center}
		\includegraphics[width=0.9\linewidth]{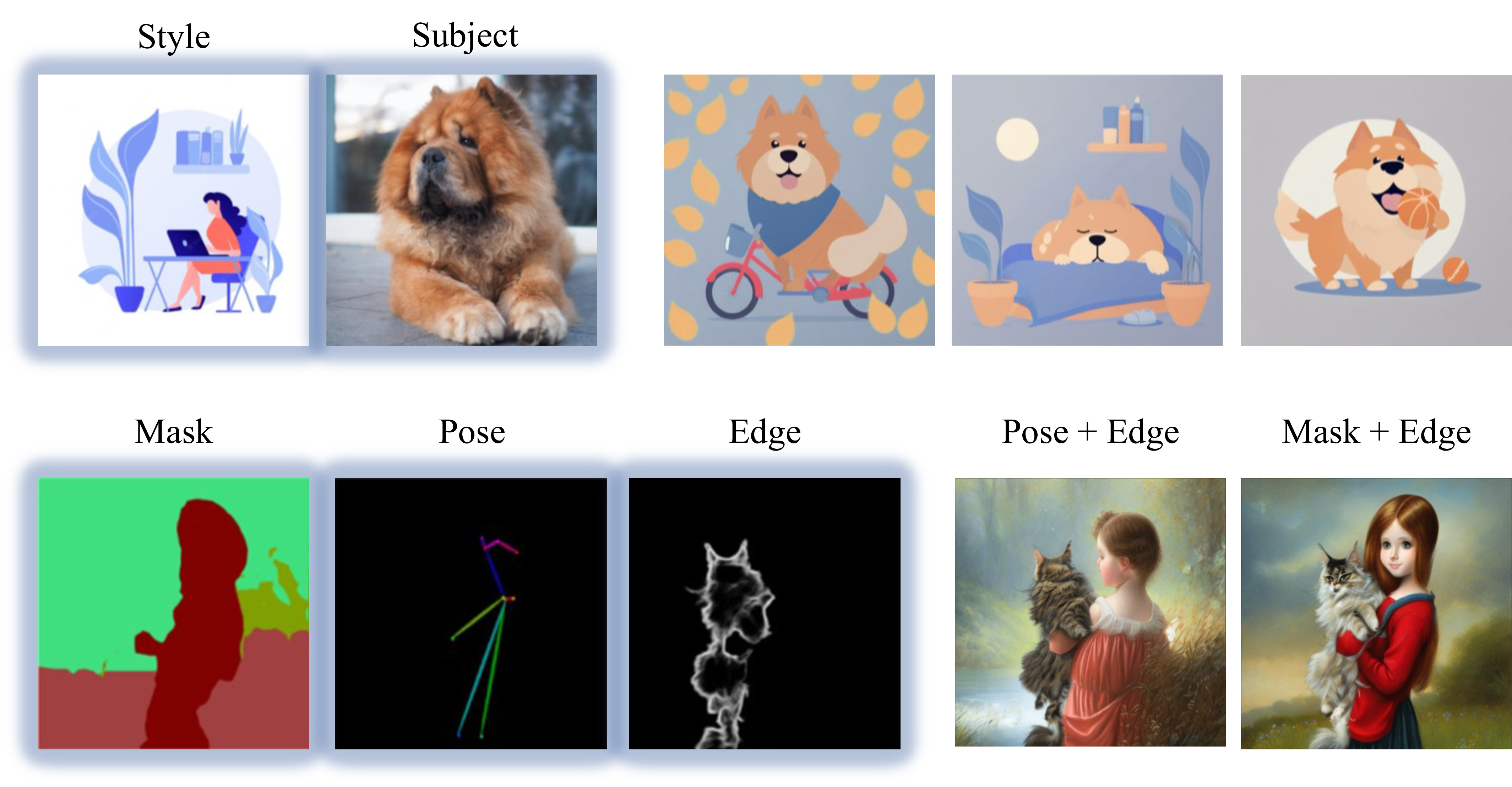}
	\end{center}
	\captionsetup{font=small}
\vspace{-5mm}
	\caption{\small{
\textbf{Illustration of multi-conditioned generation.} The condition is marked in blue background. Examples are sourced from \cite{shah2023ziplora,hu2023cocktail}.}}
	\label{fig:multi}
\vspace{-5mm}
\end{figure}

The multi-condition generation task aims to generate images under multiple conditions (see Fig.~\ref{fig:multi}). In this section, we conduct a comprehensive overview of these methods from a technical perspective, categorizing them into joint training (Sec.~\ref{sec:multi_joint}), continual learning (Sec.~\ref{sec:multi_continual}), weight fusion (Sec.~\ref{sec:multi_fusion}), attention-based integration (Sec.~\ref{sec:multi_attention}), and guidance composition (Sec.~\ref{sec:multi_guidance}). 
\begin{revrange}
Note that some of the other controllable generation methods also demonstrate multi-condition synthesis capability without dedicated designs~\cite{zhang2023adding,gal2022image,ruiz2023dreambooth,qin2023unicontrol}.
\end{revrange}

\subsection{Joint Training}
\label{sec:multi_joint}
Designing a multi-condition framework and jointly training them is a simple yet effective route to realize multi-condition generation. These methods generally focus on multi-condition encoders and training strategies~\cite{frenkel2024implicit,huang2023composer,hu2023cocktail,wang2025unicombine,han2023svdiff,han2023svdiff,xiao2023fastcomposer,lin2024non,nair2024maxfusion,lu2024coarse}.

Composer~\cite{huang2023composer} projects all conditions (including text caption, depthmap, sketch, and \emph{etc.}) into uniform-dimensional embeddings with the same spatial size as the noisy latent using stacked convolutional layers. It leverages a joint training strategy to generate images from a set of representations, where it uses an independent dropout probability of 0.5 for each condition, a probability of 0.1 for dropping all conditions, and a probability of 0.1 for retaining all conditions.
Additionally, Cocktail~\cite{hu2023cocktail} proposes the controllable normalization method (ControlNorm), which has an additional layer to generate two sets of learnable parameters conditioned on all modalities. These two sets of parameters are used to fuse the external conditional signals and the original signals.
UniCombine~\cite{wang2025unicombine} designs a Conditional MMDiT Attention mechanism, where condition-specific LoRA modules mask the attention across different conditions to naturally support multiple conditions.

From a data perspective, SVDiff~\cite{han2023svdiff} utilizes a cut-mix-unmix mechanism for a multi-subject generation. It augments multi-concept data by a CutMix-like data augmentation and rewrites the correspondence text prompt. It also leverages an unmix regularization on cross-attention maps, ensuring text embeddings are only effective in the correspondence areas. This attention map constraint mechanism is also applied in FastComposer~\cite{xiao2023fastcomposer}.

\subsection{Continual Learning}
\label{sec:multi_continual}
Continual learning methods are generally proposed to address knowledge ``catastrophic forgetting" in tuning-based conditional score prediction works~\cite{liu2025museummaker,smith2023continual,sun2023create,smith2023continual2,guo2025conceptguard}.
Specifically, C-LoRA~\cite{smith2023continual} is composed of a continually self-regularized LoRA in cross-attention layers. It utilizes the past LoRA weight deltas to regulate the new LoRA weight deltas by guiding which parameters are most available to be updated for continual concept learning.
Moreover, L$^2$DM~\cite{sun2023create} devises a task-aware memory enhancement module and an elastic-concept distillation module, which could respectively safeguard the knowledge of both prior concepts and each past personalized concept. It utilizes a rainbow-memory bank strategy to manage long-term and short-term memory and provide regularization samples to safeguard the knowledge in the personalization process. During training, the authors further propose a concept attention artist module and orthogonal attention artist module to update noisec latent for better performance.
STAMINA~\cite{smith2023continual2} introduces forgetting-regularization and sparsity-regularization in continual learning, avoiding forgetting learned concepts and ensuring no cost to storage or inference.
ConceptGuard~\cite{guo2025conceptguard} combines shift embedding, concept-binding prompts, and memory preservation regularization to support new concepts.

\subsection{Weight Fusion}
\label{sec:multi_fusion}
In the realm of adapting T2I diffusion models to novel conditions via fine-tuning, weight fusion presents itself as an intuitive approach for merging multiple conditions. These methods focus on achieving a cohesive blend of weights that incorporates each condition while ensuring that the controllability of individual conditions is retained. The goal is to seamlessly integrate various conditional aspects into a unified model, thereby enhancing its versatility and applicability across diverse scenarios. This requires a delicate balance between maintaining the integrity of each condition's influence and achieving an effective overall synthesis.

Since personalized conditions usually represent UNet's weight or text embeddings, weight fusion is an intuitive and effective way to generate images under multiple personalized conditions.
Specifically, Cones~\cite{liu2023cones} further fine-tunes the concept neurons after personalization for better generation quality and multi-subject generating capability.
Custom Diffusion~\cite{kumari2023multi} introduces a constrained optimization method to merge fine-tuned key and value matrices, as follows:
\begin{align}
\label{eq:custom_diffusion}
    \hat{W} = \arg\min_W \|&WC_{\text{reg}} - W_0C_{\text{reg}}\|_F \\
\text{s.t. } WC^T &= V, \text{ where } C = [c_1 \ldots c_N]^T \nonumber\\
\text{ and } V &= [W_1c_1^T \ldots W_Nc_N^T]^T\nonumber
\end{align}
where $\{W_{n,l}^k,W_{n,l}^v\}_{n=1}^N$ represent the corresponding updated key and value matrices for added $N$ concepts and $C_{\text{reg}}$ is a randomly sampled text features for regularization. The objective of Eq.~\ref{eq:custom_diffusion} is intuitively designed to ensure that the words in the target captions are consistently aligned with the values derived from the concept matrices that have undergone fine-tuning. 
Similarly, Mix-of-Show~\cite{gu2023mix} introduces the gradient fusion, updating weight $W$ by $W = \arg\min_W \sum_{i=1}^{n} \left\| (W_0 + \Delta W_i)X_i - WX_i \right\|_F^2$ where $X_i$ represents the input activation of the $i$-th concept, and $|\cdot|_F$ denotes the Frobenius norm. To integrate subject-centric and style-centric conditions, ZipLoRA~\cite{shah2023ziplora} merges LoRA-style weights by minimizing the difference between subject/style images generated by the mixed and original LoRA models and the cosine similarity between the columns of content and style LoRAs.
Po \etal~\cite{po2023orthogonal} present orthogonal adaption to replace LoRA in fine-tuning, encouraging the customized models to have orthogonal residual weights for efficient fusion.

\subsection{Attention-based Integration}
\label{sec:multi_attention}
Attention-based integration methods modulate attention maps to strategically position subjects within the synthesized image, allowing for precise control over where and how each condition is represented in the final composition~\cite{liu2023cones2,gu2023mix,kwon2024concept}. 

For example, Cones2~\cite{liu2023cones2} edits cross-attention map by $\text{Edited}CA \leftarrow Softmax(CA\oplus\{\eta(t)\cdot M_{s_i}|i=1,\cdots,N\}$, where $\oplus$ denotes the operation that adds the corresponding dimension of cross-attention map $CA$ and pre-defined layout $M$ and $\eta(t)$ is a concave function controlling the edit intensity at different timestep $t$.
Similarly, Mix-of-Show~\cite{gu2023mix} employs a regionally controllable sampling method, integrating global prompt and multiple regional prompts with pre-defined masks in cross-attention.

\subsection{Guidance Composition}
\label{sec:multi_guidance}
Guidance composition is an integration mechanism for synthesizing images under multiple conditions, integrating the independent denoising results of each condition~\cite{huang2023collaborative,wang2023high,cao2025image,wang2023decompose,chan2024improving,wang2024text}. This process is mathematically represented as:
\begin{equation}
    \hat\epsilon(z_t,c_1,\cdots,c_N)=\sum_{i=1}^K w_i\cdot\mathcal{M}_i\cdot \epsilon(z_t,c_i)
\end{equation}
where $\epsilon(z_t,c_i)$ denotes the guidance of each condition, while $w_i$ and $\mathcal{M}_i$ are the respective weights and spatial mask used to integrate these results.

To integrate multiple concepts, Decompose and Realign~\cite{wang2023decompose} obtains the corresponding $\mathcal{M}_i$ by their cross-attention map.
Similarly, Face-diffuser~\cite{wang2023high} presents a saliency-adaptive noise fusion method to combine results from a text-driven diffusion model and a proposed subject-augmented diffusion model.
 
Besides, to realize controllable generation in user-specific domain, Cao \etal~\cite{cao2025image} train a null-text UNet to provide domain guidance and utilize the original diffusion prior to provide control guidance.

\section{Universal Controllable Text-to-Image Generation}
\label{sec:universal}
Beyond approaches tailored to specific types of conditions, there exist universal methods designed to accommodate arbitrary conditions in image generation. These methods are broadly categorized into two groups based on their theoretical foundations: universal conditional score prediction framework and universal condition-guided score estimation. 

\subsection{Universal Conditional Score Prediction Framework}
\label{sec:uni_cond}
Universal conditional score prediction framework involves creating a framework capable of encoding any given conditions and utilizing them to predict the noise at each timestep during the image synthesis process. This approach provides a universal solution that adapts flexibly to diverse conditions. By integrating the conditional information directly into the generative model, this method allows for the dynamic adaptation of the image generation process in response to a wide array of conditions, making it versatile and applicable to various image synthesis scenarios.

DiffBlender~\cite{kim2023diffblender} is proposed to incorporate conditions from diverse types of modalities. It categorizes conditions into multiple types to employ different techniques for guiding generation. First, image-form conditions, which contain spatially rich information, are injected in ResNet Blocks~\cite{he2016deep}. Then, spatial conditions, including grounding box and keypoints, are passed through a local self-attention module to accurately locate the desired positions of synthesized results. Moreover, non-spatial conditions like color palette and style are concatenated with textual tokens through a global self-attention module and then fed into cross-attention layers. 
Additionally, Emu2~\cite{sun2023generative1} leverages a large generative multimodal model with 37 billion parameters for task-agnostic in-context learning to construct a universal controllable T2I generation framework.

\subsection{Universal Condition-Guided Score Estimation}
\label{sec:uni_guide}
Other approaches utilize condition-guided score estimation to incorporate various conditions into the text-to-image diffusion models. The primary challenge lies in obtaining condition-specific guidance from the latent during the denoising process.

Universal Guidance~\cite{bansal2023universal} observes that the reconstructed clean image proposed in the denoising diffusion implicit model (DDIM)~\cite{song2020denoising} is appropriate for a generic guidance function to provide informative feedback to guide the image generation. Given any condition $c$ and off-the-shelf predictor $f$, the denoising process is guided by:
\begin{align}
\hat{\epsilon}_{\theta}(z_t, t) = \epsilon_{\theta}(z_t, t) + s(t) \cdot \nabla_{z_t} \mathcal{L}(c, f(\hat{z}_0))
\end{align}
where $\hat{z}_0$ is the predicted clean image following~\cite{song2020denoising}:
\begin{align}
\hat{z}_0 = \frac{z_t - (\sqrt{1 - \alpha_t})\epsilon_{\theta}(z_t, t)}{\sqrt{\alpha_t}}
\end{align}
UG employs various predictors, including CLIP~\cite{radford2021learning} (for text or style conditions), segmentation network~\cite{howard2019searching} (for segmentation map conditions), face recognition model~\cite{zhang2016joint,schroff2015facenet} (for identity conditions), and object detector~\cite{ren2015faster} (for bounding box conditions), in experiments to exhibit conditional generation capabilities with various conditions. 

Similar to Universal Guidance~\cite{bansal2023universal}, FreeDom~\cite{yu2023freedom} leverages off-the-shelf predictors to construct time-independent energy functions to guide the generation process. It also develops the efficient time-travel strategy, taking the current intermediate result $z_t$ back by $j$ steps to $z_{t+j}$ and resampling it to the $t$-th timestep. This mechanism solves the problem of misalignment with conditions on large data domains, \eg ImageNet~\cite{deng2009imagenet}.

While above mentioned condition-guided sampling approaches leverage off-the-shelf models and one-step estimation procedure to predict condition-related conditions, Pan \etal~\cite{pan2023towards} present Symplecit Adjoint Guidance (SAG) in two inner stages, where SAG first estimate the clean image via $n$ function calls and then uses the symplectic adjoint method to obtain the gradients accurately.

\begin{figure*}[t!]
	\begin{center}
		\includegraphics[width=0.9\linewidth]{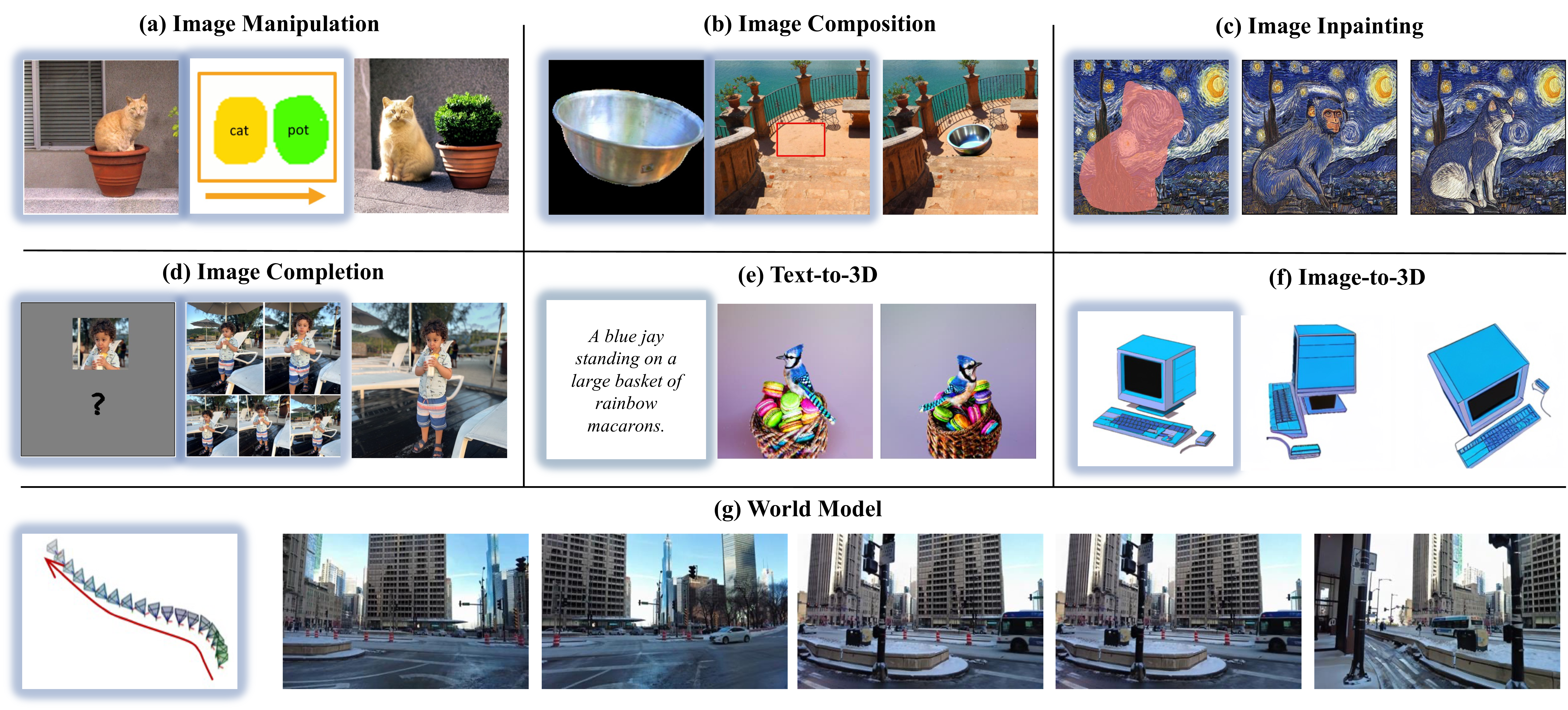}
	\end{center}
	\vspace{-5mm}
	\captionsetup{font=small}
	\caption{\small{
\textbf{Illustration of the application of controllable text-to-image generation.} The condition is marked in blue background. Examples are sourced from \cite{zhang2023continuous, song2023objectstitch, xie2023dreaminpainter,tang2023realfill,wang2023prolificdreamer,liu2023zero,he2025cameractrl}.}}
\vspace{-5mm}
\label{fig:application}
\end{figure*}

\section{Applications}
\label{sec:application}
In this section, we focus on innovative methods that utilize novel conditions in the generation process to address specific tasks. By emphasizing these pioneering approaches, we aim to highlight how conditional generation is not only reshaping the landscape of content creation but also broadening the horizons of creativity and functionality in various fields. 
The subsequent discussions will provide insights into the transformative impact of these models and their potential in diverse applications. We illustrate the example of the applications in Fig.~\ref{fig:application}.

\subsection{Image Manipulation}
Advancements in the control of pre-trained text-to-image diffusion models have allowed for more versatile image editing techniques. For instance, inspired by DreamBooth\cite{ruiz2023dreambooth}, SINE\cite{Zhang_2023_CVPR} constructs the text prompt for fine-tuning the pretrained text-to-image model by the source image as ''a photo/painting of a [$*$] [class]" and edits the image by a novel adapter-based classifier-free guidance.
Moreover, the versatility of control conditions further enhances the editing process by integrating conditions beyond mere text. For example, Choi \etal \cite{choi2023custom} customize the diffusion model to employ specific elements from the reference image as editing criteria, such as substituting the cat in the source image with the cat's appearance in the reference image. 
\begin{revrange}
Recently, Zhou \etal~\cite{zhou2025multi} modify the score estimation in multi-turn editing, introducing a dual-objective Linear Quadratic Regulators (LQR)to effectively mitigate error accumulation.
\end{revrange}

\subsection{Image Completion and Inpainting}
The advancement of flexible control mechanisms has also significantly expanded the capabilities in the field of image inpainting and completion. Specifically, DreamInpainter \cite{xie2023dreaminpainter} utilizes a subject-driven generation approach to personalize the filling of masked areas with the aid of reference images. Besides, Realfill \cite{tang2023realfill} takes similar methods that employ reference images to facilitate realistic and coherent image completions. Moreover, by multiple condition controlling, Uni-inpaint \cite{yang2023uni} integrates a diverse set of control conditions such as text descriptions, strokes, and exemplar images to simultaneously direct the generation within the masked regions. 

\subsection{Image Composition}
Image composition is a challenging task that involves multiple complex image process stages like color harmonization, geometry correction, shadow generation, and so on. While the strong prior in large-scale pre-trained diffusion model can address the problem in a unified manner. Through adding adapters to control the pre-trained text-to-image diffusion model, ObjectStitch\cite{song2023objectstitch} presents an object composition framework that can handle multiple aspects such as viewpoint, geometry, lighting, and shadow. Moreover, DreamCom\cite{lu2023dreamcom} customizes the text-to-image model on several foreground object images to enhance the object details' preservability. Besides, by inserting the task indicator vector into U-Net to control the generating process, ControlCom\cite{zhang2023controlcom} proposes a controllable image composition method that unifies four composition-related tasks with an indicator vector.

\subsection{Text/Image-to-3D Generation}
Text/image-to-3D task aims to reconstruct 3D representations from text descriptions or images (pairs). 
Recent advancements in text/image-to-3D generation represent a significant milestone with the development of Score Distillation Sampling (SDS) loss. This innovative approach, introduced by DreamFusion\cite{poole2022dreamfusion}, marks a successful adaptation of large-scale 2D diffusion models for 3D generation. Through SDS, the control method of the text-to-image model can be transferred to text-to-3D generation. Typically, DreamBooth3D\cite{raj2023dreambooth3d} combines DreamBooth\cite{ruiz2023dreambooth} and DreamFusion\cite{poole2022dreamfusion} that personalizes text-to-3d generative models from a few captured images of a subject. Similarly, some approaches\cite{chen2023control3d, huang2023dreamcontrol, yu2023points, huang2023dreamcontrol} adapt ControlNet\cite{zhang2023adding} to the SDS process, enabling the control of 3D generation through spatial signals (\emph{\eg,} depth map, sketch). 

\begin{revrange}

\subsection{World Model} 
The condition-injection mechanisms provide effective support for the development of video-generation-based world models. A representative example is camera-controlled video generation~\cite{wang2024motionctrl, he2025cameractrl, bahmani2024vd3d, yuegosim,bahmani2025ac3d,ren2025gen3c,bai2025recammaster}, which focuses on aligning sequences of camera parameters with the diffusion-based video generation process. For instance, ReCamMaster~\cite{bai2025recammaster} incorporates camera parameters into the original DiT blocks via frame-dimension conditioning. Similarly, AC3D~\cite{bahmani2025ac3d} introduces camera information through lightweight 128-dimensional DiT-SX blocks. In addition, several approaches tackle this problem in a training-free manner~\cite{hou2024training, ling2024motionclone, song2025worldforge}. Typically, WorldForge~\cite{song2025worldforge}, which leverages 3D/4D foundation models~\cite{wang2025vggt} to project video frames into a static point cloud. The point cloud is then adjusted according to camera trajectories and subsequently used as guidance for video generation.

\end{revrange}

\section{Conclusion}
\label{sec:conclusion}
In this comprehensive survey, we delve into the realm of conditional generation with text-to-image diffusion models, unveiling the novel conditions incorporated in the text-guided generation process. Initially, we equip readers with foundational knowledge, introducing the denoising diffusion probability models, prominent text-to-image diffusion models, and a well-structured taxonomy. Subsequently, we reveal the mechanisms of introducing novel conditions into T2I diffusion models.
Then, we present a summary of previous conditional generation methods and analyze them in terms of theoretical foundations, technical advancements, and solution strategies. Furthermore, we explore the practical applications of controllable generation, underscoring its vital role and immense potential in the era of AI-generated content. This survey aims to provide a comprehensive understanding of the current landscape of controllable T2I generation, thereby contributing to the ongoing evolution and expansion of this dynamic research area.

\begin{revrange}
Although controllable generation with text-to-image diffusion models has achieved remarkable progress, several promising directions remain open for future exploration:

(1) \emph{Towards a universal and cross-modal control paradigm.}  
Existing controllable diffusion models are often tailored to specific conditions or tasks. Future research could focus on developing unified and generalizable control frameworks capable of flexibly accommodating diverse forms of conditions, including spatial, semantic, and multimodal inputs, within a single generative system. Extending controllability beyond text-to-image synthesis to other modalities such as audio, video, and 3D generation will further enhance the model’s adaptability and cross-modal reasoning ability, paving the way toward general multimodal intelligence.  

(2) \emph{Building world models with controlling mechanisms.}  
The condition-injection and controllability principles of diffusion models provide a strong foundation for constructing world models based on video generation. Future studies could explore how diffusion-based systems simulate dynamic, camera-controllable environments while maintaining temporal–spatial consistency. Such world models will play a crucial role in connecting generative modeling, embodied AI, and interactive virtual environments.  

\end{revrange}

\ifCLASSOPTIONcaptionsoff
  \newpage
\fi

{\small
\bibliographystyle{unsrt2authabbrvpp}
\bibliography{negbib}

@String(IJCV = {Int. J. Comput. Vis.})

@String(CVPR= {IEEE Conf. Comput. Vis. Pattern Recog.})

@String(ICCV= {Int. Conf. Comput. Vis.})

@String(ECCV= {Eur. Conf. Comput. Vis.})

@String(NIPS= {Adv. Neural Inform. Process. Syst.})

@String(BMVC= {Brit. Mach. Vis. Conf.})

@String(TOG= {ACM Trans. Graph.})

@String(ACMMM= {ACM Int. Conf. Multimedia})

@String(ICASSP=	{ICASSP})

@String(ICLR = {Int. Conf. Learn. Represent.})

@String(IJCAI = {IJCAI})

@String(AAAI = {AAAI})

@String(IJCV  = {IJCV})

@String(CVPR  = {CVPR})

@String(ICCV  = {ICCV})

@String(ECCV  = {ECCV})

@String(NIPS  = {NeurIPS})

@String(BMVC  =	{BMVC})

@String(TOG   = {ACM TOG})

@String(ACMMM = {ACM MM})

@String(ICLR  = {ICLR})

@inproceedings{liu2023zero,
  title={Zero-1-to-3: Zero-shot one image to 3d object},
  author={Liu, Ruoshi and Wu, Rundi and Van Hoorick, Basile and Tokmakov, Pavel and Zakharov, Sergey and Vondrick, Carl},
  booktitle={ICCV},
  pages={9298--9309},
  year={2023}
}

@article{zhang2023controlcom,
  title={Controlcom: Controllable image composition using diffusion model},
  author={Zhang, Bo and Duan, Yuxuan and Lan, Jun and Hong, Yan and Zhu, Huijia and Wang, Weiqiang and Niu, Li},
  journal={arXiv preprint arXiv:2308.10040},
  year={2023}
}

@article{huang2023dreamcontrol,
  title={DreamControl: Control-Based Text-to-3D Generation with 3D Self-Prior},
  author={Huang, Tianyu and Zeng, Yihan and Zhang, Zhilu and Xu, Wan and Xu, Hang and Xu, Songcen and Lau, Rynson WH and Zuo, Wangmeng},
  journal={arXiv preprint arXiv:2312.06439},
  year={2023}
}

@inproceedings{yu2023points,
  title={Points-to-3d: Bridging the gap between sparse points and shape-controllable text-to-3d generation},
  author={Yu, Chaohui and Zhou, Qiang and Li, Jingliang and Zhang, Zhe and Wang, Zhibin and Wang, Fan},
  booktitle={ACMMM},
  pages={6841--6850},
  year={2023}
}

@inproceedings{chen2023control3d,
  title={Control3d: Towards controllable text-to-3d generation},
  author={Chen, Yang and Pan, Yingwei and Li, Yehao and Yao, Ting and Mei, Tao},
  booktitle={ACMMM},
  pages={1148--1156},
  year={2023}
}

@article{raj2023dreambooth3d,
  title={Dreambooth3d: Subject-driven text-to-3d generation},
  author={Raj, Amit and Kaza, Srinivas and Poole, Ben and Niemeyer, Michael and Ruiz, Nataniel and Mildenhall, Ben and Zada, Shiran and Aberman, Kfir and Rubinstein, Michael and Barron, Jonathan and others},
  journal={arXiv preprint arXiv:2303.13508},
  year={2023}
}

@inproceedings{poole2022dreamfusion,
  title={DreamFusion: Text-to-3D using 2D Diffusion},
  author={Poole, Ben and Jain, Ajay and Barron, Jonathan T and Mildenhall, Ben},
  booktitle={ICLR},
  year={2022}
}

@article{lu2023dreamcom,
  title={Dreamcom: Finetuning text-guided inpainting model for image composition},
  author={Lu, Lingxiao and Zhang, Bo and Niu, Li},
  journal={arXiv preprint arXiv:2309.15508},
  year={2023}
}

@inproceedings{song2023objectstitch,
  title={ObjectStitch: Object Compositing With Diffusion Model},
  author={Song, Yizhi and Zhang, Zhifei and Lin, Zhe and Cohen, Scott and Price, Brian and Zhang, Jianming and Kim, Soo Ye and Aliaga, Daniel},
  booktitle={CVPR},
  pages={18310--18319},
  year={2023}
}

@article{tang2023realfill,
  title={Realfill: Reference-driven generation for authentic image completion},
  author={Tang, Luming and Ruiz, Nataniel and Chu, Qinghao and Li, Yuanzhen and Holynski, Aleksander and Jacobs, David E and Hariharan, Bharath and Pritch, Yael and Wadhwa, Neal and Aberman, Kfir and others},
  journal={arXiv preprint arXiv:2309.16668},
  year={2023}
}

@inproceedings{yang2023uni,
  title={Uni-paint: A Unified Framework for Multimodal Image Inpainting with Pretrained Diffusion Model},
  author={Yang, Shiyuan and Chen, Xiaodong and Liao, Jing},
  booktitle={ACMMM},
  pages={3190--3199},
  year={2023}
}

@article{xie2023dreaminpainter,
  title={DreamInpainter: Text-Guided Subject-Driven Image Inpainting with Diffusion Models},
  author={Xie, Shaoan and Zhao, Yang and Xiao, Zhisheng and Chan, Kelvin CK and Li, Yandong and Xu, Yanwu and Zhang, Kun and Hou, Tingbo},
  journal={arXiv preprint arXiv:2312.03771},
  year={2023}
}

@inproceedings{zhang2023continuous,
  title={Continuous layout editing of single images with diffusion models},
  author={Zhang, Zhiyuan and Huang, Zhitong and Liao, Jing},
  booktitle={Computer Graphics Forum},
  volume={42},
  number={7},
  pages={e14966},
  year={2023},
  organization={Wiley Online Library}
}

@InProceedings{Zhang_2023_CVPR,
    author    = {Zhang, Zhixing and Han, Ligong and Ghosh, Arnab and Metaxas, Dimitris N. and Ren, Jian},
    title     = {SINE: SINgle Image Editing With Text-to-Image Diffusion Models},
    booktitle = {CVPR},
    month     = {June},
    year      = {2023},
    pages     = {6027-6037}
}

@article{choi2023custom,
  title={Custom-edit: Text-guided image editing with customized diffusion models},
  author={Choi, Jooyoung and Choi, Yunjey and Kim, Yunji and Kim, Junho and Yoon, Sungroh},
  journal={arXiv preprint arXiv:2305.15779},
  year={2023}
}

@article{krishna2017visual,
  title={Visual genome: Connecting language and vision using crowdsourced dense image annotations},
  author={Krishna, Ranjay and Zhu, Yuke and Groth, Oliver and Johnson, Justin and Hata, Kenji and Kravitz, Joshua and Chen, Stephanie and Kalantidis, Yannis and Li, Li-Jia and Shamma, David A and others},
  journal={IJCV},
  volume={123},
  pages={32--73},
  year={2017},
  publisher={Springer}
}

@inproceedings{lin2014microsoft,
  title={Microsoft coco: Common objects in context},
  author={Lin, Tsung-Yi and Maire, Michael and Belongie, Serge and Hays, James and Perona, Pietro and Ramanan, Deva and Doll{\'a}r, Piotr and Zitnick, C Lawrence},
  booktitle={ECCV},
  pages={740--755},
  year={2014},
  organization={Springer}
}

@article{luo2022understanding,
  title={Understanding diffusion models: A unified perspective},
  author={Luo, Calvin},
  journal={arXiv preprint arXiv:2208.11970},
  year={2022}
}

@article{podell2023sdxl,
  title={Sdxl: Improving latent diffusion models for high-resolution image synthesis},
  author={Podell, Dustin and English, Zion and Lacey, Kyle and Blattmann, Andreas and Dockhorn, Tim and M{\"u}ller, Jonas and Penna, Joe and Rombach, Robin},
  journal={arXiv preprint arXiv:2307.01952},
  year={2023}
}

@article{nichol2021glide,
  title={Glide: Towards photorealistic image generation and editing with text-guided diffusion models},
  author={Nichol, Alex and Dhariwal, Prafulla and Ramesh, Aditya and Shyam, Pranav and Mishkin, Pamela and McGrew, Bob and Sutskever, Ilya and Chen, Mark},
  journal={arXiv preprint arXiv:2112.10741},
  year={2021}
}

@article{vaswani2017attention,
  title={Attention is all you need},
  author={Vaswani, Ashish and Shazeer, Noam and Parmar, Niki and Uszkoreit, Jakob and Jones, Llion and Gomez, Aidan N and Kaiser, {\L}ukasz and Polosukhin, Illia},
  journal={NIPS},
  volume={30},
  year={2017}
}

@inproceedings{ramesh2021zero,
  title={Zero-shot text-to-image generation},
  author={Ramesh, Aditya and Pavlov, Mikhail and Goh, Gabriel and Gray, Scott and Voss, Chelsea and Radford, Alec and Chen, Mark and Sutskever, Ilya},
  booktitle={ICML},
  pages={8821--8831},
  year={2021},
  organization={PMLR}
}

@article{schuhmann2021laion,
  title={Laion-400m: Open dataset of clip-filtered 400 million image-text pairs},
  author={Schuhmann, Christoph and Vencu, Richard and Beaumont, Romain and Kaczmarczyk, Robert and Mullis, Clayton and Katta, Aarush and Coombes, Theo and Jitsev, Jenia and Komatsuzaki, Aran},
  journal={arXiv preprint arXiv:2111.02114},
  year={2021}
}

@article{schuhmann2022laion,
  title={Laion-5b: An open large-scale dataset for training next generation image-text models},
  author={Schuhmann, Christoph and Beaumont, Romain and Vencu, Richard and Gordon, Cade and Wightman, Ross and Cherti, Mehdi and Coombes, Theo and Katta, Aarush and Mullis, Clayton and Wortsman, Mitchell and others},
  journal={NIPS},
  volume={35},
  pages={25278--25294},
  year={2022}
}

@inproceedings{ram2025dreamblend,
  title={DreamBlend: Advancing Personalized Fine-Tuning of Text-to-Image Diffusion Models},
  author={Ram, Shwetha and Neiman, Tal and Feng, Qianli and Stuart, Andrew and Tran, Son and Chilimbi, Trishul},
  booktitle={WACV},
  pages={3614--3623},
  year={2025},
  organization={IEEE}
}

@inproceedings{arar2024palp,
  title={Palp: Prompt aligned personalization of text-to-image models},
  author={Arar, Moab and Voynov, Andrey and Hertz, Amir and Avrahami, Omri and Fruchter, Shlomi and Pritch, Yael and Cohen-Or, Daniel and Shamir, Ariel},
  booktitle={SIGGRAPH Asia 2024 Conference Papers},
  pages={1--11},
  year={2024}
}

@inproceedings{pang2024cross,
title={Cross initialization for face personalization of text-to-image models},
author={Pang, Lianyu and Yin, Jian and Xie, Haoran and Wang, Qiping and Li, Qing and Mao, Xudong},
booktitle={CVPR},
pages={8393--8403},
year={2024}
}

@article{zhou2024magictailor,
  title={Magictailor: Component-controllable personalization in text-to-image diffusion models},
  author={Zhou, Donghao and Huang, Jiancheng and Bai, Jinbin and Wang, Jiaze and Chen, Hao and Chen, Guangyong and Hu, Xiaowei and Heng, Pheng-Ann},
  journal={arXiv preprint arXiv:2410.13370},
  year={2024}
}

@inproceedings{ham2024personalized,
title={Personalized residuals for concept-driven text-to-image generation},
author={Ham, Cusuh and Fisher, Matthew and Hays, James and Kolkin, Nicholas and Liu, Yuchen and Zhang, Richard and Hinz, Tobias},
booktitle={CVPR},
pages={8186--8195},
year={2024}
}

@inproceedings{ng2024partcraft,
  title={Partcraft: Crafting creative objects by parts},
  author={Ng, Kam Woh and Zhu, Xiatian and Song, Yi-Zhe and Xiang, Tao},
  booktitle={ECCV},
  pages={420--437},
  year={2024},
  organization={Springer}
}

@inproceedings{zhu2025multibooth,
  title={Multibooth: Towards generating all your concepts in an image from text},
  author={Zhu, Chenyang and Li, Kai and Ma, Yue and He, Chunming and Li, Xiu},
  booktitle={AAAI},
  volume={39},
  number={10},
  pages={10923--10931},
  year={2025}
}

@inproceedings{
  song2021scorebased,
  title={Score-Based Generative Modeling through Stochastic Differential Equations},
  author={Yang Song and Jascha Sohl-Dickstein and Diederik P Kingma and Abhishek Kumar and Stefano Ermon and Ben Poole},
  booktitle={International Conference on Learning Representations},
  year={2021},
  url={https://openreview.net/forum?id=PxTIG12RRHS}
}

@article{dhariwal2021diffusion,
  title={Diffusion models beat gans on image synthesis},
  author={Dhariwal, Prafulla and Nichol, Alexander},
  journal={NIPS},
  volume={34},
  pages={8780--8794},
  year={2021}
}

@inproceedings{marjit2024diffusekrona,
title={DiffuseKronA: A Parameter Efficient Fine-tuning Method for Personalized Diffusion Models},
author={Marjit, Shyam and Singh, Harshit and Mathur, Nityanand and Paul, Sayak and Yu, Chia-Mu and Chen, Pin-Yu},
booktitle={WACV},
pages={3529--3538},
year={2025},
organization={IEEE}
}

@inproceedings{jones2024customizing,
  title={Customizing text-to-image models with a single image pair},
  author={Jones, Maxwell and Wang, Sheng-Yu and Kumari, Nupur and Bau, David and Zhu, Jun-Yan},
  booktitle={SIGGRAPH Asia 2024 Conference Papers},
  pages={1--13},
  year={2024}
}

@inproceedings{zeng2024infusion,
  title={Infusion: Preventing customized text-to-image diffusion from overfitting},
  author={Zeng, Weili and Yan, Yichao and Zhu, Qi and Chen, Zhuo and Chu, Pengzhi and Zhao, Weiming and Yang, Xiaokang},
  booktitle={ACMMM},
  pages={3568--3577},
  year={2024}
}

@inproceedings{kong2024omg,
  title={Omg: Occlusion-friendly personalized multi-concept generation in diffusion models},
  author={Kong, Zhe and Zhang, Yong and Yang, Tianyu and Wang, Tao and Zhang, Kaihao and Wu, Bizhu and Chen, Guanying and Liu, Wei and Luo, Wenhan},
  booktitle={ECCV},
  pages={253--270},
  year={2024},
  organization={Springer}
}

@article{yang2023diffusion,
  title={Diffusion models: A comprehensive survey of methods and applications},
  author={Yang, Ling and Zhang, Zhilong and Song, Yang and Hong, Shenda and Xu, Runsheng and Zhao, Yue and Zhang, Wentao and Cui, Bin and Yang, Ming-Hsuan},
  journal={ACM Computing Surveys},
  volume={56},
  number={4},
  pages={1--39},
  year={2023},
  publisher={ACM New York, NY, USA}
}

@article{ulhaq2022efficient,
  title={Efficient diffusion models for vision: A survey},
  author={Ulhaq, Anwaar and Akhtar, Naveed and Pogrebna, Ganna},
  journal={arXiv preprint arXiv:2210.09292},
  year={2022}
}

@article{raffel2020exploring,
  title={Exploring the limits of transfer learning with a unified text-to-text transformer},
  author={Raffel, Colin and Shazeer, Noam and Roberts, Adam and Lee, Katherine and Narang, Sharan and Matena, Michael and Zhou, Yanqi and Li, Wei and Liu, Peter J},
  journal={The Journal of Machine Learning Research},
  volume={21},
  number={1},
  pages={5485--5551},
  year={2020},
  publisher={JMLRORG}
}

@article{zhan2023multimodal,
  title={Multimodal image synthesis and editing: A survey and taxonomy},
  author={Zhan, Fangneng and Yu, Yingchen and Wu, Rongliang and Zhang, Jiahui and Lu, Shijian and Liu, Lingjie and Kortylewski, Adam and Theobalt, Christian and Xing, Eric},
  journal={IEEE Transactions on Pattern Analysis and Machine Intelligence},
  year={2023},
  publisher={IEEE}
}

@article{zhang2023text,
  title={Text-to-image diffusion model in generative ai: A survey},
  author={Zhang, Chenshuang and Zhang, Chaoning and Zhang, Mengchun and Kweon, In So},
  journal={arXiv preprint arXiv:2303.07909},
  year={2023}
}

@article{li2023generative,
  title={Generative AI meets 3D: A Survey on Text-to-3D in AIGC Era},
  author={Li, Chenghao and Zhang, Chaoning and Waghwase, Atish and Lee, Lik-Hang and Rameau, Francois and Yang, Yang and Bae, Sung-Ho and Hong, Choong Seon},
  journal={arXiv preprint arXiv:2305.06131},
  year={2023}
}

@article{xing2023survey,
  title={A survey on video diffusion models},
  author={Xing, Zhen and Feng, Qijun and Chen, Haoran and Dai, Qi and Hu, Han and Xu, Hang and Wu, Zuxuan and Jiang, Yu-Gang},
  journal={arXiv preprint arXiv:2310.10647},
  year={2023}
}

@article{croitoru2023diffusion,
  title={Diffusion models in vision: A survey},
  author={Croitoru, Florinel-Alin and Hondru, Vlad and Ionescu, Radu Tudor and Shah, Mubarak},
  journal={IEEE Transactions on Pattern Analysis and Machine Intelligence},
  year={2023},
  publisher={IEEE}
}

@inproceedings{he2016deep,
  title={Deep residual learning for image recognition},
  author={He, Kaiming and Zhang, Xiangyu and Ren, Shaoqing and Sun, Jian},
  booktitle={CVPR},
  pages={770--778},
  year={2016}
}

@inproceedings{sun2023generative1,
  title={Generative Multimodal Models are In-Context Learners},
  author={Sun, Quan and Cui, Yufeng and Zhang, Xiaosong and Zhang, Fan and Yu, Qiying and Wang, Yueze and Rao, Yongming and Liu, Jingjing and Huang, Tiejun and Wang, Xinlong},
  booktitle={CVPR},
  pages={14398--14409},
  year={2024}
}

@inproceedings{song2020denoising,
  title={Denoising Diffusion Implicit Models},
  author={Song, Jiaming and Meng, Chenlin and Ermon, Stefano},
  booktitle={International Conference on Learning Representations},
  year={2020}
}

@inproceedings{howard2019searching,
  title={Searching for mobilenetv3},
  author={Howard, Andrew and Sandler, Mark and Chu, Grace and Chen, Liang-Chieh and Chen, Bo and Tan, Mingxing and Wang, Weijun and Zhu, Yukun and Pang, Ruoming and Vasudevan, Vijay and others},
  booktitle={ICCV},
  pages={1314--1324},
  year={2019}
}

@article{zhang2016joint,
  title={Joint face detection and alignment using multitask cascaded convolutional networks},
  author={Zhang, Kaipeng and Zhang, Zhanpeng and Li, Zhifeng and Qiao, Yu},
  journal={IEEE signal processing letters},
  volume={23},
  number={10},
  pages={1499--1503},
  year={2016},
  publisher={IEEE}
}

@inproceedings{schroff2015facenet,
  title={Facenet: A unified embedding for face recognition and clustering},
  author={Schroff, Florian and Kalenichenko, Dmitry and Philbin, James},
  booktitle={CVPR},
  pages={815--823},
  year={2015}
}

@article{ren2015faster,
  title={Faster r-cnn: Towards real-time object detection with region proposal networks},
  author={Ren, Shaoqing and He, Kaiming and Girshick, Ross and Sun, Jian},
  journal={NIPS},
  volume={28},
  year={2015}
}

@inproceedings{deng2009imagenet,
  title={Imagenet: A large-scale hierarchical image database},
  author={Deng, Jia and Dong, Wei and Socher, Richard and Li, Li-Jia and Li, Kai and Fei-Fei, Li},
  booktitle={CVPR},
  pages={248--255},
  year={2009},
  organization={Ieee}
}

@article{hu2021lora,
  title={Lora: Low-rank adaptation of large language models},
  author={Hu, Edward J and Shen, Yelong and Wallis, Phillip and Allen-Zhu, Zeyuan and Li, Yuanzhi and Wang, Shean and Wang, Lu and Chen, Weizhu},
  journal={arXiv preprint arXiv:2106.09685},
  year={2021}
}

@article{valipour2022dylora,
  title={Dylora: Parameter efficient tuning of pre-trained models using dynamic search-free low-rank adaptation},
  author={Valipour, Mojtaba and Rezagholizadeh, Mehdi and Kobyzev, Ivan and Ghodsi, Ali},
  journal={arXiv preprint arXiv:2210.07558},
  year={2022}
}

@article{chavan2023one,
  title={One-for-All: Generalized LoRA for Parameter-Efficient Fine-tuning},
  author={Chavan, Arnav and Liu, Zhuang and Gupta, Deepak and Xing, Eric and Shen, Zhiqiang},
  journal={arXiv preprint arXiv:2306.07967},
  year={2023}
}

@inproceedings{houlsby2019parameter,
  title={Parameter-efficient transfer learning for NLP},
  author={Houlsby, Neil and Giurgiu, Andrei and Jastrzebski, Stanislaw and Morrone, Bruna and De Laroussilhe, Quentin and Gesmundo, Andrea and Attariyan, Mona and Gelly, Sylvain},
  booktitle={ICML},
  pages={2790--2799},
  year={2019},
  organization={PMLR}
}

@inproceedings{yeh2023navigating,
  title={Navigating Text-To-Image Customization: From LyCORIS Fine-Tuning to Model Evaluation},
  author={Yeh, Shih-Ying and Hsieh, Yu-Guan and Gao, Zhidong and Yang, Bernard BW and Oh, Giyeong and Gong, Yanmin},
  booktitle={ICLR},
  year={2023}
}

@inproceedings{abbeel2004apprenticeship,
  title={Apprenticeship learning via inverse reinforcement learning},
  author={Abbeel, Pieter and Ng, Andrew Y},
  booktitle={ICML},
  pages={1},
  year={2004}
}

@inproceedings{radford2021learning,
  title={Learning transferable visual models from natural language supervision},
  author={Radford, Alec and Kim, Jong Wook and Hallacy, Chris and Ramesh, Aditya and Goh, Gabriel and Agarwal, Sandhini and Sastry, Girish and Askell, Amanda and Mishkin, Pamela and Clark, Jack and others},
  booktitle={ICML},
  pages={8748--8763},
  year={2021},
  organization={PMLR}
}

@misc{stabilityai2023deepfloydif,
  title        = {DeepFloyd IF: A Modular Cascaded Text-to-Image Model},
  author       = {{Stability AI}},
  year         = {2023},
  howpublished = {\url{https://stability.ai/news/deepfloyd-if-text-to-image-model}},
  note         = {Accessed: 2025-09-29}
}

@inproceedings{esser2024scaling,
  title={Scaling rectified flow transformers for high-resolution image synthesis},
  author={Esser, Patrick and Kulal, Sumith and Blattmann, Andreas and Entezari, Rahim and M{\"u}ller, Jonas and Saini, Harry and Levi, Yam and Lorenz, Dominik and Sauer, Axel and Boesel, Frederic and others},
  booktitle={Forty-first ICML},
  year={2024}
}

@article{chen2023pixart,
  title={Pixart-$alpha $: Fast training of diffusion transformer for photorealistic text-to-image synthesis},
  author={Chen, Junsong and Yu, Jincheng and Ge, Chongjian and Yao, Lewei and Xie, Enze and Wu, Yue and Wang, Zhongdao and Kwok, James and Luo, Ping and Lu, Huchuan and others},
  journal={arXiv preprint arXiv:2310.00426},
  year={2023}
}

@inproceedings{peebles2023scalable,
  title={Scalable diffusion models with transformers},
  author={Peebles, William and Xie, Saining},
  booktitle={ICCV},
  pages={4195--4205},
  year={2023}
}

@inproceedings{Li2023BLIP2BL,
  title={BLIP-2: Bootstrapping Language-Image Pre-training with Frozen Image Encoders and Large Language Models},
  author={Junnan Li and Dongxu Li and Silvio Savarese and Steven C. H. Hoi},
  booktitle={ICML},
  year={2023},
  url={https://api.semanticscholar.org/CorpusID:256390509}
}

@inproceedings{zhao2025local,
  title={Local conditional controlling for text-to-image diffusion models},
  author={Zhao, Yibo and Peng, Liang and Yang, Yang and Luo, Zekai and Li, Hengjia and Chen, Yao and Yang, Zheng and He, Xiaofei and Zhao, Wei and Lu, Qinglin and others},
  booktitle={AAAI},
  volume={39},
  number={10},
  pages={10492--10500},
  year={2025}
}

@inproceedings{dosovitskiy2020image,
  title={An Image is Worth 16x16 Words: Transformers for Image Recognition at Scale},
  author={Dosovitskiy, Alexey and Beyer, Lucas and Kolesnikov, Alexander and Weissenborn, Dirk and Zhai, Xiaohua and Unterthiner, Thomas and Dehghani, Mostafa and Minderer, Matthias and Heigold, Georg and Gelly, Sylvain and others},
  booktitle={International Conference on Learning Representations},
  year={2020}
}

@inproceedings{cao2018vggface2,
  title={Vggface2: A dataset for recognising faces across pose and age},
  author={Cao, Qiong and Shen, Li and Xie, Weidi and Parkhi, Omkar M and Zisserman, Andrew},
  booktitle={2018 13th IEEE international conference on automatic face \& gesture recognition (FG 2018)},
  pages={67--74},
  year={2018},
  organization={IEEE}
}

@inproceedings{szegedy2017inception,
  title={Inception-v4, inception-resnet and the impact of residual connections on learning},
  author={Szegedy, Christian and Ioffe, Sergey and Vanhoucke, Vincent and Alemi, Alexander},
  booktitle={AAAI},
  volume={31},
  number={1},
  year={2017}
}

@article{wang2023hifi,
  title={HiFi Tuner: High-Fidelity Subject-Driven Fine-Tuning for Diffusion Models},
  author={Wang, Zhonghao and Wei, Wei and Zhao, Yang and Xiao, Zhisheng and Hasegawa-Johnson, Mark and Shi, Humphrey and Hou, Tingbo},
  journal={arXiv preprint arXiv:2312.00079},
  year={2023}
}

@inproceedings{avrahami2023break,
  title={Break-A-Scene: Extracting Multiple Concepts from a Single Image},
author={Avrahami, Omri and Aberman, Kfir and Fried, Ohad and Cohen-Or, Daniel and Lischinski, Dani},
booktitle={SIGGRAPH Asia 2023 Conference Papers},
pages={1--12},
year={2023}
}

@article{wang2023context,
  title={In-context learning unlocked for diffusion models},
  author={Wang, Zhendong and Jiang, Yifan and Lu, Yadong and He, Pengcheng and Chen, Weizhu and Wang, Zhangyang and Zhou, Mingyuan and others},
  journal={NIPS},
  volume={36},
  pages={8542--8562},
  year={2023}
}

@inproceedings{wang2023high,
  title={High-fidelity Person-centric Subject-to-Image Synthesis},
  author={Wang, Yibin and Zhang, Weizhong and Zheng, Jianwei and Jin, Cheng},
  booktitle={CVPR},
  pages={7675--7684},
  year={2024}
}

@inproceedings{ruiz2023dreambooth,
  title={Dreambooth: Fine tuning text-to-image diffusion models for subject-driven generation},
  author={Ruiz, Nataniel and Li, Yuanzhen and Jampani, Varun and Pritch, Yael and Rubinstein, Michael and Aberman, Kfir},
  booktitle={CVPR},
  pages={22500--22510},
  year={2023}
}

@inproceedings{ma2023pea,
  title={PEA-Diffusion: Parameter-Efficient Adapter with Knowledge Distillation in non-English Text-to-Image Generation},
  author={Ma, Jian and Chen, Chen and Xie, Qingsong and Lu, Haonan},
  booktitle={ECCV},
  pages={89--105},
  year={2024},
  organization={Springer}
}

@article{voynov2023anylens,
  title={AnyLens: A Generative Diffusion Model with Any Rendering Lens},
  author={Voynov, Andrey and Hertz, Amir and Arar, Moab and Fruchter, Shlomi and Cohen-Or, Daniel},
  journal={CoRR},
  year={2023}
}

@inproceedings{najdenkoska2024context,
  title={Context diffusion: In-context aware image generation},
  author={Najdenkoska, Ivona and Sinha, Animesh and Dubey, Abhimanyu and Mahajan, Dhruv and Ramanathan, Vignesh and Radenovic, Filip},
  booktitle={ECCV},
  pages={375--391},
  year={2024},
  organization={Springer}
}

@inproceedings{ju2023humansd,
  title={HumanSD: A Native Skeleton-Guided Diffusion Model for Human Image Generation},
  author={Ju, Xuan and Zeng, Ailing and Zhao, Chenchen and Wang, Jianan and Zhang, Lei and Xu, Qiang},
  booktitle={ICCV},
  pages={15988--15998},
  year={2023}
}

@inproceedings{rombach2022high,
  title={High-resolution image synthesis with latent diffusion models},
  author={Rombach, Robin and Blattmann, Andreas and Lorenz, Dominik and Esser, Patrick and Ommer, Bj{\"o}rn},
  booktitle={CVPR},
  pages={10684--10695},
  year={2022}
}

@inproceedings{voynov2023sketch,
  title={Sketch-guided text-to-image diffusion models},
  author={Voynov, Andrey and Aberman, Kfir and Cohen-Or, Daniel},
  booktitle={ACM SIGGRAPH 2023 conference proceedings},
  pages={1--11},
  year={2023}
}

@article{chen2025consislora,
  title={Consislora: Enhancing content and style consistency for lora-based style transfer},
  author={Chen, Bolin and Zhao, Baoquan and Xie, Haoran and Cai, Yi and Li, Qing and Mao, Xudong},
  journal={arXiv preprint arXiv:2503.10614},
  year={2025}
}

@inproceedings{lin2024non,
  title={Non-confusing generation of customized concepts in diffusion models},
  author={Lin, Wang and Chen, Jingyuan and Shi, Jiaxin and Zhu, Yichen and Liang, Chen and Miao, Junzhong and Jin, Tao and Zhao, Zhou and Wu, Fei and Yan, Shuicheng and others},
  booktitle={ICML},
  pages={29935--29948},
  year={2024}
}

@inproceedings{nair2024maxfusion,
  title={Maxfusion: Plug\&play multi-modal generation in text-to-image diffusion models},
  author={Nair, Nithin Gopalakrishnan and Valanarasu, Jeya Maria Jose and Patel, Vishal M},
  booktitle={ECCV},
  pages={93--110},
  year={2024},
  organization={Springer}
}

@article{jang2024identity,
  title={Identity decoupling for multi-subject personalization of text-to-image models},
  author={Jang, Sangwon and Jo, Jaehyeong and Lee, Kimin and Hwang, Sung Ju},
  journal={NIPS},
  volume={37},
  pages={100895--100937},
  year={2024}
}

@inproceedings{guo2025conceptguard,
title={ConceptGuard: Continual personalized text-to-image generation with forgetting and confusion mitigation},
author={Guo, Zirun and Jin, Tao},
booktitle={CVPR},
pages={2945--2954},
year={2025}
}

@article{liu2025museummaker,
  title={MuseumMaker: Continual Style Customization without Catastrophic Forgetting Supplementary Material},
  author={Liu, Chenxi and Sun, Gan and Liang, Wenqi and Dong, Jiahua and Qin, Can and Cong, Yang},
  journal={IEEE Transactions on Image Processing},
  year={2025},
  publisher={IEEE}
}

@inproceedings{kwon2024concept,
  title={Concept weaver: Enabling multi-concept fusion in text-to-image models},
  author={Kwon, Gihyun and Jenni, Simon and Li, Dingzeyu and Lee, Joon-Young and Ye, Jong Chul and Heilbron, Fabian Caba},
  booktitle={CVPR},
  pages={8880--8889},
  year={2024}
}

@inproceedings{lu2024coarse,
title={Coarse-to-fine latent diffusion for pose-guided person image synthesis},
author={Lu, Yanzuo and Zhang, Manlin and Ma, Andy J and Xie, Xiaohua and Lai, Jianhuang},
booktitle={CVPR},
pages={6420--6429},
year={2024}
}

@article{wang2025unicombine,
  title={Unicombine: Unified multi-conditional combination with diffusion transformer},
  author={Wang, Haoxuan and Peng, Jinlong and He, Qingdong and Yang, Hao and Jin, Ying and Wu, Jiafu and Hu, Xiaobin and Pan, Yanjie and Gan, Zhenye and Chi, Mingmin and others},
  journal={arXiv preprint arXiv:2503.09277},
  year={2025}
}

@article{zhang2023motioncrafter,
  title={MotionCrafter: One-Shot Motion Customization of Diffusion Models},
  author={Zhang, Yuxin and Tang, Fan and Huang, Nisha and Huang, Haibin and Ma, Chongyang and Dong, Weiming and Xu, Changsheng},
  journal={arXiv preprint arXiv:2312.05288},
  year={2023}
}

@inproceedings{
zhao2025dreamdistribution,
title={DreamDistribution: Learning Prompt Distribution for Diverse In-distribution Generation},
author={Brian Nlong Zhao and Yuhang Xiao and Jiashu Xu and XINYANG JIANG and Yifan Yang and Dongsheng Li and Laurent Itti and Vibhav Vineet and Yunhao Ge},
booktitle={The Thirteenth International Conference on Learning Representations},
year={2025},
url={https://openreview.net/forum?id=oQoQ4u6MQC}
}

@inproceedings{xiao2023comcat,
  title={COMCAT: Towards Efficient Compression and Customization of Attention-Based Vision Models},
author={Xiao, Jinqi and Yin, Miao and Gong, Yu and Zang, Xiao and Ren, Jian and Yuan, Bo},
booktitle={ICML},
pages={38125--38136},
year={2023}
}

@article{xiang2023closer,
  title={A closer look at parameter-efficient tuning in diffusion models},
  author={Xiang, Chendong and Bao, Fan and Li, Chongxuan and Su, Hang and Zhu, Jun},
  journal={arXiv preprint arXiv:2303.18181},
  year={2023}
}

@article{chen2023subject,
  title={Subject-driven text-to-image generation via apprenticeship learning},
  author={Chen, Wenhu and Hu, Hexiang and Li, Yandong and Ruiz, Nataniel and Jia, Xuhui and Chang, Ming-Wei and Cohen, William W},
  journal={NIPS},
  volume={36},
  pages={30286--30305},
  year={2023}
}

@article{kim2023diffblender,
  title={DiffBlender: Scalable and Composable Multimodal Text-to-Image Diffusion Models},
  author={Kim, Sungnyun and Lee, Junsoo and Hong, Kibeom and Kim, Daesik and Ahn, Namhyuk},
  journal={arXiv preprint arXiv:2305.15194},
  year={2023}
}

@inproceedings{avrahami2023spatext,
  title={Spatext: Spatio-textual representation for controllable image generation},
  author={Avrahami, Omri and Hayes, Thomas and Gafni, Oran and Gupta, Sonal and Taigman, Yaniv and Parikh, Devi and Lischinski, Dani and Fried, Ohad and Yin, Xi},
  booktitle={CVPR},
  pages={18370--18380},
  year={2023}
}

@article{qin2023unicontrol,
  title={UniControl: A Unified Diffusion Model for Controllable Visual Generation In the Wild},
author={Qin, Can and Zhang, Shu and Yu, Ning and Feng, Yihao and Yang, Xinyi and Zhou, Yingbo and Wang, Huan and Niebles, Juan Carlos and Xiong, Caiming and Savarese, Silvio and others},
journal={NIPS},
volume={36},
pages={42961--42992},
year={2023}
}

@article{chen2023photoverse,
  title={Photoverse: Tuning-free image customization with text-to-image diffusion models},
  author={Chen, Li and Zhao, Mengyi and Liu, Yiheng and Ding, Mingxu and Song, Yangyang and Wang, Shizun and Wang, Xu and Yang, Hao and Liu, Jing and Du, Kang and others},
  journal={arXiv preprint arXiv:2309.05793},
  year={2023}
}

@article{ma2023unified,
  title={Unified multi-modal latent diffusion for joint subject and text conditional image generation},
  author={Ma, Yiyang and Yang, Huan and Wang, Wenjing and Fu, Jianlong and Liu, Jiaying},
  journal={arXiv preprint arXiv:2303.09319},
  year={2023}
}

@inproceedings{ma2023subject,
  title={Subject-diffusion: Open domain personalized text-to-image generation without test-time fine-tuning},
  author={Ma, Jian and Liang, Junhao and Chen, Chen and Lu, Haonan},
  booktitle={ACM SIGGRAPH 2024 Conference Papers},
  pages={1--12},
  year={2024}
}

@inproceedings{hu2023cocktail,
  title={Cocktail: Mixing Multi-Modality Controls for Text-Conditional Image Generation},
  author={Hu, Minghui and Zheng, Jianbin and Liu, Daqing and Zheng, Chuanxia and Wang, Chaoyue and Tao, Dacheng and Cham, Tat-Jen},
  booktitle={Thirty-seventh Conference on Neural Information Processing Systems},
  year={2023}
}

@inproceedings{mou2023t2i,
  title={T2i-adapter: Learning adapters to dig out more controllable ability for text-to-image diffusion models},
  author={Mou, Chong and Wang, Xintao and Xie, Liangbin and Wu, Yanze and Zhang, Jian and Qi, Zhongang and Shan, Ying},
  booktitle={AAAI},
  volume={38},
  number={5},
  pages={4296--4304},
  year={2024}
}

@article{yang2023meta,
  title={Meta ControlNet: Enhancing Task Adaptation via Meta Learning},
  author={Yang, Junjie and Zhao, Jinze and Wang, Peihao and Wang, Zhangyang and Liang, Yingbin},
  journal={arXiv preprint arXiv:2312.01255},
  year={2023}
}

@inproceedings{shi2023instantbooth,
  title={Instantbooth: Personalized text-to-image generation without test-time finetuning},
  author={Shi, Jing and Xiong, Wei and Lin, Zhe and Jung, Hyun Joon},
  booktitle={CVPR},
  pages={8543--8552},
  year={2024}
}

@inproceedings{
liu2023hyperhuman,
title={HyperHuman: Hyper-Realistic Human Generation with Latent Structural Diffusion},
author={Xian Liu and Jian Ren and Aliaksandr Siarohin and Ivan Skorokhodov and Yanyu Li and Dahua Lin and Xihui Liu and Ziwei Liu and Sergey Tulyakov},
booktitle={ICLR},
year={2024},
url={https://openreview.net/forum?id=duyA42HlCK}
}

@article{yang2023align,
  title={Align, Adapt and Inject: Sound-guided Unified Image Generation},
  author={Yang, Yue and Zhang, Kaipeng and Ge, Yuying and Shao, Wenqi and Xue, Zeyue and Qiao, Yu and Luo, Ping},
  journal={arXiv preprint arXiv:2306.11504},
  year={2023}
}

@inproceedings{
sohn2023styledrop,
title={StyleDrop: Text-to-Image Synthesis of Any Style},
author={Kihyuk Sohn and Lu Jiang and Jarred Barber and Kimin Lee and Nataniel Ruiz and Dilip Krishnan and Huiwen Chang and Yuanzhen Li and Irfan Essa and Michael Rubinstein and Yuan Hao and Glenn Entis and Irina Blok and Daniel Castro Chin},
booktitle={NIPS},
year={2023},
url={https://openreview.net/forum?id=KoaFh16uOc}
}

@article{smith2023continual,
  title={Continual diffusion: Continual customization of text-to-image diffusion with c-lora},
  author={Smith, James Seale and Hsu, Yen-Chang and Zhang, Lingyu and Hua, Ting and Kira, Zsolt and Shen, Yilin and Jin, Hongxia},
  journal={arXiv preprint arXiv:2304.06027},
  year={2023}
}

@inproceedings{tian2023interactdiffusion,
  title={InteractDiffusion: Interaction Control in Text-to-Image Diffusion Models},
  author={Hoe, Jiun Tian and Jiang, Xudong and Chan, Chee Seng and Tan, Yap-Peng and Hu, Weipeng},
  booktitle={CVPR},
  pages={6180--6189},
  year={2024}
}

@inproceedings{zheng2023layoutdiffusion,
  title={LayoutDiffusion: Controllable Diffusion Model for Layout-to-image Generation},
  author={Zheng, Guangcong and Zhou, Xianpan and Li, Xuewei and Qi, Zhongang and Shan, Ying and Li, Xi},
  booktitle={CVPR},
  pages={22490--22499},
  year={2023}
}

@article{xiao2023ccm,
  title={CCM: Adding Conditional Controls to Text-to-Image Consistency Models},
  author={Xiao, Jie and Zhu, Kai and Zhang, Han and Liu, Zhiheng and Shen, Yujun and Liu, Yu and Fu, Xueyang and Zha, Zheng-Jun},
  journal={arXiv preprint arXiv:2312.06971},
  year={2023}
}

@inproceedings{bhat2023loosecontrol,
  title={LooseControl: Lifting ControlNet for Generalized Depth Conditioning},
  author={Bhat, Shariq Farooq and Mitra, Niloy and Wonka, Peter},
  booktitle={ACM SIGGRAPH 2024 Conference Papers},
  pages={1--11},
  year={2024}
}

@inproceedings{chen2023artadapter,
  title={ArtAdapter: Text-to-Image Style Transfer using Multi-Level Style Encoder and Explicit Adaptation},
  author={Chen, Dar-Yen and Tennent, Hamish and Hsu, Ching-Wen},
  booktitle={CVPR},
  pages={8619--8628},
  year={2024}
}

@inproceedings{valevski2023face0,
  title={Face0: Instantaneously conditioning a text-to-image model on a face},
  author={Valevski, Dani and Lumen, Danny and Matias, Yossi and Leviathan, Yaniv},
  booktitle={SIGGRAPH Asia 2023 Conference Papers},
  pages={1--10},
  year={2023}
}

@inproceedings{kumari2023multi,
  title={Multi-concept customization of text-to-image diffusion},
  author={Kumari, Nupur and Zhang, Bingliang and Zhang, Richard and Shechtman, Eli and Zhu, Jun-Yan},
  booktitle={CVPR},
  pages={1931--1941},
  year={2023}
}

@inproceedings{li2023gligen,
  title={Gligen: Open-set grounded text-to-image generation},
  author={Li, Yuheng and Liu, Haotian and Wu, Qingyang and Mu, Fangzhou and Yang, Jianwei and Gao, Jianfeng and Li, Chunyuan and Lee, Yong Jae},
  booktitle={CVPR},
  pages={22511--22521},
  year={2023}
}

@InProceedings{Wang2024TokenCompose,
author    = {Wang, Zirui and Sha, Zhizhou and Ding, Zheng and Wang, Yilin and Tu, Zhuowen},
title     = {TokenCompose: Text-to-Image Diffusion with Token-level Supervision},
booktitle = {CVPR},
month     = {June},
year      = {2024},
pages     = {8553-8564}
}

@inproceedings{jin2024image,
  title={An image is worth multiple words: Discovering object level concepts using multi-concept prompt learning},
  author={Jin, Chen and Tanno, Ryutaro and Saseendran, Amrutha and Diethe, Tom and Teare, Philip Alexander},
  booktitle={Forty-first ICML},
  year={2024}
}

@article{jiang2024comat,
  title={Comat: Aligning text-to-image diffusion model with image-to-text concept matching},
  author={Jiang, Dongzhi and Song, Guanglu and Wu, Xiaoshi and Zhang, Renrui and Shen, Dazhong and Zong, Zhuofan and Liu, Yu and Li, Hongsheng},
  journal={NIPS},
  volume={37},
  pages={76177--76209},
  year={2024}
}

@article{zhao2025catversion,
  title={Catversion: Concatenating embeddings for diffusion-based text-to-image personalization},
  author={Zhao, Ruoyu and Zhu, Mingrui and Dong, Shiyin and Cheng, De and Wang, Nannan and Gao, Xinbo},
  journal={IEEE Transactions on Circuits and Systems for Video Technology},
  year={2025},
  publisher={IEEE}
}

@inproceedings{zhang2024generative,
  title={Generative active learning for image synthesis personalization},
  author={Zhang, Xulu and Zhang, Wengyu and Wei, Xiaoyong and Wu, Jinlin and Zhang, Zhaoxiang and Lei, Zhen and Li, Qing},
  booktitle={ACMMM},
  pages={10669--10677},
  year={2024}
}

@inproceedings{parihar2024precisecontrol,
title={Precisecontrol: Enhancing text-to-image diffusion models with fine-grained attribute control},
author={Parihar, Rishubh and Sachidanand, VS and Mani, Sabariswaran and Karmali, Tejan and Venkatesh Babu, R},
booktitle={ECCV},
pages={469--487},
year={2024},
organization={Springer}
}

@article{voynov2023p+,
  title={$ P+ $: Extended Textual Conditioning in Text-to-Image Generation},
  author={Voynov, Andrey and Chu, Qinghao and Cohen-Or, Daniel and Aberman, Kfir},
  journal={arXiv preprint arXiv:2303.09522},
  year={2023}
}

@inproceedings{chen2024textdiffuser,
  title={Textdiffuser-2: Unleashing the power of language models for text rendering},
  author={Chen, Jingye and Huang, Yupan and Lv, Tengchao and Cui, Lei and Chen, Qifeng and Wei, Furu},
  booktitle={ECCV},
  pages={386--402},
  year={2024},
  organization={Springer}
}

@inproceedings{wei2023elite,
  title={Elite: Encoding visual concepts into textual embeddings for customized text-to-image generation},
  author={Wei, Yuxiang and Zhang, Yabo and Ji, Zhilong and Bai, Jinfeng and Zhang, Lei and Zuo, Wangmeng},
  booktitle={ICCV},
  pages={15943--15953},
  year={2023}
}

@inproceedings{peng2024portraitbooth,
  title={Portraitbooth: A versatile portrait model for fast identity-preserved personalization},
  author={Peng, Xu and Zhu, Junwei and Jiang, Boyuan and Tai, Ying and Luo, Donghao and Zhang, Jiangning and Lin, Wei and Jin, Taisong and Wang, Chengjie and Ji, Rongrong},
  booktitle={CVPR},
  pages={27080--27090},
  year={2024}
}

@article{sun2023create,
  title={Create your world: Lifelong text-to-image diffusion},
  author={Sun, Gan and Liang, Wenqi and Dong, Jiahua and Li, Jun and Ding, Zhengming and Cong, Yang},
  journal={IEEE Transactions on Pattern Analysis and Machine Intelligence},
  volume={46},
  number={9},
  pages={6454--6470},
  year={2024},
  publisher={IEEE}
}

@article{li2023blip,
  title={Blip-diffusion: Pre-trained subject representation for controllable text-to-image generation and editing},
author={Li, Dongxu and Li, Junnan and Hoi, Steven},
journal={NIPS},
volume={36},
pages={30146--30166},
year={2023}
}

@article{wang2024high,
  title={High-fidelity person-centric subject-to-image synthesis},
author={Wang, Yibin and Zhang, Weizhong and Zheng, Jianwei and Jin, Cheng},
journal={CoRR},
year={2024}
}

@inproceedings{han2024face,
title={Face-adapter for pre-trained diffusion models with fine-grained id and attribute control},
author={Han, Yue and Zhu, Junwei and He, Keke and Chen, Xu and Ge, Yanhao and Li, Wei and Li, Xiangtai and Zhang, Jiangning and Wang, Chengjie and Liu, Yong},
booktitle={ECCV},
pages={20--36},
year={2024},
organization={Springer}
}

@article{guo2024pulid,
  title={Pulid: Pure and lightning id customization via contrastive alignment},
  author={Guo, Zinan and Wu, Yanze and Zhuowei, Chen and Zhang, Peng and He, Qian and others},
  journal={NIPS},
  volume={37},
  pages={36777--36804},
  year={2024}
}

@inproceedings{wang2024moa,
  title={Moa: Mixture-of-attention for subject-context disentanglement in personalized image generation},
  author={Wang, Kuan-Chieh and Ostashev, Daniil and Fang, Yuwei and Tulyakov, Sergey and Aberman, Kfir},
  booktitle={SIGGRAPH Asia 2024 Conference Papers},
  pages={1--12},
  year={2024}
}

@inproceedings{ran2024x,
  title={X-adapter: Adding universal compatibility of plugins for upgraded diffusion model},
  author={Ran, Lingmin and Cun, Xiaodong and Liu, Jia-Wei and Zhao, Rui and Zijie, Song and Wang, Xintao and Keppo, Jussi and Shou, Mike Zheng},
  booktitle={CVPR},
  pages={8775--8784},
  year={2024}
}

@inproceedings{cao2024decreases,
  title={What decreases editing capability? domain-specific hybrid refinement for improved gan inversion},
  author={Cao, Pu and Yang, Lu and Liu, Dongxv and Yang, Xiaoya and Huang, Tianrui and Song, Qing},
  booktitle={WACV},
  pages={4240--4249},
  year={2024}
}

@inproceedings{zhang2024c3net,
  title={C3net: Compound conditioned controlnet for multimodal content generation},
  author={Zhang, Juntao and Liu, Yuehuai and Tai, Yu-Wing and Tang, Chi-Keung},
  booktitle={CVPR},
  pages={26886--26895},
  year={2024}
}

@inproceedings{chen2024artadapter,
  title={Artadapter: Text-to-image style transfer using multi-level style encoder and explicit adaptation},
  author={Chen, Dar-Yen and Tennent, Hamish and Hsu, Ching-Wen},
  booktitle={CVPR},
  pages={8619--8628},
  year={2024}
}

@inproceedings{xu2024prompt,
title={Prompt-free diffusion: Taking" text" out of text-to-image diffusion models},
author={Xu, Xingqian and Guo, Jiayi and Wang, Zhangyang and Huang, Gao and Essa, Irfan and Shi, Humphrey},
booktitle={CVPR},
pages={8682--8692},
year={2024}
}

@article{zhao2023uni,
  title={Uni-ControlNet: All-in-One Control to Text-to-Image Diffusion Models},
author={Zhao, Shihao and Chen, Dongdong and Chen, Yen-Chun and Bao, Jianmin and Hao, Shaozhe and Yuan, Lu and Wong, Kwan-Yee K},
journal={NIPS},
volume={36},
pages={11127--11150},
year={2023}
}

@article{chen2023integrating,
  title={Integrating Geometric Control into Text-to-Image Diffusion Models for High-Quality Detection Data Generation via Text Prompt},
  author={Chen, Kai and Xie, Enze and Chen, Zhe and Wang, Yibo and Hong, Lanqing and Li, Zhenguo and Yeung, Dit-Yan},
  journal={arXiv preprint arXiv:2306.04607},
  year={2023}
}

@inproceedings{tan2024empirical,
  title={An empirical study and analysis of text-to-image generation using large language model-powered textual representation},
  author={Tan, Zhiyu and Yang, Mengping and Qin, Luozheng and Yang, Hao and Qian, Ye and Zhou, Qiang and Zhang, Cheng and Li, Hao},
  booktitle={ECCV},
  pages={472--489},
  year={2024},
  organization={Springer}
}

@inproceedings{hyung2023magicapture,
  title={Magicapture: High-resolution multi-concept portrait customization},
  author={Hyung, Junha and Shin, Jaeyo and Choo, Jaegul},
  booktitle={AAAI},
  volume={38},
  number={3},
  pages={2445--2453},
  year={2024}
}

@article{guo2023animatediff,
  title={Animatediff: Animate your personalized text-to-image diffusion models without specific tuning},
  author={Guo, Yuwei and Yang, Ceyuan and Rao, Anyi and Liang, Zhengyang and Wang, Yaohui and Qiao, Yu and Agrawala, Maneesh and Lin, Dahua and Dai, Bo},
  journal={arXiv preprint arXiv:2307.04725},
  year={2023}
}

@article{liu2023cones,
  title={Cones: Concept neurons in diffusion models for customized generation},
  author={Liu, Zhiheng and Feng, Ruili and Zhu, Kai and Zhang, Yifei and Zheng, Kecheng and Liu, Yu and Zhao, Deli and Zhou, Jingren and Cao, Yang},
  journal={arXiv preprint arXiv:2303.05125},
  year={2023}
}

@inproceedings{shah2023ziplora,
  title={ZipLoRA: Any Subject in Any Style by Effectively Merging LoRAs},
  author={Shah, Viraj and Ruiz, Nataniel and Cole, Forrester and Lu, Erika and Lazebnik, Svetlana and Li, Yuanzhen and Jampani, Varun},
  booktitle={ECCV},
  pages={422--438},
  year={2024},
  organization={Springer}
}

@inproceedings{chen2023tailored,
  title={Tailored Visions: Enhancing Text-to-Image Generation with Personalized Prompt Rewriting},
  author={Chen, Zijie and Zhang, Lichao and Weng, Fangsheng and Pan, Lili and Lan, Zhenzhong},
  booktitle={CVPR},
  pages={7727--7736},
  year={2024}
}

@inproceedings{ham2023modulating,
  title={Modulating pretrained diffusion models for multimodal image synthesis},
  author={Ham, Cusuh and Hays, James and Lu, Jingwan and Singh, Krishna Kumar and Zhang, Zhifei and Hinz, Tobias},
  booktitle={ACM SIGGRAPH 2023 Conference Proceedings},
  pages={1--11},
  year={2023}
}

@inproceedings{wang2023decompose,
  title={Decompose and Realign: Tackling Condition Misalignment in Text-to-Image Diffusion Models},
author={Wang, Luozhou and Shen, Guibao and Ge, Wenhang and Chen, Guangyong and Li, Yijun and Chen, Yingcong},
booktitle={ECCV},
pages={21--37},
year={2024},
organization={Springer}
}

@article{ye2023ip,
  title={Ip-adapter: Text compatible image prompt adapter for text-to-image diffusion models},
  author={Ye, Hu and Zhang, Jun and Liu, Sibo and Han, Xiao and Yang, Wei},
  journal={arXiv preprint arXiv:2308.06721},
  year={2023}
}

@article{giambi2023conditioning,
  title={Conditioning Diffusion Models via Attributes and Semantic Masks for Face Generation},
  author={Giambi, Nico and Lisanti, Giuseppe},
  journal={arXiv preprint arXiv:2306.00914},
  year={2023}
}

@article{goodfellow2014generative,
  title={Generative adversarial nets},
  author={Goodfellow, Ian and Pouget-Abadie, Jean and Mirza, Mehdi and Xu, Bing and Warde-Farley, David and Ozair, Sherjil and Courville, Aaron and Bengio, Yoshua},
  journal={NIPS},
  volume={27},
  year={2014}
}

@inproceedings{karras2019style,
  title={A style-based generator architecture for generative adversarial networks},
  author={Karras, Tero and Laine, Samuli and Aila, Timo},
  booktitle={CVPR},
  pages={4401--4410},
  year={2019}
}

@inproceedings{karras2020analyzing,
  title={Analyzing and improving the image quality of stylegan},
  author={Karras, Tero and Laine, Samuli and Aittala, Miika and Hellsten, Janne and Lehtinen, Jaakko and Aila, Timo},
  booktitle={CVPR},
  pages={8110--8119},
  year={2020}
}

@inproceedings{zhang2024taming,
title={Taming stable diffusion for text to 360 panorama image generation},
author={Zhang, Cheng and Wu, Qianyi and Gambardella, Camilo Cruz and Huang, Xiaoshui and Phung, Dinh and Ouyang, Wanli and Cai, Jianfei},
booktitle={CVPR},
pages={6347--6357},
year={2024}
}

@inproceedings{kumari2024customizing,
  title={Customizing text-to-image diffusion with object viewpoint control},
  author={Kumari, Nupur and Su, Grace and Zhang, Richard and Park, Taesung and Shechtman, Eli and Zhu, Jun-Yan},
  booktitle={SIGGRAPH Asia 2024 Conference Papers},
  pages={1--13},
  year={2024}
}

@inproceedings{bernal2025precisecam,
title={PreciseCam: Precise Camera Control for Text-to-Image Generation},
author={Bernal-Berdun, Edurne and Serrano, Ana and Masia, Belen and Gadelha, Matheus and Hold-Geoffroy, Yannick and Sun, Xin and Gutierrez, Diego},
booktitle={CVPR},
pages={2724--2733},
year={2025}
}

@article{seo2024genwarp,
  title={Genwarp: Single image to novel views with semantic-preserving generative warping},
  author={Seo, Junyoung and Fukuda, Kazumi and Shibuya, Takashi and Narihira, Takuya and Murata, Naoki and Hu, Shoukang and Lai, Chieh-Hsin and Kim, Seungryong and Mitsufuji, Yuki},
  journal={NIPS},
  volume={37},
  pages={80220--80243},
  year={2024}
}

@inproceedings{bai2024integrating,
title={Integrating view conditions for image synthesis},
author={Bai, Jinbin and Dong, Zhen and Feng, Aosong and Zhang, Xiao and Ye, Tian and Zhou, Kaicheng},
booktitle={IJCAI},
pages={7591--7599},
year={2024}
}

@inproceedings{feng2025simplifying,
title={Simplifying Control Mechanism in Text-to-Image Diffusion Models},
author={Feng, Zhida and Chen, Li and Sun, Yuenan and Liu, Jiaxiang and Feng, Shikun},
booktitle={AAAI},
volume={39},
number={3},
pages={3013--3021},
year={2025}
}

@inproceedings{linctrl,
title={Ctrl-Adapter: An Efficient and Versatile Framework for Adapting Diverse Controls to Any Diffusion Model},
author={Lin, Han and Cho, Jaemin and Zala, Abhay and Bansal, Mohit},
booktitle={The Thirteenth International Conference on Learning Representations}
}

@article{yuan2023customnet,
  title={Customnet: Zero-shot object customization with variable-viewpoints in text-to-image diffusion models},
  author={Yuan, Ziyang and Cao, Mingdeng and Wang, Xintao and Qi, Zhongang and Yuan, Chun and Shan, Ying},
  journal={arXiv preprint arXiv:2310.19784},
  year={2023}
}

@inproceedings{li2024controlnet++,
  title={Controlnet++: Improving conditional controls with efficient consistency feedback: Project page: liming-ai. github. io/controlnet\_plus\_plus},
  author={Li, Ming and Yang, Taojiannan and Kuang, Huafeng and Wu, Jie and Wang, Zhaoning and Xiao, Xuefeng and Chen, Chen},
  booktitle={ECCV},
  pages={129--147},
  year={2024},
  organization={Springer}
}

@inproceedings{wang2024customizing,
  title={Customizing 360-degree panoramas through text-to-image diffusion models},
  author={Wang, Hai and Xiang, Xiaoyu and Fan, Yuchen and Xue, Jing-Hao},
  booktitle={WACV},
  pages={4933--4943},
  year={2024}
}

@inproceedings{koley2024s,
  title={It's All About Your Sketch: Democratising Sketch Control in Diffusion Models},
  author={Koley, Subhadeep and Bhunia, Ayan Kumar and Sekhri, Deeptanshu and Sain, Aneeshan and Chowdhury, Pinaki Nath and Xiang, Tao and Song, Yi-Zhe},
  booktitle={CVPR},
  pages={7204--7214},
  year={2024}
}

@inproceedings{li2024adversarial,
  title={Adversarial supervision makes layout-to-image diffusion models thrive},
  author={Li, Yumeng and Keuper, Margret and Zhang, Dan and Khoreva, Anna},
  booktitle={ICLR},
  year={2024}
}

@inproceedings{liu2024smartcontrol,
  title={Smartcontrol: Enhancing controlnet for handling rough visual conditions},
  author={Liu, Xiaoyu and Wei, Yuxiang and Liu, Ming and Lin, Xianhui and Ren, Peiran and Xie, Xuansong and Zuo, Wangmeng},
  booktitle={ECCV},
  pages={1--17},
  year={2024},
  organization={Springer}
}

@inproceedings{lv2024place,
  title={Place: Adaptive layout-semantic fusion for semantic image synthesis},
  author={Lv, Zhengyao and Wei, Yuxiang and Zuo, Wangmeng and Wong, Kwan-Yee K},
  booktitle={CVPR},
  pages={9264--9274},
  year={2024}
}

@inproceedings{wang2024instancediffusion,
  title={Instancediffusion: Instance-level control for image generation},
  author={Wang, Xudong and Darrell, Trevor and Rambhatla, Sai Saketh and Girdhar, Rohit and Misra, Ishan},
  booktitle={CVPR},
  pages={6232--6242},
  year={2024}
}

@inproceedings{cui2024idadapter,
title={Idadapter: Learning mixed features for tuning-free personalization of text-to-image models},
author={Cui, Siying and Guo, Jia and An, Xiang and Deng, Jiankang and Zhao, Yongle and Wei, Xinyu and Feng, Ziyong},
booktitle={CVPR},
pages={950--959},
year={2024}
}

@inproceedings{shiohara2024face2diffusion,
title={Face2diffusion for fast and editable face personalization},
author={Shiohara, Kaede and Yamasaki, Toshihiko},
booktitle={CVPR},
pages={6850--6859},
year={2024}
}

@inproceedings{liang2024caphuman,
  title={Caphuman: Capture your moments in parallel universes},
  author={Liang, Chao and Ma, Fan and Zhu, Linchao and Deng, Yingying and Yang, Yi},
  booktitle={CVPR},
  pages={6400--6409},
  year={2024}
}

@inproceedings{wu2024relation,
  title={Relation rectification in diffusion model},
  author={Wu, Yinwei and Yang, Xingyi and Wang, Xinchao},
  booktitle={CVPR},
  pages={7685--7694},
  year={2024}
}

@inproceedings{qiao2024facechain,
  title={Facechain-sude: Building derived class to inherit category attributes for one-shot subject-driven generation},
  author={Qiao, Pengchong and Shang, Lei and Liu, Chang and Sun, Baigui and Ji, Xiangyang and Chen, Jie},
  booktitle={CVPR},
  pages={7215--7224},
  year={2024}
}

@inproceedings{gal2024lcm,
  title={Lcm-lookahead for encoder-based text-to-image personalization},
  author={Gal, Rinon and Lichter, Or and Richardson, Elad and Patashnik, Or and Bermano, Amit H and Chechik, Gal and Cohen-Or, Daniel},
  booktitle={ECCV},
  pages={322--340},
  year={2024},
  organization={Springer}
}

@inproceedings{song2024moma,
  title={Moma: Multimodal llm adapter for fast personalized image generation},
  author={Song, Kunpeng and Zhu, Yizhe and Liu, Bingchen and Yan, Qing and Elgammal, Ahmed and Yang, Xiao},
  booktitle={ECCV},
  pages={117--132},
  year={2024},
  organization={Springer}
}

@article{wu2023human,
  title={Human preference score v2: A solid benchmark for evaluating human preferences of text-to-image synthesis},
  author={Wu, Xiaoshi and Hao, Yiming and Sun, Keqiang and Chen, Yixiong and Zhu, Feng and Zhao, Rui and Li, Hongsheng},
  journal={arXiv preprint arXiv:2306.09341},
  year={2023}
}

@article{huang2023t2i,
  title={T2i-compbench: A comprehensive benchmark for open-world compositional text-to-image generation},
  author={Huang, Kaiyi and Sun, Kaiyue and Xie, Enze and Li, Zhenguo and Liu, Xihui},
  journal={Advances in Neural Information Processing Systems},
  volume={36},
  pages={78723--78747},
  year={2023}
}

@article{kirstain2023pick,
  title={Pick-a-pic: An open dataset of user preferences for text-to-image generation},
  author={Kirstain, Yuval and Polyak, Adam and Singer, Uriel and Matiana, Shahbuland and Penna, Joe and Levy, Omer},
  journal={Advances in neural information processing systems},
  volume={36},
  pages={36652--36663},
  year={2023}
}

@misc{schuhmann2022laionaesthetics,
  title        = {LAION-Aesthetics},
  year         = {2022},
  howpublished = {\url{https://laion.ai/blog/laion-aesthetics/}},
}

@article{ghosh2023geneval,
  title={Geneval: An object-focused framework for evaluating text-to-image alignment},
  author={Ghosh, Dhruba and Hajishirzi, Hannaneh and Schmidt, Ludwig},
  journal={Advances in Neural Information Processing Systems},
  volume={36},
  pages={52132--52152},
  year={2023}
}

@inproceedings{hessel2021clipscore,
  title={Clipscore: A reference-free evaluation metric for image captioning},
  author={Hessel, Jack and Holtzman, Ari and Forbes, Maxwell and Le Bras, Ronan and Choi, Yejin},
  booktitle={EMNLP},
  pages={7514--7528},
  year={2021}
}

@inproceedings{huang2024realcustom,
  title={Realcustom: Narrowing real text word for real-time open-domain text-to-image customization},
  author={Huang, Mengqi and Mao, Zhendong and Liu, Mingcong and He, Qian and Zhang, Yongdong},
  booktitle={CVPR},
  pages={7476--7485},
  year={2024}
}

@article{dat2025vsc,
title={VSC: Visual Search Compositional Text-to-Image Diffusion Model},
author={Dat, Do Huu and Hyeonu, Nam and Mao, Po-Yuan and Oh, Tae-Hyun},
journal={arXiv preprint arXiv:2505.01104},
year={2025}
}

@inproceedings{yang2024emogen,
  title={Emogen: Emotional image content generation with text-to-image diffusion models},
  author={Yang, Jingyuan and Feng, Jiawei and Huang, Hui},
  booktitle={CVPR},
  pages={6358--6368},
  year={2024}
}

@inproceedings{zhou2024migc,
  title={Migc: Multi-instance generation controller for text-to-image synthesis},
  author={Zhou, Dewei and Li, You and Ma, Fan and Zhang, Xiaoting and Yang, Yi},
  booktitle={CVPR},
  pages={6818--6828},
  year={2024}
}

@inproceedings{song2025harmonizing,
title={Harmonizing visual and textual embeddings for zero-shot text-to-image customization},
author={Song, Yeji and Kim, Jimyeong and Park, Wonhark and Shin, Wonsik and Rhee, Wonjong and Kwak, Nojun},
booktitle={AAAI},
volume={39},
number={19},
pages={20549--20557},
year={2025}
}

@article{karras2021alias,
  title={Alias-free generative adversarial networks},
  author={Karras, Tero and Aittala, Miika and Laine, Samuli and H{\"a}rk{\"o}nen, Erik and Hellsten, Janne and Lehtinen, Jaakko and Aila, Timo},
  journal={NIPS},
  volume={34},
  pages={852--863},
  year={2021}
}

@inproceedings{liu2024towards,
  title={Towards a simultaneous and granular identity-expression control in personalized face generation},
  author={Liu, Renshuai and Ma, Bowen and Zhang, Wei and Hu, Zhipeng and Fan, Changjie and Lv, Tangjie and Ding, Yu and Cheng, Xuan},
  booktitle={CVPR},
  pages={2114--2123},
  year={2024}
}

@inproceedings{qi2024deadiff,
title={Deadiff: An efficient stylization diffusion model with disentangled representations},
author={Qi, Tianhao and Fang, Shancheng and Wu, Yanze and Xie, Hongtao and Liu, Jiawei and Chen, Lang and He, Qian and Zhang, Yongdong},
booktitle={CVPR},
pages={8693--8702},
year={2024}
}

@inproceedings{liu2024glyph,
  title={Glyph-byt5: A customized text encoder for accurate visual text rendering},
  author={Liu, Zeyu and Liang, Weicong and Liang, Zhanhao and Luo, Chong and Li, Ji and Huang, Gao and Yuan, Yuhui},
  booktitle={ECCV},
  pages={361--377},
  year={2024},
  organization={Springer}
}

@inproceedings{lv2024pick,
  title={Pick-and-draw: Training-free semantic guidance for text-to-image personalization},
  author={Lv, Henglei and Xiao, Jiayu and Li, Liang},
  booktitle={ACMMM},
  pages={10535--10543},
  year={2024}
}

@inproceedings{zhao2025ltos,
  title={Ltos: Layout-controllable text-object synthesis via adaptive cross-attention fusions},
  author={Zhao, Xiaoran and Wu, Tianhao and Lai, Yu and Tian, Zhiliang and Huang, Zhen and Liu, Yahui and He, Zejiang and Li, Dongsheng},
  booktitle={ICASSP 2025-2025 IEEE International Conference on Acoustics, Speech and Signal Processing (ICASSP)},
  pages={1--5},
  year={2025},
  organization={IEEE}
}

@inproceedings{phung2024grounded,
  title={Grounded text-to-image synthesis with attention refocusing},
  author={Phung, Quynh and Ge, Songwei and Huang, Jia-Bin},
  booktitle={CVPR},
  pages={7932--7942},
  year={2024}
}

@article{rassin2023linguistic,
  title={Linguistic binding in diffusion models: Enhancing attribute correspondence through attention map alignment},
  author={Rassin, Royi and Hirsch, Eran and Glickman, Daniel and Ravfogel, Shauli and Goldberg, Yoav and Chechik, Gal},
  journal={NIPS},
  volume={36},
  pages={3536--3559},
  year={2023}
}

@inproceedings{ge2023expressive,
  title={Expressive text-to-image generation with rich text},
  author={Ge, Songwei and Park, Taesung and Zhu, Jun-Yan and Huang, Jia-Bin},
  booktitle={ICCV},
  pages={7545--7556},
  year={2023}
}

@article{chen2024region,
title={Region-Aware Text-to-Image Generation via Hard Binding and Soft Refinement},
author={Chen, Zhennan and Li, Yajie and Wang, Haofan and Chen, Zhibo and Jiang, Zhengkai and Li, Jun and Wang, Qian and Yang, Jian and Tai, Ying},
journal={arXiv preprint arXiv:2411.06558},
year={2024}
}

@inproceedings{dahary2024yourself,
  title={Be yourself: Bounded attention for multi-subject text-to-image generation},
  author={Dahary, Omer and Patashnik, Or and Aberman, Kfir and Cohen-Or, Daniel},
  booktitle={ECCV},
  pages={432--448},
  year={2024},
  organization={Springer}
}

@article{fan2024refdrop,
  title={Refdrop: Controllable consistency in image or video generation via reference feature guidance},
  author={Fan, Jiaojiao and Xue, Haotian and Zhang, Qinsheng and Chen, Yongxin},
  journal={NIPS},
  volume={37},
  pages={33602--33637},
  year={2024}
}

@inproceedings{li2024tuning,
  title={Tuning-free image customization with image and text guidance},
  author={Li, Pengzhi and Nie, Qiang and Chen, Ying and Jiang, Xi and Wu, Kai and Lin, Yuhuan and Liu, Yong and Peng, Jinlong and Wang, Chengjie and Zheng, Feng},
  booktitle={ECCV},
  pages={233--250},
  year={2024},
  organization={Springer}
}

@inproceedings{hertz2024style,
  title={Style aligned image generation via shared attention},
  author={Hertz, Amir and Voynov, Andrey and Fruchter, Shlomi and Cohen-Or, Daniel},
  booktitle={CVPR},
  pages={4775--4785},
  year={2024}
}

@article{liu2025training,
  title={Training-free Subject-Enhanced Attention Guidance for Compositional Text-to-image Generation},
  author={Liu, Shengyuan and Wang, Bo and Ma, Ye and Yang, Te and Chen, Quan and Dong, Di},
  journal={Pattern Recognition},
  pages={112111},
  year={2025},
  publisher={Elsevier}
}

@article{gu2024analogist,
  title={Analogist: Out-of-the-box visual in-context learning with image diffusion model},
  author={Gu, Zheng and Yang, Shiyuan and Liao, Jing and Huo, Jing and Gao, Yang},
  journal={TOG},
  volume={43},
  number={4},
  pages={1--15},
  year={2024},
  publisher={ACM New York, NY, USA}
}

@inproceedings{heaid,
title={AID: Attention Interpolation of Text-to-Image Diffusion},
author={He, Qiyuan and Wang, Jinghao and Liu, Ziwei and Yao, Angela},
booktitle={The Thirty-eighth Annual Conference on Neural Information Processing Systems}
}

@inproceedings{
feng2023trainingfree,
title={Training-Free Structured Diffusion Guidance for Compositional Text-to-Image Synthesis},
author={Weixi Feng and Xuehai He and Tsu-Jui Fu and Varun Jampani and Arjun Reddy Akula and Pradyumna Narayana and Sugato Basu and Xin Eric Wang and William Yang Wang},
booktitle={ICLR },
year={2023},
url={https://openreview.net/forum?id=PUIqjT4rzq7}
}

@inproceedings{bar2023multidiffusion,
title={MultiDiffusion: fusing diffusion paths for controlled image generation},
author={Bar-Tal, Omer and Yariv, Lior and Lipman, Yaron and Dekel, Tali},
booktitle={ICML},
pages={1737--1752},
year={2023}
}

@article{zhou2025exploring,
  title={Exploring Position Encoding in Diffusion U-Net for Training-free High-resolution Image Generation},
author={Zhou, Feng and Cao, Pu and Ma, Yiyang and Yang, Lu and Yin, Jianqin},
journal={arXiv preprint arXiv:2503.09830},
year={2025}
}

@inproceedings{si2024freeu,
title={Freeu: Free lunch in diffusion u-net},
author={Si, Chenyang and Huang, Ziqi and Jiang, Yuming and Liu, Ziwei},
booktitle={CVPR},
pages={4733--4743},
year={2024}
}

@inproceedings{wang2024compositional,
title={Compositional text-to-image synthesis with attention map control of diffusion models},
author={Wang, Ruichen and Chen, Zekang and Chen, Chen and Ma, Jian and Lu, Haonan and Lin, Xiaodong},
booktitle={AAAI},
volume={38},
number={6},
pages={5544--5552},
year={2024}
}

@inproceedings{ding2024freecustom,
title={Freecustom: Tuning-free customized image generation for multi-concept composition},
author={Ding, Ganggui and Zhao, Canyu and Wang, Wen and Yang, Zhen and Liu, Zide and Chen, Hao and Shen, Chunhua},
booktitle={CVPR},
pages={9089--9098},
year={2024}
}

@inproceedings{couairon2023zero,
  title={Zero-shot spatial layout conditioning for text-to-image diffusion models},
  author={Couairon, Guillaume and Careil, Marlene and Cord, Matthieu and Lathuiliere, St{\'e}phane and Verbeek, Jakob},
  booktitle={ICCV},
  pages={2174--2183},
  year={2023}
}

@inproceedings{xiaor,
title={R\&B: Region and Boundary Aware Zero-shot Grounded Text-to-image Generation},
author={Xiao, Jiayu and Lv, Henglei and Li, Liang and Wang, Shuhui and Huang, Qingming},
booktitle={ICLR}
}

@inproceedings{mo2024freecontrol,
  title={Freecontrol: Training-free spatial control of any text-to-image diffusion model with any condition},
  author={Mo, Sicheng and Mu, Fangzhou and Lin, Kuan Heng and Liu, Yanli and Guan, Bochen and Li, Yin and Zhou, Bolei},
  booktitle={CVPR},
  pages={7465--7475},
  year={2024}
}

@inproceedings{lee2024reground,
title={Reground: Improving textual and spatial grounding at no cost},
author={Lee, Phillip Y and Sung, Minhyuk},
booktitle={ECCV},
pages={275--292},
year={2024},
organization={Springer}
}

@inproceedings{basu2024mechanistic,
  title={On mechanistic knowledge localization in text-to-image generative models},
  author={Basu, Samyadeep and Rezaei, Keivan and Kattakinda, Priyatham and Morariu, Vlad and Zhao, Nanxuan and Rossi, R\_A and Manjunatha, Varun and Feizi, Soheil},
  year={2024},
  organization={ICML}
}

@inproceedings{guo2024initno,
  title={Initno: Boosting text-to-image diffusion models via initial noise optimization},
  author={Guo, Xiefan and Liu, Jinlin and Cui, Miaomiao and Li, Jiankai and Yang, Hongyu and Huang, Di},
  booktitle={CVPR},
  pages={9380--9389},
  year={2024}
}

@article{heusel2017gans,
  title={Gans trained by a two time-scale update rule converge to a local nash equilibrium},
  author={Heusel, Martin and Ramsauer, Hubert and Unterthiner, Thomas and Nessler, Bernhard and Hochreiter, Sepp},
  journal={NIPS},
  volume={30},
  year={2017}
}

@article{salimans2016improved,
  title={Improved techniques for training gans},
  author={Salimans, Tim and Goodfellow, Ian and Zaremba, Wojciech and Cheung, Vicki and Radford, Alec and Chen, Xi},
  journal={NIPS},
  volume={29},
  year={2016}
}

@inproceedings{szegedy2016rethinking,
  title={Rethinking the inception architecture for computer vision},
  author={Szegedy, Christian and Vanhoucke, Vincent and Ioffe, Sergey and Shlens, Jon and Wojna, Zbigniew},
  booktitle={CVPR},
  pages={2818--2826},
  year={2016}
}

@inproceedings{fei2023gradient,
  title={Gradient-Free Textual Inversion},
  author={Fei, Zhengcong and Fan, Mingyuan and Huang, Junshi},
  booktitle={ACMMM},
  pages={1364--1373},
  year={2023}
}

@inproceedings{huang2023collaborative,
  title={Collaborative diffusion for multi-modal face generation and editing},
  author={Huang, Ziqi and Chan, Kelvin CK and Jiang, Yuming and Liu, Ziwei},
  booktitle={CVPR},
  pages={6080--6090},
  year={2023}
}

@inproceedings{li2023photomaker,
  title={PhotoMaker: Customizing Realistic Human Photos via Stacked ID Embedding},
  author={Li, Zhen and Cao, Mingdeng and Wang, Xintao and Qi, Zhongang and Cheng, Ming-Ming and Shan, Ying},
  booktitle={CVPR},
  pages={8640--8650},
  year={2024}
}

@inproceedings{he2023data,
  title={A Data Perspective on Enhanced Identity Preservation for Diffusion Personalization},
author={He, Xingzhe and Cao, Zhiwen and Kolkin, Nicholas and Yu, Lantao and Wan, Kun and Rhodin, Helge and Kalarot, Ratheesh},
booktitle={WACV},
pages={3782--3791},
year={2025},
organization={IEEE}
}

@inproceedings{
zhang2023jointnet,
title={JointNet: Extending Text-to-Image Diffusion for Dense Distribution Modeling},
author={Jingyang Zhang and Shiwei Li and Yuanxun Lu and Tian Fang and David Neil McKinnon and Yanghai Tsin and Long Quan and Yao Yao},
booktitle={ICLR},
year={2024},
url={https://openreview.net/forum?id=kv5xE1p3jz}
}

@inproceedings{huang2023composer,
  title={Composer: creative and controllable image synthesis with composable conditions},
  author={Huang, Lianghua and Chen, Di and Liu, Yu and Shen, Yujun and Zhao, Deli and Zhou, Jingren},
  booktitle={ICML},
  pages={13753--13773},
  year={2023}
}

@article{achlioptas2023stellar,
  title={Stellar: Systematic Evaluation of Human-Centric Personalized Text-to-Image Methods},
  author={Achlioptas, Panos and Benetatos, Alexandros and Fostiropoulos, Iordanis and Skourtis, Dimitris},
  journal={arXiv preprint arXiv:2312.06116},
  year={2023}
}

@inproceedings{peng2023portraitbooth,
  title={PortraitBooth: A Versatile Portrait Model for Fast Identity-preserved Personalization},
  author={Peng, Xu and Zhu, Junwei and Jiang, Boyuan and Tai, Ying and Luo, Donghao and Zhang, Jiangning and Lin, Wei and Jin, Taisong and Wang, Chengjie and Ji, Rongrong},
  booktitle={CVPR},
  pages={27080--27090},
  year={2024}
}

@inproceedings{mo2023freecontrol,
  title={FreeControl: Training-Free Spatial Control of Any Text-to-Image Diffusion Model with Any Condition},
  author={Mo, Sicheng and Mu, Fangzhou and Lin, Kuan Heng and Liu, Yanli and Guan, Bochen and Li, Yin and Zhou, Bolei},
  booktitle={CVPR},
  pages={7465--7475},
  year={2024}
}

@inproceedings{han2023svdiff,
  title={Svdiff: Compact parameter space for diffusion fine-tuning},
  author={Han, Ligong and Li, Yinxiao and Zhang, Han and Milanfar, Peyman and Metaxas, Dimitris and Yang, Feng},
  booktitle={ICCV},
  pages={7323--7334},
  year={2023}
}

@inproceedings{xue2023freestyle,
  title={Freestyle Layout-to-Image Synthesis},
  author={Xue, Han and Huang, Zhiwu and Sun, Qianru and Song, Li and Zhang, Wenjun},
  booktitle={CVPR},
  pages={14256--14266},
  year={2023}
}

@article{chen2023textdiffuser,
  title={TextDiffuser: Diffusion Models as Text Painters},
author={Chen, Jingye and Huang, Yupan and Lv, Tengchao and Cui, Lei and Chen, Qifeng and Wei, Furu},
journal={NIPS},
volume={36},
pages={9353--9387},
year={2023}
}

@article{zhao2023videoassembler,
  title={VideoAssembler: Identity-Consistent Video Generation with Reference Entities using Diffusion Model},
  author={Zhao, Haoyu and Lu, Tianyi and Gu, Jiaxi and Zhang, Xing and Wu, Zuxuan and Xu, Hang and Jiang, Yu-Gang},
  journal={arXiv preprint arXiv:2311.17338},
  year={2023}
}

@article{qi2023layered,
  title={Layered Rendering Diffusion Model for Zero-Shot Guided Image Synthesis},
  author={Qi, Zipeng and Huang, Guoxi and Huang, Zebin and Guo, Qin and Chen, Jinwen and Han, Junyu and Wang, Jian and Zhang, Gang and Liu, Lufei and Ding, Errui and others},
  journal={arXiv preprint arXiv:2311.18435},
  year={2023}
}

@inproceedings{chen2023dreamidentity,
  title={Dreamidentity: Improved editability for efficient face-identity preserved image generation},
  author={Chen, Zhuowei and Fang, Shancheng and Liu, Wei and He, Qian and Huang, Mengqi and Mao, Zhendong},
  booktitle={AAAI},
  volume={38},
  number={2},
  pages={1281--1289},
  year={2024}
}

@article{zhang2023prospect,
  title={ProSpect: Prompt Spectrum for Attribute-Aware Personalization of Diffusion Models},
  author={Zhang, Yuxin and Dong, Weiming and Tang, Fan and Huang, Nisha and Huang, Haibin and Ma, Chongyang and Lee, Tong-Yee and Deussen, Oliver and Xu, Changsheng},
  journal={TOG},
  volume={42},
  number={6},
  pages={1--14},
  year={2023},
  publisher={ACM New York, NY, USA}
}

@article{qiu2023controlling,
  title={Controlling text-to-image diffusion by orthogonal finetuning},
  author={Qiu, Zeju and Liu, Weiyang and Feng, Haiwen and Xue, Yuxuan and Feng, Yao and Liu, Zhen and Zhang, Dan and Weller, Adrian and Sch{\"o}lkopf, Bernhard},
  journal={NIPS},
  volume={36},
  pages={79320--79362},
  year={2023}
}

@inproceedings{xie2023boxdiff,
  title={Boxdiff: Text-to-image synthesis with training-free box-constrained diffusion},
  author={Xie, Jinheng and Li, Yuexiang and Huang, Yawen and Liu, Haozhe and Zhang, Wentian and Zheng, Yefeng and Shou, Mike Zheng},
  booktitle={ICCV},
  pages={7452--7461},
  year={2023}
}

@inproceedings{chen2025versagen,
title={VersaGen: Unleashing Versatile Visual Control for Text-to-Image Synthesis},
author={Chen, Zhipeng and Yang, Lan and Qi, Yonggang and Zhang, Honggang and Pang, Kaiyue and Li, Ke and Song, Yi-Zhe},
booktitle={AAAI},
volume={39},
number={3},
pages={2394--2402},
year={2025}
}

@inproceedings{zhang2024object,
  title={Object-conditioned energy-based attention map alignment in text-to-image diffusion models},
  author={Zhang, Yasi and Yu, Peiyu and Wu, Ying Nian},
  booktitle={ECCV},
  pages={55--71},
  year={2024},
  organization={Springer}
}

@inproceedings{patel2025enhancing,
title={Enhancing image layout control with loss-guided diffusion models},
author={Patel, Zakaria and Serkh, Kirill},
booktitle={WACV},
pages={3916--3924},
year={2025},
organization={IEEE}
}

@inproceedings{wang2024magic,
title={Magic: Multi-modality guided image completion},
author={Wang, Hao and Yu, Yongsheng and Luo, Tiejian and Fan, Heng and Zhang, Libo},
booktitle={ICLR},
year={2024}
}

@inproceedings{liang2025vodiff,
title={VODiff: Controlling Object Visibility Order in Text-to-Image Generation},
author={Liang, Dong and Jia, Jinyuan and Liu, Yuhao and Ke, Zhanghan and Fu, Hongbo and Lau, Rynson WH},
booktitle={CVPR},
pages={18379--18389},
year={2025}
}

@article{luo2025adding,
title={Adding Additional Control to One-Step Diffusion with Joint Distribution Matching},
author={Luo, Yihong and Hu, Tianyang and Song, Yifan and Sun, Jiacheng and Li, Zhenguo and Tang, Jing},
journal={arXiv preprint arXiv:2503.06652},
year={2025}
}

@inproceedings{hertz2023style,
  title={Style Aligned Image Generation via Shared Attention},
  author={Hertz, Amir and Voynov, Andrey and Fruchter, Shlomi and Cohen-Or, Daniel},
  booktitle={CVPR},
  pages={4775--4785},
  year={2024}
}

@inproceedings{zavadski2023controlnet,
  title={ControlNet-XS: Designing an Efficient and Effective Architecture for Controlling Text-to-Image Diffusion Models},
author={Zavadski, Denis and Feiden, Johann-Friedrich and Rother, Carsten},
booktitle={ECCV},
pages={343--362},
year={2024},
organization={Springer}
}

@article{li2023stylegan,
  title={When StyleGAN Meets Stable Diffusion: a $\mathscr {W} _+ $ Adapter for Personalized Image Generation},
  author={Li, Xiaoming and Hou, Xinyu and Loy, Chen Change},
  journal={arXiv preprint arXiv:2311.17461},
  year={2023}
}

@article{liu2023late,
  title={Late-Constraint Diffusion Guidance for Controllable Image Synthesis},
  author={Liu, Chang and Liu, Dong},
  journal={arXiv preprint arXiv:2305.11520},
  year={2023}
}

@inproceedings{arar2023domain,
  title={Domain-agnostic tuning-encoder for fast personalization of text-to-image models},
  author={Arar, Moab and Gal, Rinon and Atzmon, Yuval and Chechik, Gal and Cohen-Or, Daniel and Shamir, Ariel and H. Bermano, Amit},
  booktitle={SIGGRAPH Asia 2023 Conference Papers},
  pages={1--10},
  year={2023}
}

@inproceedings{tewel2023key,
  title={Key-locked rank one editing for text-to-image personalization},
  author={Tewel, Yoad and Gal, Rinon and Chechik, Gal and Atzmon, Yuval},
  booktitle={ACM SIGGRAPH 2023 conference proceedings},
  pages={1--11},
  year={2023}
}

@article{cheng2023layoutdiffuse,
  title={Layoutdiffuse: Adapting foundational diffusion models for layout-to-image generation},
  author={Cheng, Jiaxin and Liang, Xiao and Shi, Xingjian and He, Tong and Xiao, Tianjun and Li, Mu},
  journal={arXiv preprint arXiv:2302.08908},
  year={2023}
}

@inproceedings{agarwal2023image,
  title={An Image is Worth Multiple Words: Multi-attribute Inversion for Constrained Text-to-Image Synthesis},
author={Agarwal, Aishwarya and Karanam, Srikrishna and Shukla, Tripti and Srinivasan, Balaji Vasan},
booktitle={WACV},
pages={6053--6062},
year={2025},
organization={IEEE}
}

@inproceedings{jin2023image,
  title={An image is worth multiple words: Learning object level concepts using multi-concept prompt learning},
  author={Jin, Chen and Tanno, Ryutaro and Saseendran, Amrutha and Diethe, Tom and Teare, Philip Alexander},
  booktitle={Forty-first ICML},
  year={2024}
}

@article{gal2022image,
  title={An image is worth one word: Personalizing text-to-image generation using textual inversion},
  author={Gal, Rinon and Alaluf, Yuval and Atzmon, Yuval and Patashnik, Or and Bermano, Amit H and Chechik, Gal and Cohen-Or, Daniel},
  journal={arXiv preprint arXiv:2208.01618},
  year={2022}
}

@article{motamed2023lego,
  title={Lego: Learning to Disentangle and Invert Concepts Beyond Object Appearance in Text-to-Image Diffusion Models},
  author={Motamed, Saman and Paudel, Danda Pani and Van Gool, Luc},
  journal={arXiv preprint arXiv:2311.13833},
  year={2023}
}

@article{gal2023designing,
  title={Designing an encoder for fast personalization of text-to-image models},
  author={Gal, Rinon and Arar, Moab and Atzmon, Yuval and Bermano, Amit H and Chechik, Gal and Cohen-Or, Daniel},
  journal={arXiv preprint arXiv:2302.12228},
  year={2023}
}

@article{cheong2023visconet,
  title={ViscoNet: Bridging and Harmonizing Visual and Textual Conditioning for ControlNet},
  author={Cheong, Soon Yau and Mustafa, Armin and Gilbert, Andrew},
  journal={arXiv preprint arXiv:2312.03154},
  year={2023}
}

@inproceedings{safaee2023clic,
  title={CLiC: Concept Learning in Context},
  author={Safaee, Mehdi and Mikaeili, Aryan and Patashnik, Or and Cohen-Or, Daniel and Mahdavi-Amiri, Ali},
  booktitle={CVPR},
  pages={6924--6933},
  year={2024}
}

@inproceedings{huang2023learning,
  title={Learning Disentangled Identifiers for Action-Customized Text-to-Image Generation},
  author={Huang, Siteng and Gong, Biao and Feng, Yutong and Chen, Xi and Fu, Yuqian and Liu, Yu and Wang, Donglin},
  booktitle={CVPR},
  pages={7797--7806},
  year={2024}
}

@inproceedings{jia2023ssmg,
  title={SSMG: Spatial-Semantic Map Guided Diffusion Model for Free-form Layout-to-Image Generation},
  author={Jia, Chengyou and Luo, Minnan and Dang, Zhuohang and Dai, Guang and Chang, Xiaojun and Wang, Mengmeng and Wang, Jingdong},
  booktitle={AAAI},
  volume={38},
  number={3},
  pages={2480--2488},
  year={2024}
}

@inproceedings{smith2023continual2,
  title={Continual Diffusion with STAMINA: STack-And-Mask INcremental Adapters},
  author={Smith, James Seale and Hsu, Yen-Chang and Kira, Zsolt and Shen, Yilin and Jin, Hongxia},
  booktitle={CVPR},
  pages={1744--1754},
  year={2024}
}

@inproceedings{zhou2023customization,
  title={Customization Assistant for Text-to-image Generation},
  author={Zhou, Yufan and Zhang, Ruiyi and Gu, Jiuxiang and Sun, Tong},
  booktitle={CVPR},
  pages={9182--9191},
  year={2024}
}

@article{li2023generate,
  title={Generate Anything Anywhere in Any Scene},
  author={Li, Yuheng and Liu, Haotian and Wen, Yangming and Lee, Yong Jae},
  journal={arXiv preprint arXiv:2306.17154},
  year={2023}
}

@inproceedings{jiang2023videobooth,
  title={VideoBooth: Diffusion-based Video Generation with Image Prompts},
  author={Jiang, Yuming and Wu, Tianxing and Yang, Shuai and Si, Chenyang and Lin, Dahua and Qiao, Yu and Loy, Chen Change and Liu, Ziwei},
  booktitle={CVPR},
  pages={6689--6700},
  year={2024}
}

@article{voronov2023loss,
  title={Is This Loss Informative? Faster Text-to-Image Customization by Tracking Objective Dynamics},
  author={Voronov, Anton and Khoroshikh, Mikhail and Babenko, Artem and Ryabinin, Max},
  journal={NIPS},
  volume={36},
  pages={37491--37510},
  year={2023}
}

@article{chen2023disenbooth,
  title={DisenBooth: Disentangled Parameter-Efficient Tuning for Subject-Driven Text-to-Image Generation},
  author={Chen, Hong and Zhang, Yipeng and Wang, Xin and Duan, Xuguang and Zhou, Yuwei and Zhu, Wenwu},
  journal={arXiv preprint arXiv:2305.03374},
  year={2023}
}

@inproceedings{zhang2023adding,
  title={Adding conditional control to text-to-image diffusion models},
  author={Zhang, Lvmin and Rao, Anyi and Agrawala, Maneesh},
  booktitle={ICCV},
  pages={3836--3847},
  year={2023}
}

@inproceedings{yu2023freedom,
  title={Freedom: Training-free energy-guided conditional diffusion model},
  author={Yu, Jiwen and Wang, Yinhuai and Zhao, Chen and Ghanem, Bernard and Zhang, Jian},
  booktitle={ICCV},
  pages={23174--23184},
  year={2023}
}

@article{roy2023diffnat,
  title={DIFFNAT: Improving Diffusion Image Quality Using Natural Image Statistics},
  author={Roy, Aniket and Suin, Maiterya and Shah, Anshul and Shah, Ketul and Liu, Jiang and Chellappa, Rama},
  journal={arXiv preprint arXiv:2311.09753},
  year={2023}
}

@inproceedings{ruiz2023hyperdreambooth,
  title={Hyperdreambooth: Hypernetworks for fast personalization of text-to-image models},
  author={Ruiz, Nataniel and Li, Yuanzhen and Jampani, Varun and Wei, Wei and Hou, Tingbo and Pritch, Yael and Wadhwa, Neal and Rubinstein, Michael and Aberman, Kfir},
  booktitle={CVPR},
  pages={6527--6536},
  year={2024}
}

@article{xiao2023fastcomposer,
  title={FastComposer: Tuning-Free Multi-Subject Image Generation with Localized Attention},
  author={Xiao, Guangxuan and Yin, Tianwei and Freeman, William T and Durand, Fr{\'e}do and Han, Song},
  journal={IJCV},
  volume={133},
  number={3},
  pages={1175--1194},
  year={2025},
  publisher={Springer}
}

@article{gu2023mix,
  title={Mix-of-Show: Decentralized Low-Rank Adaptation for Multi-Concept Customization of Diffusion Models},
  author={Gu, Yuchao and Wang, Xintao and Wu, Jay Zhangjie and Shi, Yujun and Chen, Yunpeng and Fan, Zihan and Xiao, Wuyou and Zhao, Rui and Chang, Shuning and Wu, Weijia and others},
  journal={NIPS},
  volume={36},
  pages={15890--15902},
  year={2023}
}

@article{yang2023controllable,
  title={Controllable Textual Inversion for Personalized Text-to-Image Generation},
  author={Yang, Jianan and Wang, Haobo and Zhang, Yanming and Xiao, Ruixuan and Wu, Sai and Chen, Gang and Zhao, Junbo},
  journal={arXiv preprint arXiv:2304.05265},
  year={2023}
}

@inproceedings{cai2024decoupled,
  title={Decoupled textual embeddings for customized image generation},
  author={Cai, Yufei and Wei, Yuxiang and Ji, Zhilong and Bai, Jinfeng and Han, Hu and Zuo, Wangmeng},
  booktitle={AAAI},
  volume={38},
  number={2},
  pages={909--917},
  year={2024}
}

@inproceedings{wei2024dreamvideo,
  title={Dreamvideo: Composing your dream videos with customized subject and motion},
  author={Wei, Yujie and Zhang, Shiwei and Qing, Zhiwu and Yuan, Hangjie and Liu, Zhiheng and Liu, Yu and Zhang, Yingya and Zhou, Jingren and Shan, Hongming},
  booktitle={CVPR},
  pages={6537--6549},
  year={2024}
}

@article{pan2023towards,
  title={Towards Accurate Guided Diffusion Sampling through Symplectic Adjoint Method},
  author={Pan, Jiachun and Yan, Hanshu and Liew, Jun Hao and Feng, Jiashi and Tan, Vincent YF},
  journal={arXiv preprint arXiv:2312.12030},
  year={2023}
}

@inproceedings{wu2023harnessing,
  title={Harnessing the spatial-temporal attention of diffusion models for high-fidelity text-to-image synthesis},
  author={Wu, Qiucheng and Liu, Yujian and Zhao, Handong and Bui, Trung and Lin, Zhe and Zhang, Yang and Chang, Shiyu},
  booktitle={ICCV},
  pages={7766--7776},
  year={2023}
}

@inproceedings{wang2024detdiffusion,
  title={Detdiffusion: Synergizing generative and perceptive models for enhanced data generation and perception},
  author={Wang, Yibo and Gao, Ruiyuan and Chen, Kai and Zhou, Kaiqiang and Cai, Yingjie and Hong, Lanqing and Li, Zhenguo and Jiang, Lihui and Yeung, Dit-Yan and Xu, Qiang and others},
  booktitle={CVPR},
  pages={7246--7255},
  year={2024}
}

@inproceedings{zhang2024attention,
  title={Attention calibration for disentangled text-to-image personalization},
  author={Zhang, Yanbing and Yang, Mengping and Zhou, Qin and Wang, Zhe},
  booktitle={CVPR},
  pages={4764--4774},
  year={2024}
}

@article{zhu2024isolated,
  title={Isolated diffusion: Optimizing multi-concept text-to-image generation training-freely with isolated diffusion guidance},
  author={Zhu, Jingyuan and Ma, Huimin and Chen, Jiansheng and Yuan, Jian},
  journal={IEEE Transactions on Visualization and Computer Graphics},
  year={2024},
  publisher={IEEE}
}

@inproceedings{kim2024selectively,
  title={Selectively informative description can reduce undesired embedding entanglements in text-to-image personalization},
  author={Kim, Jimyeong and Park, Jungwon and Rhee, Wonjong},
  booktitle={CVPR},
  pages={8312--8322},
  year={2024}
}

@inproceedings{frenkel2024implicit,
title={Implicit style-content separation using b-lora},
author={Frenkel, Yarden and Vinker, Yael and Shamir, Ariel and Cohen-Or, Daniel},
booktitle={ECCV},
pages={181--198},
year={2024},
organization={Springer}
}

@inproceedings{zhang2024compositional,
  title={Compositional inversion for stable diffusion models},
  author={Zhang, Xulu and Wei, Xiao-Yong and Wu, Jinlin and Zhang, Tianyi and Zhang, Zhaoxiang and Lei, Zhen and Li, Qing},
  booktitle={AAAI},
  volume={38},
  number={7},
  pages={7350--7358},
  year={2024}
}

@inproceedings{wu2024u,
  title={U-VAP: User-specified visual appearance personalization via decoupled self augmentation},
  author={Wu, You and Liu, Kean and Mi, Xiaoyue and Tang, Fan and Cao, Juan and Li, Jintao},
  booktitle={CVPR},
  pages={9482--9491},
  year={2024}
}

@inproceedings{huang2024reversion,
  title={Reversion: Diffusion-based relation inversion from images},
  author={Huang, Ziqi and Wu, Tianxing and Jiang, Yuming and Chan, Kelvin CK and Liu, Ziwei},
  booktitle={SIGGRAPH Asia 2024 Conference Papers},
  pages={1--11},
  year={2024}
}

@inproceedings{hyung2024magicapture,
  title={Magicapture: High-resolution multi-concept portrait customization},
  author={Hyung, Junha and Shin, Jaeyo and Choo, Jaegul},
  booktitle={AAAI},
  volume={38},
  number={3},
  pages={2445--2453},
  year={2024}
}

@inproceedings{safaee2024clic,
  title={Clic: Concept learning in context},
  author={Safaee, Mehdi and Mikaeili, Aryan and Patashnik, Or and Cohen-Or, Daniel and Mahdavi-Amiri, Ali},
  booktitle={CVPR},
  pages={6924--6933},
  year={2024}
}

@inproceedings{huang2024learning,
  title={Learning disentangled identifiers for action-customized text-to-image generation},
  author={Huang, Siteng and Gong, Biao and Feng, Yutong and Chen, Xi and Fu, Yuqian and Liu, Yu and Wang, Donglin},
  booktitle={CVPR},
  pages={7797--7806},
  year={2024}
}

@inproceedings{chatterjee2024getting,
  title={Getting it right: Improving spatial consistency in text-to-image models},
  author={Chatterjee, Agneet and Stan, Gabriela Ben Melech and Aflalo, Estelle and Paul, Sayak and Ghosh, Dhruba and Gokhale, Tejas and Schmidt, Ludwig and Hajishirzi, Hannaneh and Lal, Vasudev and Baral, Chitta and others},
  booktitle={ECCV},
  pages={204--222},
  year={2024},
  organization={Springer}
}

@article{cao2022lsap,
  title={Lsap: Rethinking inversion fidelity, perception and editability in gan latent space},
  author={Cao, Pu and Yang, Lu and Liu, Dongxu and Liu, Zhiwei and Li, Shan and Song, Qing},
  journal={arXiv preprint arXiv:2209.12746},
  year={2022}
}

@article{devlin2018bert,
  title={Bert: Pre-training of deep bidirectional transformers for language understanding},
  author={Devlin, Jacob and Chang, Ming-Wei and Lee, Kenton and Toutanova, Kristina},
  journal={arXiv preprint arXiv:1810.04805},
  year={2018}
}

@article{dong2022dreamartist,
  title={Dreamartist: Towards controllable one-shot text-to-image generation via contrastive prompt-tuning},
  author={Dong, Ziyi and Wei, Pengxu and Lin, Liang},
  journal={arXiv preprint arXiv:2211.11337},
  year={2022}
}

@article{chefer2023attend,
  title={Attend-and-excite: Attention-based semantic guidance for text-to-image diffusion models},
  author={Chefer, Hila and Alaluf, Yuval and Vinker, Yael and Wolf, Lior and Cohen-Or, Daniel},
  journal={TOG},
  volume={42},
  number={4},
  pages={1--10},
  year={2023},
  publisher={ACM New York, NY, USA}
}

@inproceedings{li2023divide,
  title={Divide \& Bind Your Attention for Improved Generative Semantic Nursing},
author={Li, Yumeng and Keuper, Margret and Zhang, Dan and Khoreva, Anna},
booktitle={BMVC},
year={2023}
}

@article{liu2023cones2,
  title={Cones 2: Customizable Image Synthesis with Multiple Subjects},
  author={Liu, Zhiheng and Feng, Ruili and Zhu, Kai and Zhang, Yifei and Zheng, Kecheng and Liu, Yu and Zhao, Deli and Zhou, Jingren and Cao, Yang},
  journal={arXiv preprint arXiv:2303.05125},
  year={2023}
}

@inproceedings{phung2023grounded,
  title={Grounded Text-to-Image Synthesis with Attention Refocusing},
  author={Phung, Quynh and Ge, Songwei and Huang, Jia-Bin},
  booktitle={CVPR},
  pages={7932--7942},
  year={2024}
}

@inproceedings{Kim2023DenseTG,
title={Dense Text-to-Image Generation with Attention Modulation},
  author={Kim, Yunji and Lee, Jiyoung and Kim, Jin-Hwa and Ha, Jung-Woo and Zhu, Jun-Yan},
  booktitle={ICCV},
  pages={7701--7711},
  year={2023}
}

@article{wang2023enhancing,
  title={Enhancing Object Coherence in Layout-to-Image Synthesis},
  author={Wang, Yibin and Xu, Honghui and Zhou, Changhai and Zhang, Weizhong and Jin, Cheng},
  journal={arXiv preprint arXiv:2311.10522},
  year={2023}
}

@article{zhao2023loco,
  title={LoCo: Locally Constrained Training-Free Layout-to-Image Synthesis},
  author={Zhao, Peiang and Li, Han and Jin, Ruiyang and Zhou, S Kevin},
  journal={arXiv preprint arXiv:2311.12342},
  year={2023}
}

@article{wu2023paragraph,
  title={Paragraph-to-Image Generation with Information-Enriched Diffusion Model},
  author={Wu, Weijia and Li, Zhuang and He, Yefei and Shou, Mike Zheng and Shen, Chunhua and Cheng, Lele and Li, Yan and Gao, Tingting and Zhang, Di},
  journal={IJCV},
  pages={1--22},
  year={2025},
  publisher={Springer}
}

@inproceedings{po2023orthogonal,
  title={Orthogonal Adaptation for Modular Customization of Diffusion Models},
  author={Po, Ryan and Yang, Guandao and Aberman, Kfir and Wetzstein, Gordon},
  booktitle={CVPR},
  pages={7964--7973},
  year={2024}
}

@inproceedings{cao2025image,
  title={Image is All You Need to Empower Large-scale Diffusion Models for In-Domain Generation},
author={Cao, Pu and Zhou, Feng and Yang, Lu and Huang, Tianrui and Song, Qing},
booktitle={CVPR},
pages={18358--18368},
year={2025}
}

@inproceedings{chan2024improving,
  title={Improving subject-driven image synthesis with subject-agnostic guidance},
  author={Chan, Kelvin CK and Zhao, Yang and Jia, Xuhui and Yang, Ming-Hsuan and Wang, Huisheng},
  booktitle={CVPR},
  pages={6733--6742},
  year={2024}
}

}
\vspace{-15mm}
\begin{IEEEbiography}[{\includegraphics[width=1in,height=1.25in,clip,keepaspectratio]{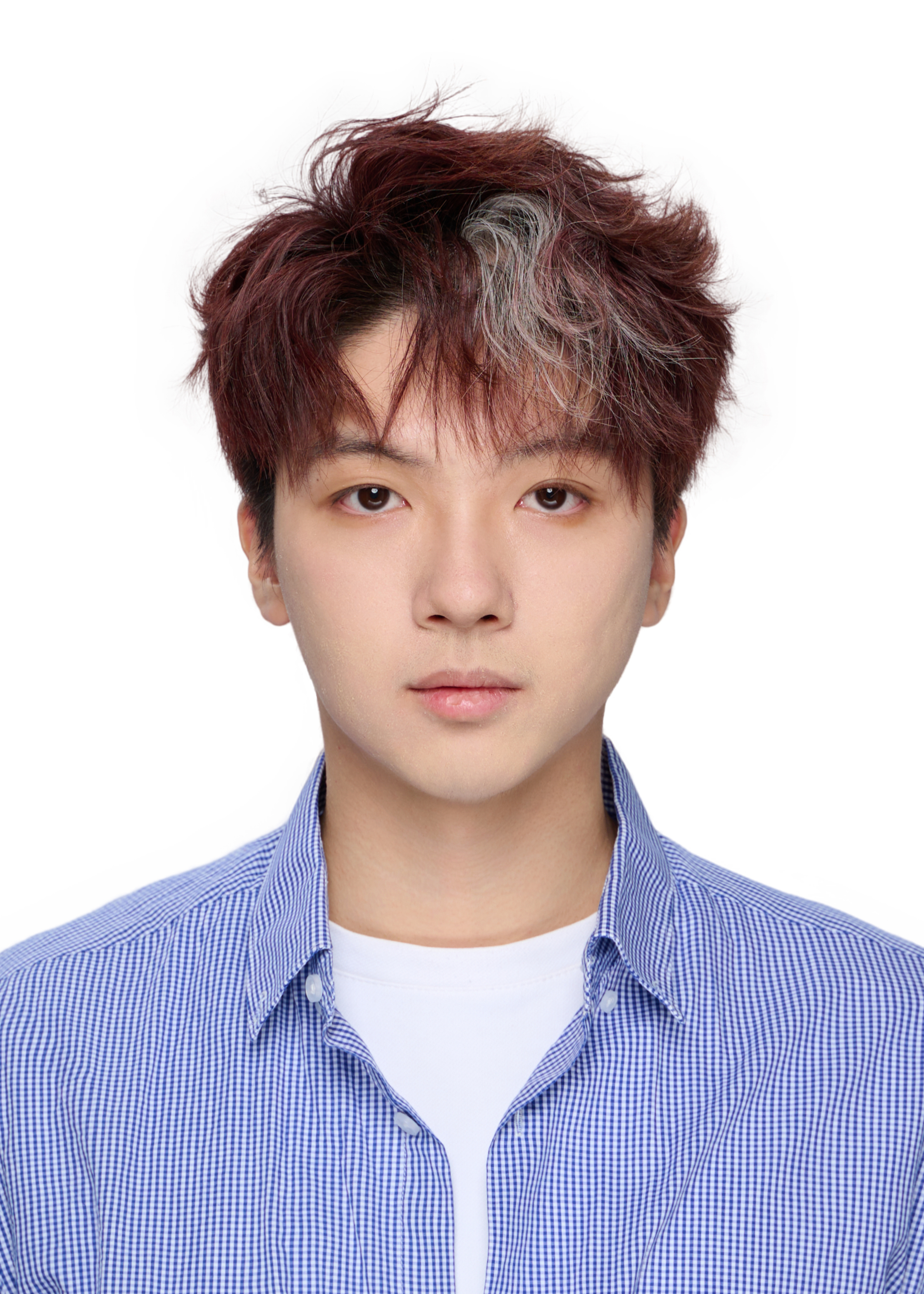}}]{Pu Cao}
received his bachelor’s degree from University of Science and Technology Beijing (USTB), Beijing, China, in 2022, and is currently a Ph.D. candidate at the School of Intelligent Engineering and Automation, Beijing University of Posts and Telecommunications (BUPT), since 2022. His research interests include multimodal understanding and generation, especially focusing on multimodal large-language models and diffusion models.
\end{IEEEbiography}
\vspace{-15mm}
\begin{IEEEbiography}[{\includegraphics[width=1in,height=1.25in,clip,keepaspectratio]{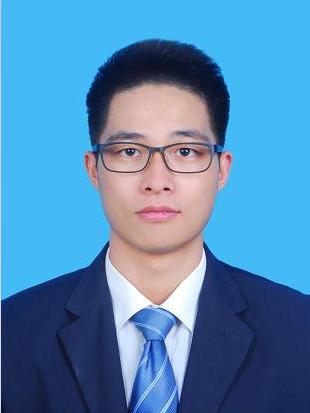}}]{Feng Zhou}
received his bachelor’s degree from Beijing University of Posts and Telecommunications, Beijing, China, in 2022, and is currently a Ph.D. candidate at the School of Intelligent Engineering and Automation, Beijing University of Posts and Telecommunications. His research interests include computer vision, and machine learning, especially
focusing on 3d vision.
\end{IEEEbiography}
\vspace{-15mm}
\begin{IEEEbiography}[{\includegraphics[width=1in,height=1.25in,clip,keepaspectratio]{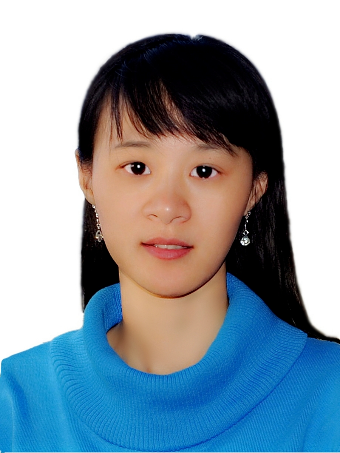}}]{Qing Song}
received the Ph.D. degree from Tianjin University, Tianjin, China, in 2006. She is currently a Scientific Researcher with Beijing University of Posts and Telecommunications (BUPT), where she is engaged in the field of computer vision technology. She is the Founder of the Pattern Recognition and Intelligent Vision Laboratory (PRIV). 
\end{IEEEbiography}
\vspace{-15mm}
\begin{IEEEbiography}[{\includegraphics[width=1in,height=1.25in,clip,keepaspectratio]{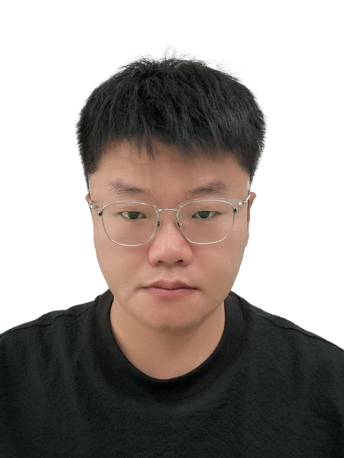}}]{Lu Yang}
is currently an associate professor in the Beijing University of Posts and Telecommunications (BUPT), China. He received his Ph.D. degree from the BUPT in 2021. He has been involved in research work with the Pattern Recognition and Intelligent Vision Laboratory (PRIV), since 2012. His current research interests include the fields of HumanCentric Al and GenAI.
\end{IEEEbiography}

\end{document}